\documentclass{article}

\PassOptionsToPackage{numbers, compress}{natbib}

\usepackage[preprint]{neurips_2026}

\usepackage[utf8]{inputenc}
\usepackage[T1]{fontenc}
\usepackage{hyperref}
\usepackage{xurl}
\usepackage{booktabs}
\usepackage{tabularx}
\usepackage{colortbl}
\usepackage{multirow}
\usepackage{array}
\usepackage{arydshln}
\usepackage{amsfonts}
\usepackage{nicefrac}
\usepackage{microtype}
\usepackage{xcolor}
\usepackage{enumitem}
\usepackage{pifont}
\usepackage{fontawesome5}
\usepackage[most]{tcolorbox}
\usepackage{graphicx}
\usepackage{wrapfig}
\usepackage{tikz}
\usepackage[font=footnotesize]{caption}
\usepackage{etoc}
\usepackage{placeins}
\newcommand{\cmark}{\ding{51}}
\newcommand{\xmark}{\ding{55}}
\newcommand{\circled}[1]{\tikz[baseline=(char.base)]{\node[shape=circle,draw,inner sep=1pt,minimum size=12pt,font=\scriptsize] (char) {#1};}}
\definecolor{agentrow}{HTML}{F0F6EE}
\definecolor{modelrow}{HTML}{FFF6E8}
\definecolor{harnessrow}{HTML}{EEF5FB}
\definecolor{termrow}{HTML}{F5F0FA}
\definecolor{lastexamred}{HTML}{D00000}
\definecolor{sc0}{HTML}{9C2F2F}
\definecolor{sc1}{HTML}{B45F06}
\definecolor{sc2}{HTML}{8A6D1D}
\definecolor{sc3}{HTML}{4F7C2C}
\definecolor{sc4}{HTML}{2E6F3E}
\definecolor{sc5}{HTML}{1D5B35}
\newcommand{\hsc}[1]{%
  \ifdim #1 pt < 5pt \textcolor{sc0}{#1}%
  \else\ifdim #1 pt < 20pt \textcolor{sc1}{#1}%
  \else\ifdim #1 pt < 40pt \textcolor{sc2}{#1}%
  \else\ifdim #1 pt < 60pt \textcolor{sc3}{#1}%
  \else\ifdim #1 pt < 80pt \textcolor{sc4}{#1}%
  \else \textcolor{sc5}{#1}%
  \fi\fi\fi\fi\fi}
\newcommand{\hscb}[1]{%
  \ifdim #1 pt < 5pt \textcolor{sc0}{\textbf{#1}}%
  \else\ifdim #1 pt < 20pt \textcolor{sc1}{\textbf{#1}}%
  \else\ifdim #1 pt < 40pt \textcolor{sc2}{\textbf{#1}}%
  \else\ifdim #1 pt < 60pt \textcolor{sc3}{\textbf{#1}}%
  \else\ifdim #1 pt < 80pt \textcolor{sc4}{\textbf{#1}}%
  \else \textcolor{sc5}{\textbf{#1}}%
  \fi\fi\fi\fi\fi}

\definecolor{papertodobg}{HTML}{FFF8E1}
\definecolor{papertodoframe}{HTML}{E65100}
\newcommand{\papertodotag}{%
  \tcbox[on line, arc=2pt, boxrule=0.45pt,%
    colback=papertodobg, colframe=papertodoframe!55!black,%
    boxsep=2pt, left=3.5pt, right=3.5pt, top=1pt, bottom=1pt,%
    nobeforeafter]{\textbf{[TODO]}}}

\newtcolorbox{lastexambox}{%
  enhanced, sharp corners,
  colback=lastexamred!4, colframe=lastexamred!70!black,
  boxrule=0pt, leftrule=3pt,
  left=7pt, right=7pt, top=5pt, bottom=5pt,
  before skip=6pt, after skip=8pt,
  fontupper=\small
}
\renewcommand{\answerTODO}[1][]{\papertodotag}
\renewcommand{\justificationTODO}[1][]{\papertodotag}

\graphicspath{{./}}
\newcommand{\taskcount}{55}
\newcommand{\clustercount}{13}

\newcommand{\totaltasks}{960}
\newcommand{\totalvariants}{1{,}490}
\newcommand{\neartermcount}{67}
\newcommand{\fullspectrumcount}{55}
\newcommand{\lastexamcount}{38}
\newcommand{\publictaskcount}{152}
\newcommand{\aleclicount}{105}

\definecolor{halinkcol}{HTML}{1B4F9C}
\newcommand{\hficon}{\raisebox{-0.22\height}{\includegraphics[height=1.02em]{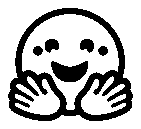}}}
\newcommand{\halink}[3]{{\hypersetup{hidelinks}\href{#3}{{\color{black!80}#1}\,\textcolor{halinkcol}{#2}}}}
\newcommand{\halinksep}{\hspace{0.9em}\textcolor{black!25}{\rule[-0.15ex]{0.5pt}{1.5ex}}\hspace{0.9em}}
\newcommand{\resourcelinks}{%
  {\small
    \halink{\faGlobe}{Website}{https://agents-last-exam.org}\halinksep
    \halink{\faGithub}{GitHub}{https://github.com/rdi-berkeley/agents-last-exam}\halinksep
    \halink{\hficon}{HuggingFace}{https://huggingface.co/datasets/agents-last-exam/agents-last-exam}\halinksep
    \halink{\faTrophy}{Leaderboard}{https://agents-last-exam.org/leaderboard}%
  }}

\title{%
  \raisebox{-0.12\height}{\includegraphics[height=1.15em]{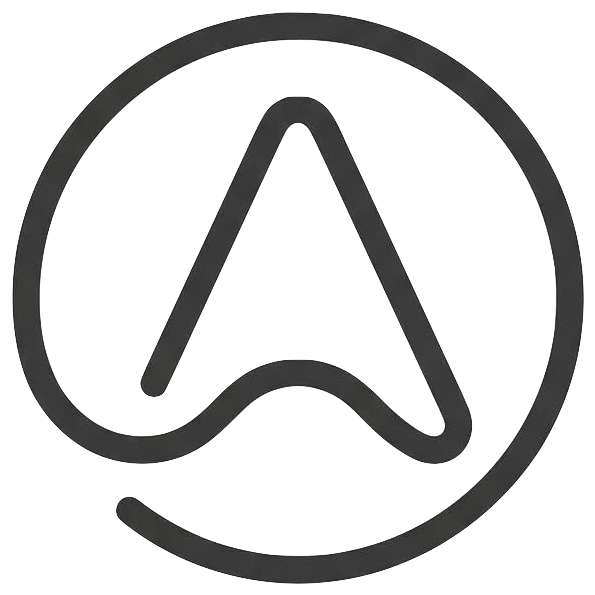}}\hspace{0.35em}%
  Agents' Last Exam%
}

\author{%
}

\vspace{-1cm}

\begin{document}

\maketitle
\vspace{-5.2em}
\centerline{\resourcelinks}
\vspace{1.8em}
\begingroup
\setlength{\parindent}{0pt}
\setlength{\parskip}{0pt}
\centering\small

{\normalsize\bfseries Organization \& Execution Team}\par\vspace{4pt}
Yiyou Sun\textsuperscript{*}, Xinyang Han\textsuperscript{*}, Weichen Zhang\textsuperscript{*}, Yuanbo Pang\textsuperscript{*}, Tianyu Wang\textsuperscript{*}, Yuhan Cao\textsuperscript{*}, Yixiao Huang\textsuperscript{*}, Xiang Liu, Zonglin Di, Chris Duroiu, Haoyun Zhang, Jeffrey Lin, Weishu Zhang, Tyler Zeng, Ying Yan, Bo Liu, Hanson Wen, Mingyang Xu, Xiaoyuan Liu, Zimeng Chen, Zhaowei Zhang, Weiyan Shi, Amanda Dsouza, Vincent Sunn Chen, Jingxu Xie, Dewen Lei, Hanxuan Zhang, Yinhe Wang, Xinran Zhang, Chao Zhou, Sheldon Huang, Dawn Song\par
\vspace{2pt}
{\footnotesize * Core contributors.\par}
\vspace{8pt}

{\normalsize\bfseries Advisory Committee}\par\vspace{4pt}
Patrick Bryant, Carl Boettiger, Yamini Rangan, Bradley Rothenberg, Kyle Steinfeld, Arvind Rao, Tapio Schneider, Georgios Yannakakis, Laure Zanna, Kaan Ozbay, Ida Sim, Tarek Zohdi, George Em Karniadakis, Jack Gallant, Teresa Head-Gordon\par
\vspace{8pt}

{\normalsize\bfseries Data Contributors}\par\vspace{4pt}
{\footnotesize
Chris Yuhao Liu, Rongwang Hu, Yafei Cheng, Aras Bacho, Zengyi Qin, Alan Yuille, Yixiong Chen, Hao Liu, Lin Zeng, Bo Chen, Zhendong Li, Litian Gong, Shashank Muralidhar Bharadwaj, Thorben Jansen, HAIXIANG TANG, Jianhong Tu, Jixian Liu, Maojia Song, Ruheng Wang, Shengcao Cao, Yanxin Zhang, Yingxuan Yang, Zongzheng Zhang, Honglin Bao, Hyunjun Kim, Peiran Xu, Shuo Lu, Zheng Zhang, Zhonghua Wang, Aditya Prakash, Bohan Lyu, Chang Ma, Haoxuan Chen, Hengduan Fan, Molei Tao, Roman Sitelew, Russell Alan Poldrack, Yang Liu, Yang Liu, Yichen Wu, Yuanze Lin, Zhenglu Li, Zibu Wei, Zijiao Chen, Abderrahim Bendahi, Alexander Schubert, Chase Broderick, Chengzhi Liu, Haoran Sun, Haoyang Li, Haoyue Hua, Jiayuan Ding, Jing Huang, Shangding Gu, Tuan Vinh, Xiatao Sun, Yanqiong Jiang, Yiran Hu, Zechen Xu, Ziao Yang, Chenguang Wang, Costas Spanos, Dongrong Yang, Haocheng Wang, Jiayi Cheng, JICHENG WANG, Junbo Peng, Kangqi Zhang, Keying Kuang, Lichao Sun, Ming Yin, Orson Xu, Vincent Siu, Wanzhe Liao, Wenqi Shi, Wenxi Deng, Xiaopan Zhang, Yangjunfeng Zhu, Yifan Shen, Yi Jiang, Yinsheng Li, Youbang Pan, Yuan Li, Yuchen Zhu, Yunshi Zhang, Zhenning Yang, Ziyan Rao, Alireza Rafiei, Andrew Glover, Bixuan Ren, Boqin Yuan, Chonghao Cai, Deren Lei, Hanwen Xing, Haoran Jin, Hector Gomez, Ishaan Singh Chandok, Jiaming Hu, Jianxin Sun, Justin Purnomo, Kunyang Sun, Lei Ding, Longtai Liao, Lukas Kim, Ming Liu, Qingpo Wuwu, Ran Jin, Ruiqi Zhang, Ting Sun, Wenbo Huang, Xuandong Zhao, Peter Chen, Chuanyang Jin, Qiming Bao, Xuhuan Shen, Yifan Zhang, Yi Li, Yingze Wang, Yixin Wang, Yushi Yang, Zhen Dong, Zhiling Yan, Zhongxing Xu, Zibo Zhao, Zixin Jiang, Ali Emami, Andy Zeyi Liu, Angela Wu, Aylin Caliskan, Baihe Huang, Baixiang Huang, Benjamin Yang, Benran Zhang, Brian Zhao, Carlo Bosio, Charlie Guo, Che Liu, Chenchen Du, Chen Dai, Chenlong Zhang, Chenyu Li, Chenyu Zhu, Chi Zhang, Core Francisco Park, Daofeng Li, Dehao Wu, Dowling Wong, Erin Kasson, Fangyu Wu, Fan Xu, Fiona Shen, Haimin Hu, Haitao Li, Hang Zhou, Hannah Szlyk, Hanning Zhang, Hanwen Zhang, Hao Fan, Haoran Ye, HaoYang Shang, Haoyue Bai, Hengyu Fu, Hongwei Li, Huiqi Wang, Jiachen Li, Jialu Wang, Jiaming Liu, Jiawei Shen, Jierui Zuo, Jingxuan Fan, Jingyan Li, Jirong Yang, Jonah So, Jorin Overwiening, Joseph Cavanagh, Joseph Lee, Junge Zhang, Kaiyuan Liu, Kaustubh Dhole, Kevin Qinghong Lin, Kunpeng Zhang, Kyle Montgomery, Layna Paraboschi, Liana, Lihui Liu, Liwei Zhou, Lu Lu, Mengyuan ``Millie'' Wu, Mikko Tolonen, Mingxuan Yan, Minqi Shi, Murphy Niu, Pan Lu, Patricia Cavazos-Rehg, Qijian Zhao, Qiuyang Mang, Rimika Jaiswal, Ruiling Xu, Run-Ze Fan, Ruoxi Ning, Santanu Bosu Antu, Scott Rankin, Sen Yang, Shi Gu, Shi Qiu, Sichen Zhu, Sihan Xu, Siqi Zhu, Song Zhang, Sophia Sun, Stefano De Castro, Swati Rajwal, Tianneng Shi, Tianxing Chen, Tianyue Zhou, Ting Wang, Ventus Shaw, Vladislav Susoy, Weihang Xiao, Wei Wang, Wenbo Lv, Wenjin Zhang, Wenshuo Zhang, Wenyu Liang, Xiaohu Yu, Xiao Li, Xiaoli Yang, Xin Lan, Xinyu Wang, Xuan Liu, Xuewen Chen, Xu Pan, Yan Zhu, Yicheng Lyu, Yifan Gu, Yimin Zhao, Ying Zhang, Yin Ren, Yipu Chen, Yiren Rong, Yi Shao, Yishu Zhu, Yizhi Zhang, Yongyi Xiong, Yuangang Li, Yuchen Wu, Yuhaohua Zheng, Yuhao Wu, Yunfu Deng, Yushan Li, Yu Wu, Yuzheng Xu, Yuzhou Nie, Zehao Wang, Zhehang Du, Zhengming Yu, Zhihan Jiang, Zhi Li, Zhixu Li, Zhuoran Yi, Tao Sun, Zhun Wang, Justin Xu, Shi Bo, Helena Casademunt, Wenjia Chen, Cheng Cheng, Nawen Deng, Rao Fu, Tianfu Fu, Yifan Han, Zhenyu He, Qiao Jin, Langlang Li, Yuetai Li, Sylvia Liu, Subhabrata Mukherjee, Yunqi Ouyang, He Ren, Dawei Shi, Haoran Wu, Zhiyue Wu, Hannah Yao, Jenny Yu, Rhea Zhan, Luqing Zhou, Blake Zhu, Junfan Zhu, Sanjiv Das, Yanjun Gao, Wenbo Guo, Erika J Schneider, Bing Liu, Zhiyong Lu, Yian Ma, Mark Mueller, Radha Poovendran, Somayeh Sojoudi, Huaxiu Yao, Yinglun Zhu, Károly Zsolnai-Fehér, Yegor Denisov-Blanch, Changran Hu, Vincent Koc, Hezheng Yin, Naomi Xia, Xinyi Bai, Arsh Shah Dilbagi, Jing Gu, Yutong Liu, Lin Shi, Qizheng Zhang, Yuexing Hao, Xiaoming Li, Xiangyi Li, Bingran You, Wenjun Ke, Daoan Zhang, Ari Nemet, Tianyi Zhang, Xingda Lyu, Pranit Chawla, Genta Indra Winata, Maosen Tang, Purva Chiniya, Qiuyi Chen, Steven Dillmann, Wanghan Xu, Wenqi Yang, Xiaojun Wu, Xin Luo, Yue Huang\par}
\vspace{8pt}

{\footnotesize Leading institution: University of California, Berkeley.\quad Corresponding to \texttt{\{sunyiyou,dawnsong\}@berkeley.edu}.\quad \\ Full affiliations in Appendix~\ref{app:authors}.}\par
\endgroup

\clearpage
\begin{abstract}
\vspace{-1em}
Recent AI systems have achieved strong results on a wide range of benchmarks, yet these gains have not translated into economically meaningful deployment across many professional domains. We argue that this gap is largely an evaluation problem: widely used benchmarks lack sustained performance measurement on real and economically valuable workflows. This paper introduces \textbf{Agents' Last Exam (ALE)}, a benchmark designed to evaluate AI agents on long horizon, economically valuable, real world tasks with verifiable outcomes. Developed in collaboration with 250+ industry experts, ALE covers non-physical industries defined with reference to O*NET / SOC 2018 (the U.S. federal occupational taxonomy). It is organized around a task taxonomy with \taskcount{} sub fields grouped into \clustercount{} industry clusters covering 1K+ tasks. Current results show that the hardest tier remains far from saturated: across mainstream harness and backbone configurations, the average full pass rate is below 1\%. ALE is designed as a living benchmark: its task pool grows continuously as new workflows and industries are onboarded. More broadly, ALE is intended not merely as another leaderboard, but as an instrument for closing the gap between benchmark success and GDP relevant impact.
\end{abstract}

\vspace{-1em}
\begin{figure}[h]
\centering
\makebox[\textwidth][c]{%
\includegraphics[width=1.1\linewidth]{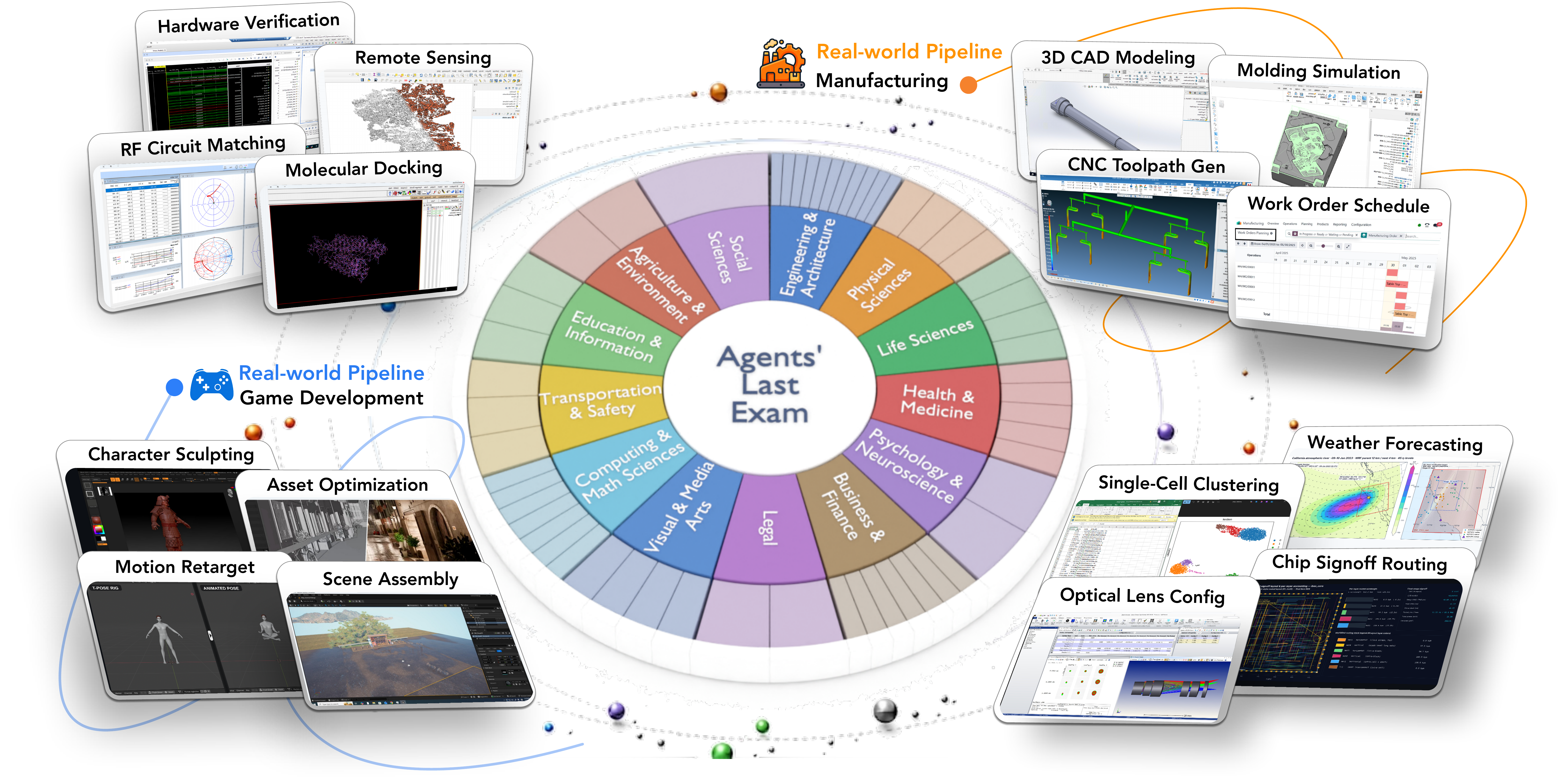}%
}
\caption{\textbf{Agents' Last Exam} spans a broad taxonomy of professional tasks and realistic workflows.}
\label{fig:teaser}
\end{figure}

\begin{figure}[t!]
  \centering
  \makebox[\textwidth][c]{%
    \includegraphics[width=1.17\textwidth]{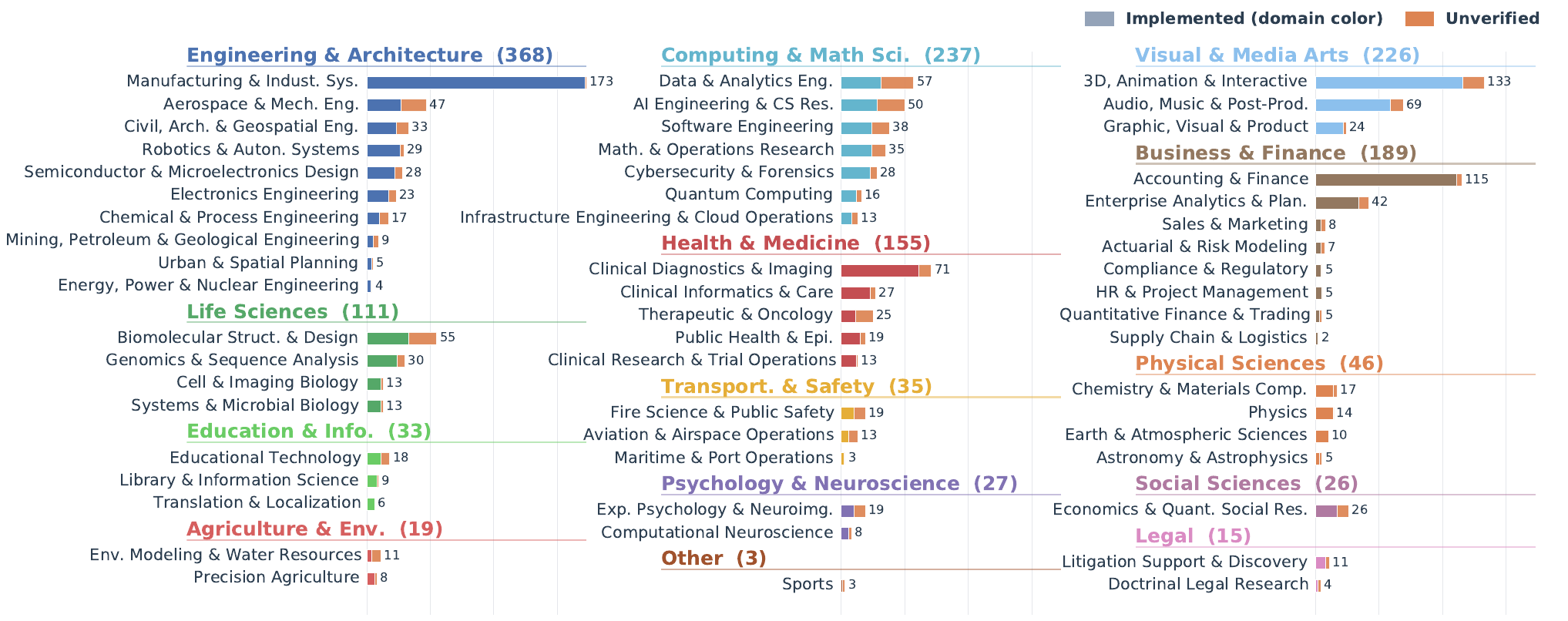}%
  }
  \vspace{-1.5em}
  \caption{\textbf{Distribution of \totalvariants{} task instances across the ALE taxonomy.} Each row is one of the \taskcount{} subdomains, grouped under the \clustercount{} top-level domains (parenthetical numbers give domain totals). Stacked bars decompose each subdomain into fully-implemented instances (domain color) and expert submissions awaiting Quality Control (QC) Process (orange). All \taskcount{} subdomains receive non-zero coverage. Current runnable task instances target either Linux or Windows virtual machines.}
  \label{fig:subdomain-distribution}
  \vspace{-1.6em}
\end{figure}

\vspace{-1em}
\section{Introduction}
\vspace{-.5em}
Over the past few years, AI systems have cleared one celebrated benchmark after another: world-champion
games~\citep{silver2016alphago}, olympiad mathematics~\citep{deepmind2025geminiimo}, and
competitive programming~\citep{google2025geminiicpc}. Yet by the metric that ultimately matters, economic output, the broader impact has remained surprisingly muted; benchmark victories have accumulated faster than measurable transformation in core industries. This gap, which we view as a utility problem for AI, suggests that the field now needs evaluations that measure not only abstract competence, but also the ability to carry out long-horizon, economically valuable work in real professional environments. 

This gap matters because AI progress is remarkably shaped by the benchmarks the field chooses to optimize. Benchmarks do not merely record capability; they focus research attention, define engineering targets, and often determine which domains become tractable for rapid improvement. The recent history of AI makes this pattern clear: once a domain is captured by a verifiable and widely used evaluation, progress in that domain tends to accelerate, and deployment often follows, like ImageNet~\citep{deng2009imagenet} played this role in computer vision. For economically central sectors such as finance, law, electrical engineering, and manufacturing, however, comparable evaluations remain underdeveloped. If such benchmarks can be built and eventually saturated, that outcome would signify more than success on a test: it would indicate that AI systems have become capable of performing the underlying professional workflows at a level sufficient for real industrial adoption.

Building such evaluations is difficult for structural reasons. First, long-horizon authentic workflows are expensive to collect because they must be sourced from real software and organizational contexts. Prior work often adopts task units that are easier to collect, whether shorter computer-use tasks~\citep{xie2024osworld}, synthetic environment construction~\citep{aggarwal2026gymanything}, or purely question-answering setups~\citep{yang2026onemillionbench}. Second, broad industry coverage with authentic, economically valuable workflows is also hard. It requires sustained access to experts across domains and deep insight into the industrial landscape. Existing benchmarks often evaluate on a limited set of domains~\citep{alphaeval2026}. Third, verification is intrinsically hard for real workflows because the output space is heterogeneous. A correct deliverable may be a file, spreadsheet, media artifact, report, design, or model. As a result, benchmarks that measure economically valuable work often rely on human judgment, as in GDPval~\citep{patwardhan2025gdpval} and the Remote Labor Index~\citep{mazeika2025rli}. These constraints help explain why existing benchmarks often trade away one of realism, breadth, or verifiability. They jointly motivate \textbf{Agents' Last Exam (ALE)}.

\begin{lastexambox}
\textbf{Why ``Last Exam''?}\enspace
The name carries a dual aspiration.
\emph{Last as competence threshold}: an agent that passes these industry exams demonstrates readiness to carry out sustained, economically valuable work in that profession, not merely to answer questions about it.
\emph{Last as difficulty frontier}: by grounding evaluation in authentic long-horizon workflows that require professional judgment, ALE sits at the boundary of what current systems can reliably accomplish.
\end{lastexambox}

\begin{wrapfigure}{r}{0.45\textwidth}
    \vspace{-0.8em}
    \centering
    \includegraphics{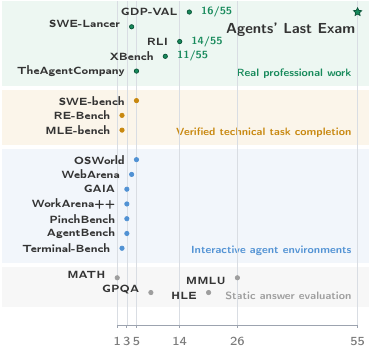}
    \vspace{-0.9em}
    \caption{\textbf{Benchmark positioning map.} Prior benchmarks are placed by mapping their
    published domains onto the ALE domain taxonomy.}
    \label{fig:positioning}
    \vspace{-1.3em}
  \end{wrapfigure}
\noindent ALE is a benchmark of 1K+ task instances spanning \taskcount{} subfields and \clustercount{} industry clusters, developed in collaboration with 250+ domain experts. To ensure broad and representative industry coverage, expert advisory committees map each domain's workflow landscape and identify economically meaningful workflow families, anchored in the O*NET / SOC 2018 occupational taxonomy~\citep{peterson2001onet,bls2018soc}. Its task workflows are sourced from real professional practice: rather than inventing synthetic scenarios, experts contribute projects they have already completed, which then undergo multi-round quality control, including first-pass review, engineer dry-runs, and final peer review by expert committees, before admission. Most tasks demand computer use that interleaves GUI interaction (desktop applications, browsers, domain-specific software) with CLI operations (shell scripting, code execution, file manipulation), requiring the union of capabilities that existing benchmarks test in isolation. To make heterogeneous real-world outputs verifiable without human judges, ALE standardizes evaluation around structured deliverable-based or milestone-based checks against expert-provided references and rubrics.

ALE's target evaluation subject is the \emph{Generalist Computer-Use Agent} (GCUA), such as Claude Code~\citep{anthropic2025claudecode} or Codex~\citep{openai2025codex}, that combines visual perception, code execution, tool use, and long-horizon planning within a single action loop. ALE's task surface is, by construction, a superset of GUI-only benchmarks like OSWorld~\citep{xie2024osworld} and CLI-only benchmarks like Terminal-Bench~\citep{merrill2026terminalbench}. For coverage comparisons, we use the \taskcount{} ALE subdomains as a common coordinate system and map each prior benchmark's published subjects, applications, repositories, or occupations onto this taxonomy (Figure~\ref{fig:positioning}). Current results confirm that ALE is far from saturated: the strongest configuration (Codex with GPT-5.5), which already achieves 82\% on Terminal-Bench, scores below 50\% even on ALE's easiest tier and under 10\% on the hardest; most mainstream agents, including Claude Code, record near-zero pass rates at that difficulty level.

More broadly, ALE is intended not merely as another leaderboard, but as an instrument for closing the gap between benchmark success and GDP-relevant impact: \textbf{if frontier AI agents can pass this last exam, then progress on the benchmark may begin to register as real economic transformation.}

\vspace{-0.6em}
\section{Benchmark Design and Dataset Construction}
\label{sec:dataset}
\vspace{-0.6em}
\subsection{Benchmark Design Principles: What Tasks are We Looking for?}
\label{sec:design-principles}

The benchmark is defined by three high-level requirements. They determine which workflows are
admitted into the dataset and which are rejected in the public submission portal:

\noindent \textbullet\ \textbf{Representativeness.} The workflow should match real professional practice and use the software that domain experts would actually use. 
\noindent For example, architectural experts would typically use \colorbox{green!8}{SolidWorks or Rhino} rather than \colorbox{red!8}{AutoCAD} to convert a 2D blueprint into a 3D model.

\noindent \textbullet\ \textbf{Complexity.} A task should be an end-to-end deliverable that would take an expert substantial time, rather than only a few UI operations. The key distinction is between a workflow and an action.

\noindent \colorbox{red!8}{\parbox{\dimexpr\linewidth-2\fboxsep\relax}{\textit{Undesired example.} ``Apply a color filter in DaVinci'' is too narrow because it's a single local edit.}}

\noindent \colorbox{green!8}{\parbox{\dimexpr\linewidth-2\fboxsep\relax}{\textit{Better example.} ``Move a running cheetah into another race video'' is suitable because it requires tracking, rotoscoping, compositing, and color matching within one coupled workflow.}}

\noindent \textbullet\ \textbf{Verifiability.} The output should admit deterministic checking or an unambiguous rubric tied to observable artifacts. The strongest case is a deterministic deliverable that can be compared directly against a reference output. When exact matching is impossible, the judgment should still reduce to a measurable artifact.

\noindent \colorbox{red!8}{\parbox{\dimexpr\linewidth-2\fboxsep\relax}{\textit{Undesired example.} ``Design an RPG game with monsters'' provides no objectively checkable target.}}

\noindent \colorbox{green!8}{\parbox{\dimexpr\linewidth-2\fboxsep\relax}{\textit{Better example.} ``Reproduce the game \textit{mota.exe} using RPGMaker XP'' is verifiable because the resulting map geometry, character attributes, and event states can be automatically compared against a reference version under identical trajectories of user operations.}}

\vspace{-0.4em}
\subsection{Benchmark Scope and Taxonomy: Which Domains Do We Cover?}
\label{sec:taxonomy}
\vspace{-0.4em}
Rather than selecting industries ad hoc or by economic ranking, we ground the ALE taxonomy in SOC 2018~\citep{bls2018soc} and O*NET~\citep{peterson2001onet}: we cluster occupations with similar software-mediated workflows into ALE industries, exclude sectors whose core work is not meaningfully digital, and group the result into \clustercount{} domains spanning \taskcount{} subdomains (Figure~\ref{fig:teaser}; full derivation in Appendix~\ref{app:taxonomy}). To enable fair cross-benchmark comparison, we map each prior benchmark's published categories (subjects, applications, repositories, or occupations) onto the same \taskcount{}-subdomain taxonomy via an LLM-assisted classifier. The result exposes a coverage gap that no existing benchmark closes: even the union of 16 major prior benchmarks leaves \textbf{13 of \taskcount{} subdomains} entirely uncovered (Table~\ref{tab:related-comparison}).

\vspace{-0.4em}
\subsection{Task Construction Pipeline: How are the Tasks Created?}
\label{sec:construction-pipeline}

\begin{figure}[htbp]
  \centering
  \includegraphics[width=1.0\linewidth]{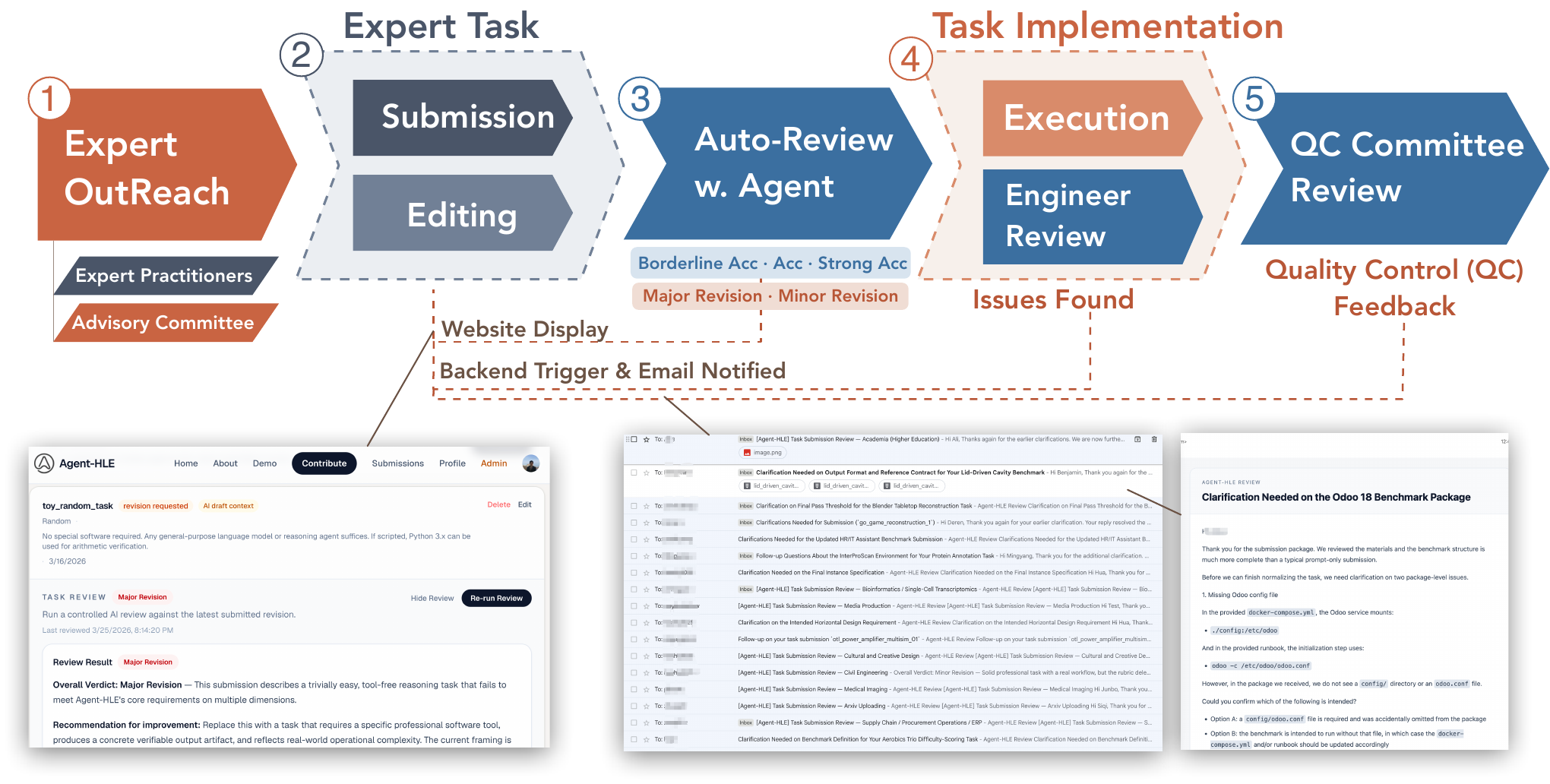}
  \caption{\textbf{Task construction pipeline.} Tasks proceed from expert sourcing through submission, first-pass review, engineering implementation, and final quality control.}
  \label{fig:pipeline}
  \vspace{-0.3cm}
\end{figure}

\begin{wrapfigure}{r}{0.5\textwidth}
  \vspace{-0.4em}
  \centering
  \includegraphics[width=\linewidth]{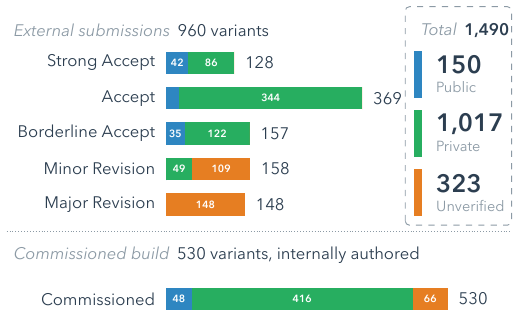}
  \caption{\textbf{Provenance and review yield.} The \totalvariants{} task instances split into 960 external submissions (top, by first-pass review verdict) and 530 commissioned tasks (bottom). Each bar is segmented by release state: 150 public, 1{,}017 private, and 323 unverified pending QC.}
  \label{fig:pipeline-yield}
  \vspace{-0.5em}
\end{wrapfigure}
Tasks in ALE cannot be crowdsourced from lay workers; they must arise from the actual routines of domain professionals and undergo rigorous screening to guarantee authenticity, complexity, and technical executability. We therefore employ a staged construction protocol (Figure~\ref{fig:pipeline}) with five gates. \textit{Expert sourcing} recruits domain specialists through an advisory committee of industry practitioners, ensuring coverage across the taxonomy. \textit{Task submission} routes proposals through a dedicated web portal (\url{https://agents-last-exam.org/submit/new/form}), where experts upload past projects that took them days or weeks of professional work and AI-assisted tools help refine each proposal until five core components are fully specified: a natural-language description, input files, target software, expected deliverable, and evaluation specification.
\textit{First-pass review} screens submissions with conference-style decisions (\texttt{major / minor revision}, \texttt{borderline accept}, \texttt{accept}, \texttt{strong accept}); revisions loop back to the expert. \textit{Task implementation} converts accepted specifications into runnable assets, provisioned software containers, and codified evaluation logic, with engineer dry-runs and automatic routing back to the expert when gaps are found. \textit{Final QC} is a peer review by the expert committee that verifies reference-output correctness, calibration of evaluation bounds (neither impossibly narrow nor spuriously permissive), and sufficiency of context before the task is admitted. More details are deferred to Appendix~\ref{app:construction-pipeline}.

\noindent \textbf{Public/private release strategy.}
Benchmark contamination, whether through pre-training data overlap or task-specific optimization, is a central threat to the long-term validity of any public evaluation.
ALE addresses this by releasing only 150 of the \totalvariants{} task instances ($\sim$10\%) publicly, with the remainder held in a private pool (Figure~\ref{fig:pipeline-yield}).
ALE is further designed for \emph{rolling evaluation}: private task instances will periodically rotate into the public set while retired public tasks are replaced, maintaining an uncontaminated evaluation surface over successive model generations.
Appendix~\ref{app:fullpool} verifies empirically that the public subset is representative of the full pool.

\vspace{-0.6em}
\section{Evaluation Pipeline}
\label{sec:evaluation-pipeline}
\vspace{-0.6em}
The previous section describes how expert submissions are collected and converted into verified benchmark instances. This section specifies what happens once a task exists: how it is executed, how the agent interacts with the environment, and how the outcome is scored. The pipeline is organized around an \textbf{uncoupling} of three components (the \emph{task specification}, the \emph{agent}, and the \emph{environment}) so that they can all be easily interchanged.

\noindent \textbf{Terminology.} Throughout the paper we use \emph{task workflow} for an end-to-end professional procedure and \emph{task instance} for one runnable case of a task workflow (one concrete (input, output) pair, sharing the same \texttt{evaluate()} but differing in input and output data). The rare appearance single ``task'' refers to the runnable instance level.

\vspace{-0.6em}
\subsection{Pipeline Architecture}
\label{sec:pipeline-architecture}
\vspace{-0.6em}
\begin{wrapfigure}{r}{0.6\textwidth}
    \vspace{-0.8em}
    \centering
    \includegraphics[width=\linewidth]{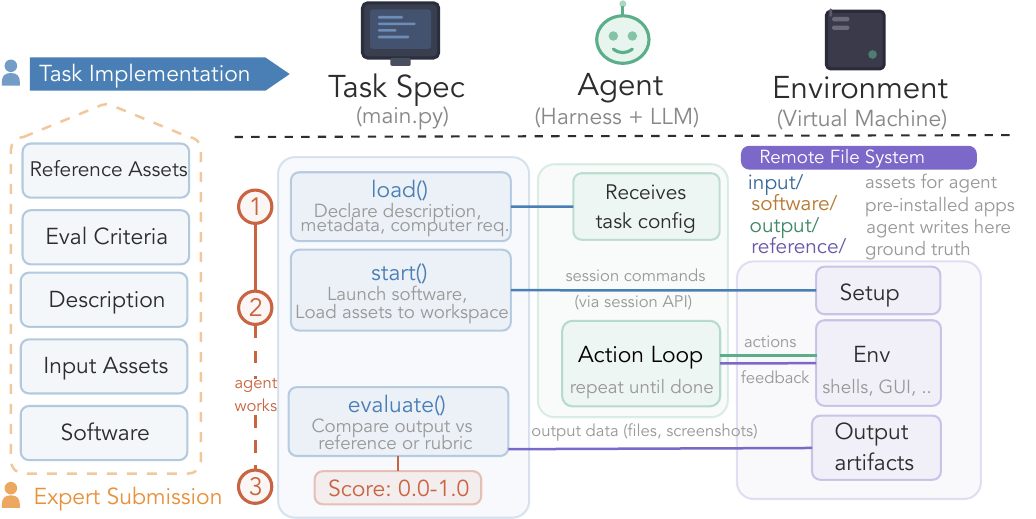}
    \caption{\textbf{Evaluation pipeline architecture.} Each benchmark instance is defined by a Task Specification (\texttt{main.py}) that orchestrates a three-phase lifecycle (\texttt{load()}, \texttt{start()}, \texttt{evaluate()}) over a remote virtual-machine environment. The agent (harness + model) receives only the task description and metadata, interacts with the environment through an action loop, and produces output artifacts that the specification scores against references or rubrics.}
    \label{fig:task-spec}
    \vspace{-0.8em}
\end{wrapfigure}

Figure~\ref{fig:task-spec} illustrates the end-to-end evaluation pipeline. A benchmark instance is realized through three decoupled components that interact via well-defined interfaces. The \textit{task specification} is an executable \texttt{main.py} encoding the five elements supplied during construction (description, input assets, target software, reference assets, evaluation criteria) and exposing three lifecycle functions: \texttt{load()} declares the task and compute requirements, \texttt{start()} provisions the VM into a deterministic starting state, and \texttt{evaluate()} scores the agent's output in $[0, 1]$. The \textit{agent} (a harness orchestrating a foundation model) receives only the task description and metadata, then runs an action loop over screenshots, shell output, mouse and keyboard, file edits, and API calls until it terminates. The \textit{environment} is a remote virtual machine with a canonical four-directory layout: \texttt{input/} (read-only assets), \texttt{software/} (pre-installed applications), \texttt{output/} (the agent's sole writable target), and \texttt{reference/} (ground-truth artifacts hidden from the agent and used only for scoring). This decoupling lets any agent that conforms to the action interface be evaluated on any task, and a single specification be deployed across cloud VMs or local containers without modification. Per-component interfaces and the directory contract are detailed in Appendix~\ref{app:pipeline-architecture}, with the lifecycle protocol in Appendix~\ref{app:task-spec-protocol}.

\vspace{-0.4em}
\subsection{Agent Architecture: From CLI/GUI-agents to Generalist CUA}
\label{sec:agent-architecture}
\vspace{-0.4em}
The tasks in ALE require agents that can read GUI screens, type in dialog boxes, execute shell commands, write and debug code, invoke APIs, and manage long-running sessions, often within a single task workflow. No single existing agent family covers this surface natively, so the benchmark targets a broader agent class that we make explicit here, together with how the prevailing harness architecture is extended to reach it.

\begin{figure}[t]
    \centering
    \begin{minipage}[b]{0.42\textwidth}
        \includegraphics[width=\textwidth]{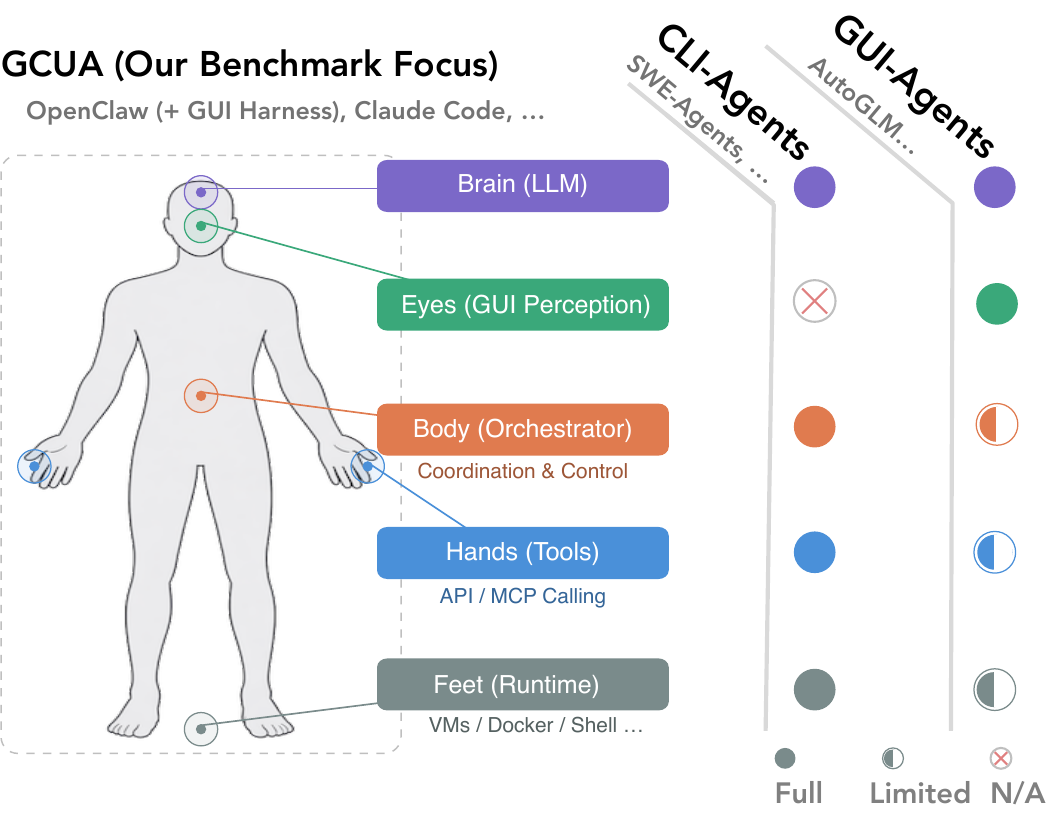}
        \caption{\textbf{Agent capability taxonomy.} Five functional layers define an agent's operational surface. Generalist CUA-agents (GCUA) possess full capability across all layers; CLI-agents lack visual perception (Eyes); GUI-agents have limited orchestration, tool use, and runtime access (Body, Hands, Feet).}
        \label{fig:agent-body}
    \end{minipage}
    \hfill
    \begin{minipage}[b]{0.55\textwidth}
        \includegraphics[width=\textwidth]{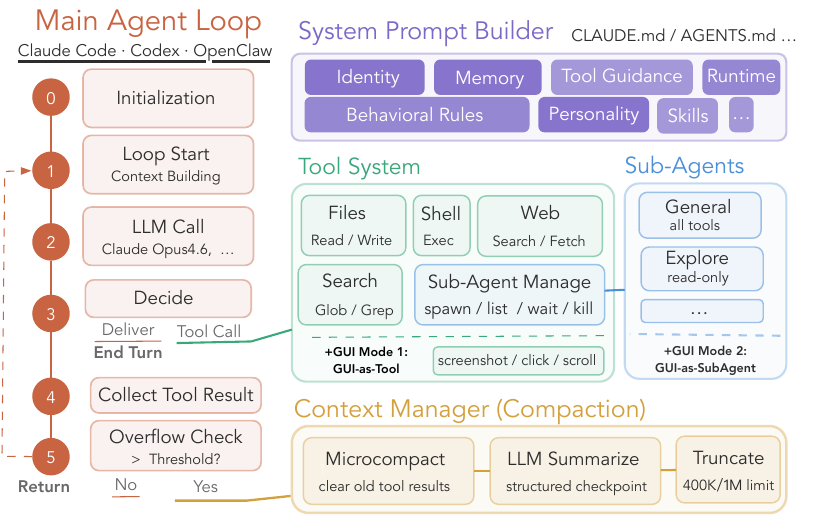}
        \caption{\textbf{Typical GCUA harness architecture.} The main agent loop (left) cycles through context building, LLM inference, action decision, tool execution, and overflow management. The system prompt builder, tool system (including GUI harness via MCP), sub-agents, and context compaction manager are shared across mainstream harness implementations.}
        \label{fig:agent-harness}
    \end{minipage}
    \vspace{-1.2em}
\end{figure}

We decompose an agent's operational capabilities into five functional layers (Figure~\ref{fig:agent-body}): \textbf{Brain} (LLM reasoning and planning), \textbf{Eyes} (GUI perception via screenshots), \textbf{Body} (orchestration and control flow), \textbf{Hands} (structured tool invocation), and \textbf{Feet} (the runtime substrate on which actions take effect). This decomposition exposes a clean split among existing families. Traditional \textbf{CLI-agents} such as SWE-agent~\citep{yang2024swe} and ForgeCode~\citep{tailcall2024forgecode} have full Brain, Body, Hands, and Feet but lack Eyes by construction; framework-style agents such as OpenClaw are not strictly CLI-only, yet they ship without a native GUI module. \textbf{GUI-agents} built on vision-language action models cover Brain and Eyes but expose only shallow Body, narrow Hands (mostly mouse and keyboard), and restricted Feet, leaving them unable to write code, manage files, or sustain long workflows. ALE's task workflows demand the union of both surfaces, so the agent class the benchmark evaluates is the \textbf{Generalist CUA-agent} (GCUA): an agent with full capability across all five layers. \textbf{We adopt the ``Generalist'' qualifier deliberately, because the industry often equates CUA-agents with GUI-agents; this conflation is incomplete.} 

The harness layer that mediates model and environment has converged toward a structure rich enough to support GCUA. Early agents revolved around a thin reasoning loop in the style of ReAct~\citep{yao2023react} (interleaved \textit{Think} and \textit{Action} steps); contemporary harnesses such as Claude Code~\citep{anthropic2025claudecode}, Codex~\citep{openai2025codex}, and OpenClaw share a richer macro-architecture that we treat as representative of the modern agent-harness layer (Figure~\ref{fig:agent-harness}): a main agent loop, a modular system prompt builder, a unified tool system, sub-agent dispatch, and a context compaction manager for long-horizon runs. Because these harnesses are CLI-native, lifting them to GCUA reduces to adding GUI capability. We use two modes: \textbf{GUI-as-Tool} exposes GUI operations as ordinary tools in the main loop, while \textbf{GUI-as-SubAgent} delegates GUI interaction to a specialized vision-language sub-agent. We currently reserve GUI-as-SubAgent for models without native vision input, such as DeepSeek V4~\citep{deepseek2026v4pro}. Our primary benchmark evaluation uses GUI-as-Tool to measure integrated visual reasoning and action over the full task. Component-level harness internals are deferred to Appendix~\ref{app:agent-harness}.

\vspace{-0.6em}
\subsection{Evaluation Modes}
\label{sec:evaluation-modes}
\vspace{-0.4em}
The deliverables that \texttt{evaluate()} must score are highly heterogeneous (CAM toolpaths, financial workbooks, 3D meshes, game world states, rendered screenshots, structured filings, free-text reports). Rather than impose a single scoring metric, ALE composes every task's evaluation along two orthogonal axes. \textbf{(i) Comparison form}: task authors pick from a small palette of artifact modes: exact / hashed values, structured numeric or tabular fields with manifest-driven tolerances, geometric surface or point-cloud distances, visual appearance (vision-LLM judge), behavioral world state under a fixed input trajectory, and free-text rubric scoring. \textbf{(ii) Composition}: the per-artifact signals are combined either by a \emph{gate-and-score} pattern or by averaging a binary checklist or per-file scores. Gate-and-score is the most common pattern: a binary precondition (e.g., ``no toolpath collision,'' ``file parses without error'') must pass before a continuous quality metric (surface deviation, dimensional accuracy, etc.) is evaluated; failure on the gate forces the task score to $0$ regardless of partial progress on other criteria. The full mode taxonomy with per-task-workflow assignments, helper APIs, and worked examples are given in Appendix~\ref{app:evaluation-modes}.

ALE \textbf{deliberately avoids LLM-as-judge} wherever a deterministic alternative exists; a task workflow whose only proposed scoring path is ``ask a model whether the result looks correct'' is rejected at QC and re-engineered to expose a checkable artifact. The minority of task workflows that genuinely require an LLM judge (video clip, game screenshot, rendered scene, etc) are scored not by general-purpose holistic prompts but by narrow, evidence-anchored yes/no probes whose answers code aggregates into the score. Each task workflow exposes a list of task instances (e.g., the 18 workpiece instances in \texttt{manufacturing/gcode}) that share a single \texttt{evaluate()} but differ in input and reference.

\begin{table}[t]
\caption{\textbf{Main results on ALE.}
Each difficulty level reports the full-pass rate~(\textbf{Pass}, \%),
the mean score~(\textbf{Score}, \%),
total API cost~(\textbf{\faIcon[regular]{money-bill-alt}}),
total wall-clock time~(\textbf{\faIcon[regular]{clock}}),
and total token use~(\textbf{Tok.}).
The final \textbf{Overall Pass Rate} column reports the full-pass rate over all distinct tasks in the three difficulty levels.
``--''~cost data not available.
Superscript~$\pm$ values denote score standard deviations estimated from three independent runs of the same task instance; due to compute budget constraints, only a subset of configurations include repeated runs.
$^\dagger$Model uses an additional visual sub-agent for visual perception.
The lower panel reports \textbf{ALE-CLI}, the Linux-only subset, comparing CLI agents alongside GCUA references~($^*$).}
\label{tab:main-results}
\centering
{\renewcommand\baselinestretch{1.25}\selectfont
\setlength{\dashlinedash}{0.7pt}%
\setlength{\dashlinegap}{1.2pt}%
\setlength{\arrayrulewidth}{0.2pt}%
\resizebox{\textwidth}{!}{
\begin{tabular}{@{\hskip 4pt}l@{\hskip 6pt}ccccc@{\hskip 7pt}:ccccc@{\hskip 7pt}:ccccc@{\hskip 6pt}:c@{\hskip 4pt}}
\toprule
& \multicolumn{5}{c}{{\large\textsf{\textbf{Near-Term}}} (67 tasks)}
& \multicolumn{5}{c}{{\large\textsf{\textbf{Full-Spectrum}}} (55 tasks)}
& \multicolumn{5}{c}{\textcolor{lastexamred}{{\large\textsf{\textbf{Last-Exam}}}} (38 tasks)}
& \textbf{Overall} \\
& Pass (\%) & Score & \faIcon[regular]{money-bill-alt} & \faIcon[regular]{clock} & Tok.
& Pass (\%) & Score & \faIcon[regular]{money-bill-alt} & \faIcon[regular]{clock} & Tok.
& Pass (\%) & Score & \faIcon[regular]{money-bill-alt} & \faIcon[regular]{clock} & Tok.
& \textbf{Pass Rate}
\\
\midrule
\rowcolor{agentrow}
\multicolumn{17}{l}{\textbf{Mainstream Agent Harnesses (paired LLM + GUI-as-Tool)}} \\
Codex~\citep{openai2025codex} (GPT-5.5~\citep{openai2026gpt55}) & \hscb{38.1} & \hsc{64.7}$^{\pm2}$ & \$243 & 154h & 550M & \hsc{22.7} & \hsc{36.0}$^{\pm3}$ & \$157 & 125h & 389M & \hsc{0.0} & \hsc{11.2}$^{\pm2}$ & \$170 & 111h & 754M & \hscb{24.0} \\
ALE-Claw (GPT-5.5~\citep{openai2026gpt55}) & \hsc{32.8} & \hscb{67.4} & \$148 & 20h & 168M & \hscb{23.6} & \hscb{41.1} & \$71 & 14h & 56M & \hscb{2.6} & \hscb{12.8} & \$107 & 18h & 127M & \hsc{23.0} \\
Claude Code~\citep{anthropic2025claudecode} (Fable 5~\citep{anthropic2026fable5}) & \hsc{34.3} & \hsc{63.4} & \$953 & 76h & 348M & \hsc{20.9} & \hsc{34.1}$^{\pm1}$ & \$609 & 67h & 333M & \hsc{0.0} & \hsc{5.2} & \$840 & 70h & 236M & \hsc{22.0} \\
Cursor~\citep{cursor2025cli} (GPT-5.5~\citep{openai2026gpt55}) & \hsc{32.1} & \hsc{60.8}$^{\pm1}$ & \$69 & 49h & 58M & \hsc{20.0} & \hsc{32.7} & \$50 & 21h & 37M & \hscb{2.6} & \hsc{10.7} & \$61 & 21h & 64M & \hsc{20.7} \\
Cursor~\citep{cursor2025cli} (Composer 2.5~\citep{cursor2026composer25}) & \hsc{34.3} & \hsc{61.1} & \$23 & 158h & 95M & \hsc{18.2} & \hsc{30.8} & \$30 & 40h & 131M & \hsc{0.0} & \hsc{8.8} & \$34 & 62h & 152M & \hsc{20.4} \\
Cursor~\citep{cursor2025cli} (Opus 4.7~\citep{anthropic2026opus47}) & \hsc{29.9} & \hsc{61.2} & \$558 & 38h & 112M & \hsc{20.0} & \hsc{37.6} & \$866 & 19h & 225M & \hscb{2.6} & \hsc{11.4} & \$588 & 21h & 122M & \hsc{20.4} \\
Droid~\citep{docker2026droid} (GPT-5.5~\citep{openai2026gpt55}) & \hsc{29.9} & \hsc{58.2} & \$106 & 32h & 99M & \hsc{16.4} & \hsc{33.4} & \$71 & 22h & 60M & \hscb{2.6} & \hsc{11.3} & \$74 & 38h & 92M & \hsc{19.1} \\
ALE-Claw (Opus 4.7~\citep{anthropic2026opus47}) & \hsc{28.4} & \hsc{60.5} & \$312 & 26h & 361M & \hsc{18.2} & \hsc{36.6} & \$258 & 19h & 302M & \hsc{0.0} & \hsc{7.9} & \$594 & 46h & 710M & \hsc{18.4} \\
Gemini CLI~\citep{google2025geminicli} (Gemini 3.1 Pro~\citep{google2026gemini31pro}) & \hsc{26.9} & \hsc{53.5} & \$342 & 113h & 240M & \hsc{12.7} & \hsc{26.4} & \$962 & 94h & 483M & \hsc{0.0} & \hsc{0.9} & \$733 & 74h & 498M & \hsc{15.8} \\
Claude Code~\citep{anthropic2025claudecode} (Opus 4.8~\citep{anthropic2026opus48}) & \hsc{26.9} & \hsc{60.2} & \$806 & 289h & 192M & \hsc{10.9} & \hsc{27.5} & \$486 & 79h & 119M & \hsc{0.0} & \hsc{5.2} & \$602 & 99h & 158M & \hsc{15.8} \\
Claude Code~\citep{anthropic2025claudecode} (Opus 4.7~\citep{anthropic2026opus47}) & \hsc{20.9} & \hsc{54.3} & \$496 & 20h & 121M & \hsc{12.7} & \hsc{29.1} & \$747 & 15h & 204M & \hsc{0.0} & \hsc{10.0} & \$685 & 20h & 171M & \hsc{13.2} \\
Droid~\citep{docker2026droid} (Opus 4.7~\citep{anthropic2026opus47}) & \hsc{27.6} & \hsc{60.2}$^{\pm1}$ & \$737 & 21h & 182M & \hsc{3.6} & \hsc{10.9} & \$167 & 5h & 40M & \hsc{0.0} & \hsc{8.0} & \$496 & 13h & 147M & \hsc{12.8} \\
ALE-Claw (GPT-5.4~\citep{openai2026gpt54}) & \hsc{20.9} & \hsc{44.3} & \$66 & 22h & 179M & \hsc{9.1} & \hsc{22.9} & \$179 & 26h & 597M & \hsc{0.0} & \hsc{8.1} & \$100 & 21h & 322M & \hsc{11.8} \\
Codex~\citep{openai2025codex} (GPT-5.4~\citep{openai2026gpt54}) & \hsc{13.4} & \hsc{22.8} & \$56 & 28h & 87M & \hsc{3.6} & \hsc{7.7} & \$42 & 10h & 67M & \hsc{0.0} & \hsc{0.0} & \$46 & 14h & 69M & \hsc{7.2} \\
Grok CLI~\citep{xai2026models} (Grok 4.3~\citep{xai2026models}) & \hsc{9.0} & \hsc{30.4}$^{\pm1}$ & \$126 & 28h & 104M & \hsc{7.3} & \hsc{17.0} & \$97 & 21h & 78M & \hsc{0.0} & \hsc{2.3} & \$73 & 18h & 60M & \hsc{6.6} \\
\midrule
\rowcolor{modelrow}
\multicolumn{17}{l}{\textbf{LLM Model Comparison (fixed OpenClaw~\citep{openclaw2026agent} + GUI-as-Tool)}} \\
GPT-5.5~\citep{openai2026gpt55} & \hscb{35.8} & \hscb{65.7} & \$218 & 48h & 224M & \hsc{18.2} & \hsc{32.1} & \$104 & 27h & 102M & \hsc{0.0} & \hscb{10.9} & \$155 & 27h & 179M & \hscb{21.1} \\
GPT-5.4~\citep{openai2026gpt54} & \hsc{33.6} & \hsc{57.8}$^{\pm1}$ & \$70 & 54h & 104M & \hscb{19.4} & \hscb{34.3}$^{\pm1}$ & \$130 & 73h & 285M & \hsc{0.0} & \hsc{7.3}$^{\pm1}$ & \$110 & 47h & 218M & \hsc{20.5} \\
Claude Opus 4.7~\citep{anthropic2026opus47} & \hsc{26.9} & \hsc{56.5} & \$493 & 57h & 218M & \hsc{10.9} & \hsc{27.5} & \$393 & 37h & 175M & \hsc{0.0} & \hsc{4.3} & \$842 & 55h & 454M & \hsc{15.1} \\
Gemini 3.1 Pro~\citep{google2026gemini31pro} & \hsc{26.1} & \hsc{48.3}$^{\pm1}$ & \$570 & 61h & 626M & \hsc{10.9} & \hsc{23.6} & \$1387 & 74h & 1854M & \hsc{0.0} & \hsc{3.1} & \$1176 & 52h & 1417M & \hsc{14.1} \\
Claude Opus 4.6~\citep{anthropic2026opus46} & \hsc{22.4} & \hsc{51.2}$^{\pm1}$ & \$242 & 58h & 238M & \hsc{13.6} & \hsc{28.5}$^{\pm1}$ & \$115 & 67h & 95M & \hsc{0.0} & \hsc{5.6}$^{\pm1}$ & \$160 & 52h & 119M & \hsc{14.1} \\
DeepSeek V4 Pro\rlap{$^\dagger$}~\citep{deepseek2026v4pro} & \hsc{19.9} & \hsc{43.8}$^{\pm2}$ & \$131 & 106h & 356M & \hsc{10.9} & \hsc{23.8}$^{\pm2}$ & \$73 & 66h & 201M & \hsc{0.0} & \hsc{2.5}$^{\pm1}$ & \$131 & 73h & 384M & \hsc{12.4} \\
Qwen3.7 Max~\citep{alibaba2026qwen37max} & \hsc{17.9} & \hsc{46.9} & \$268 & 112h & 509M & \hsc{10.9} & \hsc{27.2} & \$308 & 43h & 592M & \hsc{0.0} & \hsc{6.4} & \$196 & 44h & 382M & \hsc{11.8} \\
GLM 5.1\rlap{$^\dagger$}~\citep{zai2026glm51} & \hsc{20.1} & \hsc{45.6}$^{\pm2}$ & \$112 & 124h & 329M & \hsc{9.1} & \hsc{21.8}$^{\pm1}$ & \$237 & 116h & 580M & \hsc{0.0} & \hsc{6.2}$^{\pm1}$ & \$188 & 100h & 575M & \hsc{11.5} \\
Claude Sonnet 4.6~\citep{anthropic2026sonnet46} & \hsc{16.4} & \hsc{40.6} & \$120 & 36h & 154M & \hsc{7.3} & \hsc{19.6} & \$67 & 18h & 101M & \hsc{0.0} & \hsc{1.3} & \$48 & 13h & 79M & \hsc{9.9} \\
Kimi K2.6~\citep{moonshot2026kimi26} & \hsc{15.7} & \hsc{35.1}$^{\pm2}$ & \$47 & 119h & 208M & \hsc{6.4} & \hsc{18.2}$^{\pm1}$ & \$40 & 111h & 140M & \hsc{0.0} & \hsc{1.6} & \$36 & 85h & 121M & \hsc{9.2} \\
MIMO v2.5~\citep{mimov25} & \hsc{11.9} & \hsc{35.1}$^{\pm2}$ & \$15 & 79h & 206M & \hsc{9.1} & \hsc{20.8}$^{\pm1}$ & \$23 & 66h & 321M & \hsc{0.0} & \hsc{3.4}$^{\pm1}$ & \$13 & 62h & 221M & \hsc{8.6} \\
Qwen3.6 Plus~\citep{alibaba2026qwen36plus} & \hsc{12.7} & \hsc{35.8}$^{\pm2}$ & \$100 & 117h & 466M & \hsc{8.2} & \hsc{22.9}$^{\pm1}$ & \$120 & 89h & 480M & \hsc{0.0} & \hsc{5.0}$^{\pm1}$ & \$89 & 67h & 376M & \hsc{8.6} \\
MiniMax M2.7\rlap{$^\dagger$}~\citep{minimax2026m27} & \hsc{10.4} & \hsc{24.5}$^{\pm1}$ & \$9 & 85h & 131M & \hsc{3.6} & \hsc{8.4}$^{\pm1}$ & \$10 & 64h & 146M & \hsc{0.0} & \hsc{1.3} & \$10 & 56h & 105M & \hsc{5.9} \\
Grok 4.3~\citep{xai2026models} & \hsc{6.7} & \hsc{24.4}$^{\pm1}$ & \$38 & 78h & 143M & \hsc{3.6} & \hsc{12.9}$^{\pm2}$ & \$28 & 56h & 97M & \hsc{0.0} & \hsc{2.3}$^{\pm1}$ & \$25 & 51h & 84M & \hsc{4.3} \\
\midrule
\rowcolor{harnessrow}
\multicolumn{17}{l}{\textbf{Agent Harness Comparison (fixed GPT-5.5~\citep{openai2026gpt55} + GUI-as-Tool)}} \\
Codex~\citep{openai2025codex} & \hscb{38.1} & \hsc{64.7}$^{\pm2}$ & \$243 & 154h & 550M & \hsc{22.7} & \hsc{36.0}$^{\pm3}$ & \$157 & 125h & 389M & \hsc{0.0} & \hsc{11.2}$^{\pm2}$ & \$170 & 111h & 754M & \hscb{24.0} \\
ALE-Claw & \hsc{32.8} & \hscb{67.4} & \$148 & 20h & 168M & \hscb{23.6} & \hscb{41.1} & \$71 & 14h & 56M & \hscb{2.6} & \hscb{12.8} & \$107 & 18h & 127M & \hsc{23.0} \\
OpenClaw~\citep{openclaw2026agent} & \hsc{35.8} & \hsc{65.7} & \$218 & 48h & 224M & \hsc{18.2} & \hsc{32.1} & \$104 & 27h & 102M & \hsc{0.0} & \hsc{10.9} & \$155 & 27h & 179M & \hsc{21.1} \\
Cursor~\citep{cursor2025cli} & \hsc{32.1} & \hsc{60.8}$^{\pm1}$ & \$69 & 49h & 58M & \hsc{20.0} & \hsc{32.7} & \$50 & 21h & 37M & \hscb{2.6} & \hsc{10.7} & \$61 & 21h & 64M & \hsc{20.7} \\
Droid~\citep{docker2026droid} & \hsc{29.9} & \hsc{58.2} & \$106 & 32h & 99M & \hsc{16.4} & \hsc{33.4} & \$71 & 22h & 60M & \hscb{2.6} & \hsc{11.3} & \$74 & 38h & 92M & \hsc{19.1} \\
\midrule
\rowcolor{harnessrow}
\multicolumn{17}{l}{\textbf{Agent Harness Comparison (fixed Claude Opus 4.7~\citep{anthropic2026opus47} + GUI-as-Tool)}} \\
Cursor~\citep{cursor2025cli} & \hscb{29.9} & \hscb{61.2} & \$558 & 38h & 112M & \hscb{20.0} & \hscb{37.6} & \$866 & 19h & 225M & \hscb{2.6} & \hscb{11.4} & \$588 & 21h & 122M & \hscb{20.4} \\
ALE-Claw & \hsc{28.4} & \hsc{60.5} & \$312 & 26h & 361M & \hsc{18.2} & \hsc{36.6} & \$258 & 19h & 302M & \hsc{0.0} & \hsc{7.9} & \$594 & 46h & 710M & \hsc{18.4} \\
Claude Code~\citep{anthropic2025claudecode} & \hsc{20.9} & \hsc{54.3} & \$496 & 20h & 121M & \hsc{12.7} & \hsc{29.1} & \$747 & 15h & 204M & \hsc{0.0} & \hsc{10.0} & \$685 & 20h & 171M & \hsc{13.2} \\
\bottomrule
\end{tabular}}\par
\vspace{.35em}
\noindent{\footnotesize ALE-CLI (Linux-only subset, 105 tasks).}\par
\vspace{.12em}
\setlength{\dashlinedash}{0.7pt}%
\setlength{\dashlinegap}{1.2pt}%
\setlength{\arrayrulewidth}{0.2pt}%
\resizebox{\textwidth}{!}{
\begin{tabular}{@{\hskip 4pt}l@{\hskip 6pt}ccccc@{\hskip 7pt}:ccccc@{\hskip 7pt}:ccccc@{\hskip 6pt}:c@{\hskip 4pt}}
\toprule
& \multicolumn{5}{c}{{\large\textsf{\textbf{Near-Term}}} (47 tasks)}
& \multicolumn{5}{c}{{\large\textsf{\textbf{Full-Spectrum}}} (42 tasks)}
& \multicolumn{5}{c}{\textcolor{lastexamred}{{\large\textsf{\textbf{Last-Exam}}}} (20 tasks)}
& \textbf{Overall} \\
& Pass (\%) & Score & \faIcon[regular]{money-bill-alt} & \faIcon[regular]{clock} & Tok.
& Pass (\%) & Score & \faIcon[regular]{money-bill-alt} & \faIcon[regular]{clock} & Tok.
& Pass (\%) & Score & \faIcon[regular]{money-bill-alt} & \faIcon[regular]{clock} & Tok.
& \textbf{Pass Rate}
\\
\midrule
\rowcolor{termrow}
\multicolumn{17}{l}{\textbf{ALE-CLI: CLI Agents + GUI-as-Tool}} \\
Codex~\citep{openai2025codex} (GPT-5.5~\citep{openai2026gpt55})\rlap{$^*$} & \hscb{37.2} & \hscb{68.3}$^{\pm3}$ & \$142 & 86h & 258M & \hscb{19.0} & \hscb{32.2}$^{\pm2}$ & \$107 & 93h & 213M & \hsc{0.0} & \hscb{13.1}$^{\pm3}$ & \$101 & 62h & 394M & \hscb{23.3} \\
Claude Code~\citep{anthropic2025claudecode} (Sonnet 4.6~\citep{anthropic2026sonnet46})\rlap{$^*$} & \hsc{24.5} & \hsc{54.2}$^{\pm2}$ & \$38 & 23h & 84M & \hsc{14.3} & \hsc{28.3}$^{\pm1}$ & \$54 & 26h & 112M & \hsc{0.0} & \hsc{8.9}$^{\pm2}$ & \$63 & 38h & 118M & \hsc{16.7} \\
ForgeCode~\citep{tailcall2024forgecode} (Sonnet 4.6~\citep{anthropic2026sonnet46}) & \hsc{22.0} & \hsc{48.1}$^{\pm3}$ & \$44 & 41h & 170M & \hsc{8.7} & \hsc{17.6}$^{\pm2}$ & \$39 & 32h & 112M & \hsc{0.0} & \hsc{2.8} & \$20 & 52h & 83M & \hsc{13.3} \\
Hermes~\citep{nous2026hermes} (Sonnet 4.6~\citep{anthropic2026sonnet46}) & \hsc{21.6} & \hsc{52.4}$^{\pm3}$ & \$219 & 31h & 216M & \hsc{8.7} & \hsc{22.5}$^{\pm1}$ & \$132 & 23h & 121M & \hsc{0.0} & \hsc{4.1}$^{\pm2}$ & \$123 & 24h & 192M & \hsc{13.2} \\
Terminus~\citep{merrill2026terminalbench} (Sonnet 4.6~\citep{anthropic2026sonnet46}) & \hsc{19.1} & \hsc{50.0}$^{\pm2}$ & \$75 & 42h & 208M & \hsc{8.3} & \hsc{22.7} & \$55 & 32h & 513M & \hsc{0.0} & \hsc{4.1} & \$148 & 53h & 630M & \hsc{11.9} \\
OpenHands~\citep{wang2025openhands} (Sonnet 4.6~\citep{anthropic2026sonnet46}) & \hsc{13.8} & \hsc{31.9}$^{\pm5}$ & \$74 & 36h & 213M & \hsc{7.1} & \hsc{13.6}$^{\pm3}$ & \$21 & 26h & 191M & \hsc{0.0} & \hsc{3.7}$^{\pm2}$ & \$81 & 49h & 288M & \hsc{9.0} \\
\bottomrule
\end{tabular}}\par}
\end{table}

\section{Experiment}
\label{sec:experiment}

ALE's task instances are drawn from authentic professional workflows that experts carry out on real computer environments, routinely interleaving shell commands, GUI applications, file manipulation, and web research within a single task.
As argued in Section~\ref{sec:agent-architecture}, this operational surface requires \textbf{Generalist CUA-agents} (GCUA) with full capability across all five functional layers (Brain, Eyes, Body, Hands, Feet). We therefore evaluate all agent systems in GCUA configuration.

\subsection{Main Results}
To bring every agent into GCUA configuration, each system is extended with \textbf{GUI-as-Tool} mode: a unified CUA MCP bridge exposes 14 desktop-action tools (Table~\ref{tab:gui-tools}) as standard entries in the agent's tool system, so that a single foundation model reasons over both shell output and visual feedback within one action loop.
Table~\ref{tab:main-results} reports mean scores, pass rates, costs, and wall-clock times for agent systems and foundation models. Specifically, mean score is the mean fine-grained task score, full pass rate is the share of tasks receiving full credit. Each run is capped at five hours; the overall timeout rate is 3.8\%, ranging from ${\sim}$1\% for lightweight harnesses to 5.7\% for OpenClaw (see Appendix~\ref{app:timeout-analysis}).
Rows are grouped into five blocks: (a)~mainstream harness--backbone configurations; (b)~model sweeps with the harness fixed to OpenClaw+GUI-as-Tool; (c)~harness sweeps with the backbone fixed to GPT-5.5; (d)~harness sweeps with the backbone fixed to Claude Opus~4.7; and (e)~\textbf{ALE-CLI}, the \aleclicount{} Linux-only task instances that can be attempted by CLI-only agents (ForgeCode, Hermes, Terminus) without GUI desktop access, reported in the lower panel of Table~\ref{tab:main-results} with Codex and Claude~Code as GCUA references.

\noindent \textbf{ALE-CLI: a harder CLI-focused sub-benchmark.}
ALE-CLI is a natural comparison point to Terminal-Bench~\citep{merrill2026terminalbench}, which contains ${\sim}$100 terminal-centric tasks at a similar scale.
However, ALE-CLI tasks are substantially harder and require longer agent sessions:
Codex~\citep{openai2025codex} with GPT-5.5~\citep{openai2026gpt55}, the strongest configuration that achieves 82\% on Terminal-Bench, reaches only a 23.3\% overall pass rate on ALE-CLI (37.2\% on Near-Term, 19.0\% on Full-Spectrum, 0.0\% on \textcolor{lastexamred}{Last-Exam}).

\noindent \textbf{Three difficulty tiers.}
A single run of a frontier agent on one ALE task costs \$3--10 on average and takes tens of minutes to hours.
Evaluating the full \publictaskcount{}-task public set is therefore expensive, so ALE organizes tasks into three difficulty tiers that let the community choose a subset matching its evaluation budget and goal.
\textbf{Near-Term} (\neartermcount{} tasks) contains workflows that current frontier agents can partially complete, with top pass rates approaching 40\%; these tasks are the most cost-effective target for short-term leaderboard competition and rapid iteration.
\textbf{Full-Spectrum} (\fullspectrumcount{} tasks) covers, by design, each of ALE's \taskcount{} subdomains with at least one task instance, ensuring broad domain coverage for comprehensive evaluation.
\textcolor{lastexamred}{\textbf{Last-Exam}} (\lastexamcount{} tasks) comprises the hardest workflows, on which most agents achieve a 0\% pass rate; these tasks anchor the benchmark's long-term headroom and are best reserved for milestone evaluations rather than routine testing.

\noindent \textbf{ALE-Claw: a Self-implemented GCUA reference.}
We implement \textbf{ALE-Claw} to test whether the basic GCUA components in Section~\ref{sec:agent-architecture}, namely a single action loop, modular tools, GUI-as-Tool, and context compaction, suffice for achieving a comparable performance to frontier harnesses.
ALE-Claw is a simplified implementation derived from OpenClaw~\citep{openclaw2026agent} and scoped to isolated benchmark runs.
It omits product features such as long-term memory management and user preference customization, which are useful for interactive agents such as Claude Code~\citep{anthropic2025claudecode} but not required for single-task evaluation.
Holding the foundation model fixed, ALE-Claw has comparable performance to default OpenClaw on ALE. Appendix~\ref{app:agent-harness} documents the implementation differences.

\begin{figure}[htbp]
    \vspace{-0.8em}
    \centering
    \newcommand{\toolswatch}[2]{\textcolor[HTML]{#1}{\rule{0.7ex}{0.7ex}}\,#2}
    \newlength{\analysisfigheight}
    \setlength{\analysisfigheight}{0.25\linewidth}
    \begin{tikzpicture}[baseline=(img.south)]
      \node[inner sep=0pt,anchor=south west] (img) {\includegraphics[height=\analysisfigheight]{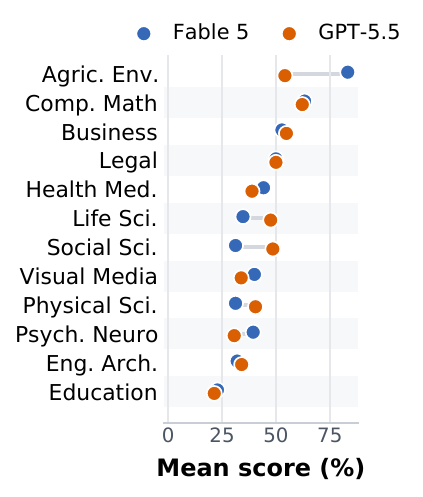}};
      \node[anchor=south west,inner sep=0.3pt,fill=white,text opacity=1,fill opacity=0.85] at ([xshift=1pt,yshift=1pt]img.south west) {\small (a)};
    \end{tikzpicture}\hspace{0.01\linewidth}%
    \begin{tikzpicture}[baseline=(img.south)]
      \node[inner sep=0pt,anchor=south west] (img) {\includegraphics[height=\analysisfigheight]{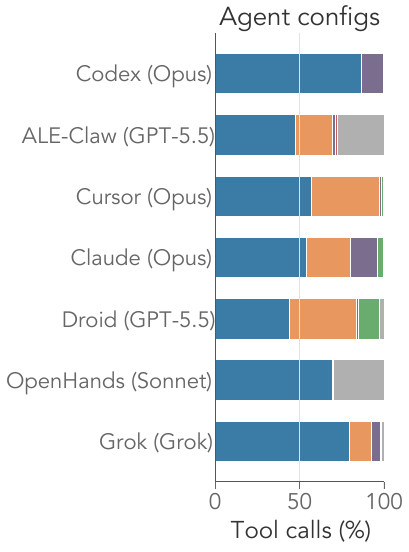}};
      \node[anchor=south west,inner sep=0.3pt,fill=white,text opacity=1,fill opacity=0.85] at ([xshift=4pt,yshift=1pt]img.south west) {\small (b)};
    \end{tikzpicture}\hspace{0.01\linewidth}%
    \begin{tikzpicture}[baseline=(img.south)]
      \node[inner sep=0pt,anchor=south west] (img) {\includegraphics[height=\analysisfigheight]{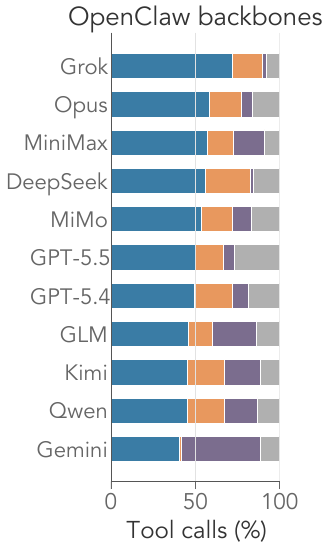}};
      \node[anchor=south west,inner sep=0.3pt,fill=white,text opacity=1,fill opacity=0.85] at ([xshift=4pt,yshift=1pt]img.south west) {\small (c)};
    \end{tikzpicture}\hspace{0.01\linewidth}%
    \begin{tikzpicture}[baseline=(img.south)]
      \node[inner sep=0pt,anchor=south west] (img) {\includegraphics[height=\analysisfigheight,trim={8pt 0pt 2pt 2pt},clip]{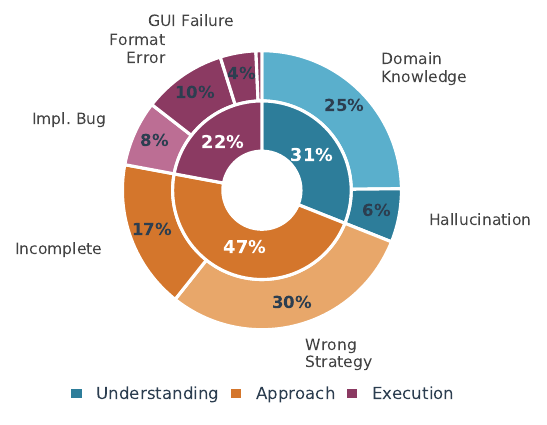}};
      \node[anchor=south west,inner sep=0.3pt,fill=white,text opacity=1,fill opacity=0.85] at ([xshift=1pt,yshift=1pt]img.south west) {\small (d)};
    \end{tikzpicture}
    \makebox[\linewidth][l]{\hspace{0.285\linewidth}{\sffamily\scriptsize
      \toolswatch{3A7CA5}{Bash}\quad
      \toolswatch{E8985E}{File}\quad
      \toolswatch{7B6D8E}{GUI}\quad
      \toolswatch{6AAB6E}{Planning}\quad
      \toolswatch{C75C5C}{Web}\quad
      \toolswatch{B0B0B0}{Other}}}
    \caption{\textbf{Experiment analysis overview.}
    (a)~Domain-level mean scores for Claude Fable~5 and GPT-5.5, each averaged over harnesses with completed runs on the selected public task set; the sparse transportation domain is omitted.
    (b)~Tool-call mix for the best available table-backed configuration per harness.
    (c)~Tool-call mix for backbone models under a fixed OpenClaw harness.
    (d)~Failure root-cause taxonomy for failed Claude Code + Opus~4.7 public-task runs.}
    \label{fig:analysis-overview}
    \label{fig:tool-usage}
    \label{fig:failure-taxonomy}
\end{figure}
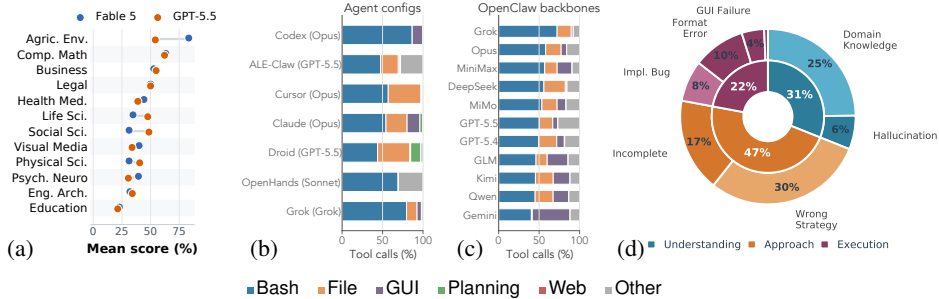

\subsection{Analysis}
\label{sec:experiment-analysis}
\noindent \textbf{Domain-level performance.}
Figure~\ref{fig:analysis-overview}(a) shows mean scores by domain for Claude Fable~5 and GPT-5.5, each averaged over harnesses.
The two frontier models exhibit similar domain profiles: computational mathematics and agriculture/environment score highest (${\sim}$55--85\%), followed by business and legal (${\sim}$50--55\%), while education remains lowest (below 25\%).
This shared ranking likely reflects both an imbalance in intrinsic model capability across domains and uneven exposure to tool-use tasks during training, where code-adjacent domains receive far more coverage than specialized professional workflows.

\noindent \textbf{Tool usage.}
The tool traces, normalized using the taxonomy in Appendix~\ref{app:tool-surface}, reveal that both harness and model shape the tool-call mix (Figure~\ref{fig:analysis-overview}(b,c)).
GUI usage remains below task demand: 34\% of public task instances designate graphical software as the primary tool, yet the GUI share stays small across most configurations, as agents execute GUI tasks through Bash/CLI substitutes.

\noindent \textbf{Failure taxonomy.}
We classified the failed tasks of Claude Code + Opus~4.7 into a two-level taxonomy (Figure~\ref{fig:analysis-overview}(d); details in Appendix~\ref{app:failure-taxonomy}).
Understanding and Approach failures together account for roughly three quarters of cases, indicating that the dominant bottleneck is domain knowledge rather than execution capability.
Lacking specialized knowledge, agents default to ad-hoc scripts instead of the intended domain software, reinforcing the GUI-underutilization pattern above.

\noindent \textbf{Additional analysis.}
Appendix~\ref{app:harness-vs-model} further decomposes the performance variation into model and harness effects, finding that the choice of foundation model accounts for roughly 3$\times$ the spread of the choice of agent harness among well-engineered systems.
Appendix~\ref{app:cost-performance} examines the cost, time, and token efficiency of each configuration (Figure~\ref{fig:cost-performance}), showing that higher resource consumption does not reliably translate to better performance.
Appendix~\ref{app:per-task-heatmap} provides per-task instance score heatmaps (Figures~\ref{fig:heatmap-nearterm}--\ref{fig:heatmap-lastexam}) that visualize every task--agent combination across the three tiers.

\section{Related Work}
\label{sec:related}

\begin{table}[t]
\centering
\caption{\textbf{Positioning of ALE against representative benchmarks.} Industry counts use the mapping in
Appendix~\ref{app:taxonomy}; ``--'' indicates that the benchmark does not declare a
domain taxonomy and is not directly comparable.}
\label{tab:related-comparison}
\renewcommand{\arraystretch}{0.95}
\setlength{\tabcolsep}{6pt}
\resizebox{\linewidth}{!}{%
\begin{tabular}{l l r c l l l l}
\toprule
\textbf{Benchmark} & \textbf{Task form} & \textbf{Size} & \textbf{Breadth} & \textbf{Domain grounding} & \textbf{Task source} & \textbf{Horizon} & \textbf{Verification} \\
\midrule
\multicolumn{8}{l}{\textit{Knowledge \& exam-style QA}} \\
MMLU~\citep{hendrycks2021measuring}        & Knowledge QA       & $\sim$16K & 26 / \taskcount{} & Academic subjects     & Mined (exams)                 & Seconds        & Auto (exact match) \\
GPQA~\citep{rein2024gpqa}                  & Knowledge QA       & $\sim$500 & 8 / \taskcount{}  & Hard sciences         & Domain experts                & Minutes        & Auto (exact match) \\
HLE~\citep{phan2026hle}                    & Knowledge QA       & $\sim$2.5K & 20 / \taskcount{} & Academic subjects     & Domain experts                & Minutes        & Auto (exact / LLM-judge) \\
\midrule
\multicolumn{8}{l}{\textit{Agentic \& computer-use benchmarks}} \\
GAIA~\citep{mialon2024gaia}                & Tool-use QA        & $\sim$500 & --                & General assistant     & Curators                      & Minutes        & Auto (exact match) \\
SWE-bench~\citep{jimenez2024swebench}      & Code patch         & $\sim$2K  & 5 / \taskcount{}  & OSS Python repos      & Mined (GitHub)                & Minutes--Hours & Auto (unit tests) \\
OSWorld~\citep{xie2024osworld}             & GUI Operation      & $\sim$400 & 5 / \taskcount{}  & Desktop apps          & Curators                      & Minutes        & Auto (state checks) \\
WebArena~\citep{zhou2024webarena}          & GUI Operation      & $\sim$800 & 4 / \taskcount{}  & Web apps (6 sites)    & Curators                      & Minutes        & Auto (state checks) \\
Terminal-Bench~2~\citep{merrill2026terminalbench} & CLI Operation & $\sim$100 & 6 / \taskcount{}  & Terminal workflows    & Curators                      & Minutes--Hours & Auto (state checks) \\
PinchBench~\citep{pinchbench2026}          & Assistant tasks    & $\sim$50  & 3 / \taskcount{}  & Productivity + coding & Curators                      & Minutes        & Auto (tests / LLM-judge) \\
\midrule
\multicolumn{8}{l}{\textit{Economically grounded / project-style benchmarks}} \\
GDPval~\citep{patwardhan2025gdpval}        & Project deliverable & $\sim$200 & 16 / \taskcount{} & Top GDP occupations  & Industry experts              & Hours--Days    & Human (expert) \\
RLI~\citep{mazeika2025rli}                 & Project deliverable & $\sim$250 & 14 / \taskcount{} & Upwork's categories & Mined (Upwork)              & Hours--Days    & Human (expert) \\
\textbf{ALE (ours)}                        & \textbf{Project deliverable} & \textbf{$\sim$1.5K} & \textbf{\taskcount{} / \taskcount{}} & \textbf{SOC 2018 + O*NET} & \textbf{Industry experts}     & \textbf{Hours--Weeks} & \textbf{Auto (deterministic scripts)} \\
\bottomrule
\end{tabular}%
}
\end{table}

Table~\ref{tab:related-comparison} positions ALE against prominent benchmarks; the ``Breadth'' column reports industry coverage by mapping each benchmark's published categories onto our \taskcount{}-industry taxonomy.
Knowledge and exam-style benchmarks such as MMLU~\citep{hendrycks2021measuring}, GPQA~\citep{rein2024gpqa}, and HLE~\citep{phan2026hle} are topically broad but test what a model \emph{knows}, not what it can \emph{do}.
Agentic benchmarks including SWE-bench~\citep{jimenez2024swebench}, OSWorld~\citep{xie2024osworld}, WebArena~\citep{zhou2024webarena}, and GAIA~\citep{mialon2024gaia} add multi-step interaction and tool use but cover only a handful of software-centric domains and typically rely on curator-authored tasks rather than real professional workflows.
The closest contemporaries, GDPval~\citep{patwardhan2025gdpval} and RLI~\citep{mazeika2025rli}, target economically grounded, project-scale evaluation but still leave large portions of the labor market untested (16 and 14 of \taskcount{} industries, respectively) and depend on expensive human grading.
ALE closes these gaps: it is the first benchmark to cover all \taskcount{} SOC/O*NET industries, draws every task from a real project completed by one of 300+ practitioners, and replaces human evaluation with deterministic, rubric-based automated verification.

\section{Conclusion}
\label{sec:conclusion}

We introduced \textbf{ALE}, a benchmark of \totaltasks{} expert-authored task workflows (\totalvariants{} task instances) across \taskcount{} digital industries, sourced from work experts have already shipped, anchored in the SOC/O*NET taxonomy, and scored through deterministic checks and structured rubrics rather than open-ended LLM judging. Frontier agents clear only a small fraction today; we release ALE as an instrument for closing the gap between benchmark success and GDP-relevant impact, where saturation would signal that agents can sustain the long-horizon, tool-intensive work professional practice actually requires.

\section{Broader Impact}
\label{sec:broader-impact}

The premise of ALE is to quantitatively measure the usefulness of agents on increasingly complex, multi-step tasks that reflect real professional work requiring planning, perception, and creative problem-solving. Although agents currently pass only a small fraction of ALE tasks, continued advances in data and training protocols in response to more sophisticated benchmarks are likely to improve agentic capabilities. As a potentially disruptive technology, agentic AI may reshape labor markets, particularly entry-level roles held by workers with bachelor's or doctoral degrees, in ways that mirror earlier disruptions to blue-collar work caused by the automation of physical labor. By providing early, measurable signals of these changes, ALE can help policymakers, educators, employers, and workers anticipate workforce transitions and create new roles that keep pace with potential job displacement as agents become more capable. Finally, by democratizing benchmark development across domain experts and making governance, access policies, and evaluation procedures explicit and transparent, benchmarks such as ALE may support more informed regulation and the responsible development of increasingly powerful agentic technologies.

\begin{ack}
We gratefully acknowledge the Tianqiao \& Chrissy Chen Institute (TCCI), Snorkel AI, and UniPat AI for their financial and credit support.
\end{ack}

{\small

}

\appendix
\addtocontents{toc}{\protect\contentsline{part}{Appendices}{}{}}

\clearpage
\begingroup
\hypersetup{linkcolor=black}
\renewcommand{\contentsname}{Appendix Table of Contents}
\setcounter{tocdepth}{2}
\localtableofcontents
\endgroup
\newpage

\section{Authors}
\label{app:authors}
Each contributor below is listed with their full affiliation (superscript), keyed to the numbered institution list at the end of this section.

\subsection{Contributors \& Affiliations}

\noindent\textbf{Organization \& Execution Team.}\quad Yiyou Sun\textsuperscript{1}, Xinyang Han\textsuperscript{1}, Weichen Zhang\textsuperscript{1}, Yuanbo Pang\textsuperscript{1}, Tianyu Wang\textsuperscript{3}, Yuhan Cao\textsuperscript{1}, Yixiao Huang\textsuperscript{1}, Xiang Liu\textsuperscript{2}, Zonglin Di\textsuperscript{11,1}, Chris Duroiu\textsuperscript{1}, Haoyun Zhang\textsuperscript{1}, Jeffrey Lin\textsuperscript{1}, Weishu Zhang\textsuperscript{1}, Tyler Zeng\textsuperscript{1}, Ying Yan\textsuperscript{3}, Bo Liu\textsuperscript{5}, Hanson Wen\textsuperscript{1}, Mingyang Xu\textsuperscript{6}, Xiaoyuan Liu\textsuperscript{1}, Zimeng Chen\textsuperscript{1}, Zhaowei Zhang\textsuperscript{4}, Weiyan Shi\textsuperscript{7}, Amanda Dsouza\textsuperscript{8}, Vincent Sunn Chen\textsuperscript{8}, Jingxu Xie\textsuperscript{1}, Dewen Lei\textsuperscript{9}, Hanxuan Zhang\textsuperscript{10}, Yinhe Wang\textsuperscript{110}, Xinran Zhang\textsuperscript{62}, Chao Zhou\textsuperscript{3}, Sheldon Huang\textsuperscript{109}, Dawn Song\textsuperscript{1}\par
% ...
\vspace{4pt}

\medskip
\noindent\textbf{Affiliations}
\vspace{4pt}

{\footnotesize\raggedright
\begin{multicols}{2}
\noindent
1.~University of California, Berkeley \\
2.~National University of Singapore \\
3.~Independent Contributor \\
4.~Individual Researcher \\
5.~Stanford University \\
6.~University of Michigan \\
7.~Northeastern University \\
8.~Snorkel AI \\
9.~University of Waterloo \\
10.~Databricks \\
11.~University of California, Santa Cruz \\
12.~SciLifeLab \\
13.~HubSpot \\
14.~nTop \\
15.~California Institute of Technology \\
16.~University of Malta \\
17.~New York University \\
18.~University of California, San Francisco \\
19.~Brown University \\
20.~OpenAGI Research Foundation \\
21.~Johns Hopkins University \\
22.~Lehigh University \\
23.~University of California, Riverside \\
24.~University of Wisconsin-Madison \\
25.~Leibniz Institute for Science and Mathematics Education \\
26.~University of Illinois Urbana-Champaign \\
27.~Singapore University of Technology and Design \\
28.~University of Texas Southwestern Medical Center \\
29.~University of Chicago \\
30.~KAIST \\
31.~University of California, Los Angeles \\
32.~Adobe \\
33.~Monash University \\
34.~IMPACT Lab at Children's National Hospital \\
35.~University of Washington \\
36.~Georgia Institute of Technology \\
37.~Independent Researcher \\
38.~North Carolina Central University \\
39.~University of Nottingham \\
40.~University of Oxford \\
41.~University of Southern California \\
42.~Meta \\
43.~University of California, Santa Barbara \\
44.~Yale University \\
45.~University of California, San Diego \\
46.~Hippocratic AI \\
47.~University of Pennsylvania \\
48.~U.S. National Institutes of Health \\
49.~Syracuse University \\
50.~Michtom School of Computer Science, Brandeis University \\
51.~University of California, Davis \\
52.~Emory University \\
53.~Duke University \\
54.~Columbia University \\
55.~Tsinghua University \\
56.~Massachusetts Institute of Technology \\
57.~McGill University \\
58.~Google \\
59.~BUPT \\
60.~UMass Chan Medical School \\
61.~ETH Zurich \\
62.~Amazon \\
63.~Purdue University \\
64.~Harvard University \\
65.~Boston University \\
66.~University of Nebraska-Lincoln \\
67.~Oracle \\
68.~Iowa State University \\
69.~AstraZeneca \\
70.~Arizona State University \\
71.~Mission San Jose High School \\
72.~X School \\
73.~Imperial College London \\
74.~Aalto University \\
75.~Cornell University \\
76.~Washington University in St. Louis \\
77.~Karlsruhe Institute of Technology \\
78.~LeverArch.ai \\
79.~BreathingCORE Limited \\
80.~Wayne State University \\
81.~University of Helsinki \\
82.~University of Massachusetts Amherst \\
83.~Texas A\&M University \\
84.~University of North Carolina at Chapel Hill \\
85.~University of Melbourne \\
86.~Michigan State University \\
87.~Jiji Information \& Communication Technology Branch \\
88.~Carnegie Mellon University \\
89.~University of California, Irvine \\
90.~University of Tokyo \\
91.~NanoFrontier \\
92.~University of Colorado Boulder \\
93.~University of Utah \\
94.~Morgan Stanley \\
95.~Photon Fund \\
96.~PIMCO \\
97.~Brix Labs \\
98.~Intellipro \\
99.~JPMorgan Chase \\
100.~Goldman Sachs \\
101.~Santa Clara University \\
102.~University of Colorado Anschutz Medical Campus \\
103.~OpenClaw Foundation \\
104.~MoE Fund \\
105.~Adaline \\
106.~Elorian AI \\
107.~MatrAIx \\
108.~BenchFlow \\
109.~University of Toronto \\
110.~Savannah College of Art and Design \\
111.~University of Auckland \\
112.~Two Minute Papers \\
113.~University of Rochester \\
114.~Mountain View High School \\
115.~Capital One \\
116.~Hithink RoyalFlush Information Network Co., Ltd. \\
117.~Shanghai Jiao Tong University \\
118.~University of Notre Dame
\end{multicols}}

\section{Benchmark Construction Details}
\label{app:benchmark-construction}

\subsection{Taxonomy Details}
\label{app:taxonomy}

\subsubsection{Taxonomy Definition}
\label{app:taxonomy-definition}

\noindent \textbf{Scope.}
ALE evaluates whether generalist computer-use agents, operating through digital interfaces, software
tools, files, and APIs, can complete valuable professional work across fields. The taxonomy centers
on workflows whose primary outputs can be produced at a computer, depend on domain expertise,
and yield artifacts that are objectively evaluable.

\noindent \textbf{Occupational backbone.}
We use SOC 2018 as the occupational backbone and O*NET to interpret occupational content through
linked task, work-activity, and tools-and-technology records~\citep{bls2018soc,peterson2001onet,onet302database}.
SOC 2018 provides a standard classification of U.S. occupations, while O*NET provides
occupation-level records describing work content, activities, tasks, and technology use. SOC 2018
describes the structure of U.S. work across 23 major groups, 98 minor groups, 459 broad occupations,
and 867 detailed occupations. To derive ALE's workflow-level taxonomy, we screen these occupation
records for in-scope digital workflows, consolidate the retained occupational evidence into
subdomains, and supplement SOC/O*NET where stable frontier workflows are absent or under-specified.

\noindent \textbf{Screening procedure.}
The initial screen applies the scope definition to the 1{,}016 entries in O*NET 30.2 Occupation
Data~\citep{onet302database} using a fixed GPT-4o mini prompt at temperature~0. Each prompt
instance includes the SOC code and title, O*NET description, task statements, work activities, and
technology examples. After O*NET variants are consolidated under shared SOC base codes, 117 unique
SOC base codes remain.

\noindent \textbf{Subdomain construction.}
The screening stage identifies occupation-level evidence, which we then organize into workflow-level
subdomains. We group SOC codes whose task statements, work activities, and tools-and-technology
records describe a common class of artifact-producing professional work. Field, methodology, and
work product are the primary dimensions used to set subdomain boundaries, with LLM-assisted
research and domain-expert review used to check borderline assignments. A SOC code can be assigned
to more than one subdomain when its work content contains separable workflows. This process yields
51 SOC-anchored subdomains.

\noindent \textbf{Frontier extension.}
SOC/O*NET anchors the taxonomy in established occupations. ALE further adds a frontier supplement
for emerging digital workflows not yet represented in SOC 2018 but already present in current
research and professional practice, as reflected in recent NIH, NSF, and field-specific technical
roadmaps~\citep{nih2021strategicplan,nsb2024sei,uschips2022,nsfchips2024,nist2023airmf}.
The supplement adds four frontier subdomains and seven extensions to SOC-anchored subdomains,
yielding \taskcount{} subdomains in total.

\noindent \textbf{Result.}
The final taxonomy contains \taskcount{} workflow-level subdomains grouped into \clustercount{}
domains.

\subsubsection{Industry Landscape Review}
\label{app:industry-landscape}

\noindent \textbf{Collection context.}
Subdomain workflow landscapes provide collection context for task workflows within each subdomain.
Each landscape summarizes the work setting, practitioner roles, digital inputs, workflow
dependencies, and output artifacts associated with a class of professional workflows. The records
are based on field references, workflow documentation, LLM-assisted research, and expert
review. They are used to organize candidate task families and to check that collected tasks specify
realistic inputs, practice-appropriate deliverables, and verifiable success criteria. This section
includes four illustrative subdomain workflow landscapes.

\subsubsection{Manufacturing \& Industrial Operations}
Manufacturing \& Industrial Operations covers workflows that turn CAD geometry, bills of materials,
process specifications, and production targets into routings, toolpaths, line layouts, control logic,
and inspection plans. A recurring handoff risk is consistency between the digital plan and the physical
process: controller dialects, fixture geometry, alarm priorities, and tolerance stacks are tracked
across CAM, controls, industrial engineering, and quality review.

Work starts from a part model, material requirements, and a production target. Process planners
assign machine classes and routings; CAM programmers generate toolpaths, post G-code, and verify
cutter motion in stock simulation; industrial engineers balance station times against takt; controls
engineers write PLC logic, SCADA screens, alarm priorities, and safety interlocks; quality engineers
author SPC plans, control limits, and first-article inspection procedures. The resulting artifacts
include routing plans, posted code, simulation outputs, control files, SPC plans, and inspection
packages. When a feature drifts out of specification, the inspection record can be compared with the
process history and, when needed, route the workflow back to fixture, parameter, or toolpath
revision.

\subsubsection{Biomolecular Structure \& Design}
Biomolecular Structure \& Design covers computational workflows that precede wet-lab execution.
Typical inputs include a target hypothesis, candidate sequences or ligands, structural templates,
assay constraints, and host-organism requirements. A recurring handoff risk is consistency between the
model and the biological material: protonation states, retained waters, MSA depth, codon usage, and
assembly-junction fidelity are tracked from computational design through experimental handoff.

Work starts from a protein to inhibit, a binder to design, or a pathway to express. Computational
chemists dock candidate ligands and run molecular dynamics; structural biologists query structure
predictors for relevant conformations; protein designers filter sequences through inverse-folding
and stability models; bioinformaticians build MSAs and codon-optimize for the host; synthetic
biologists draft plasmid maps and assembly junctions. The resulting artifacts include ranked ligand
poses, structure models, sequence designs, codon-optimized constructs, plasmid maps, and design
registry entries. If a binder fails to bind, a predicted pose is contradicted, or a pathway misses
titer, the recorded assumptions may help identify the in-silico layer to revisit.

\subsubsection{3D, Animation \& Interactive Media}
3D, Animation \& Interactive Media covers workflows that convert scripts, storyboards, look briefs,
and gameplay requirements into rendered frames or runtime states. Assets move through geometry,
rigging, material, animation, lighting, compositing, and engine stages. A recurring handoff risk is
that scale, coordinate frames, import settings, color spaces, and timing metadata are preserved as
authored assets move across tools.

Work starts from a script, storyboard, and look brief. Concept artists establish the visual language;
modelers build geometry within scale, topology, and polygon-budget constraints; look-development
artists author materials and textures; riggers wire skeletons and deformers; animators key motion
against the rig and camera; lighting technical directors stage scenes; FX artists simulate particles,
fluids, and destruction; compositors layer rendered elements, while runtime engineers and technical
artists integrate assets into gameplay logic, shaders, naming conventions, and pipeline layouts. The
resulting artifacts include models, rigs, texture sets, material graphs, animation clips, lighting
setups, compositing files, and engine scenes. When a shot or build returns with incorrect scale,
color, or timing, the asset graph can help locate the layer that introduced the drift.

\subsubsection{Robotics \& Autonomous Systems}
Robotics \& Autonomous Systems covers workflows that translate task specifications and environment
models into robot descriptions, perception calibration, controllers, planners, safety logic, and
simulation assets. A recurring handoff risk is the sim-to-real gap: kinematic descriptions, sensor
transforms, controller gains, planner parameters, and simulator assumptions are tracked from digital
twin to hardware.

Work starts from a task specification, such as picking a part, reaching a target site, or navigating
a constrained route. Mechanical engineers fix the URDF and joint limits; perception engineers
calibrate sensors, validate transforms, and test perception against expected noise; controls
engineers tune gains and design safety controllers; motion planners and trajectory optimizers compose
plans under kinematic and dynamic constraints; behavior engineers connect state machines, behavior
trees, or learned policies to failure modes; simulation engineers maintain the digital twin; safety
engineers map the system to hazard analyses. The resulting artifacts include URDFs, calibration
files, controller parameters, planner configurations, behavior specifications, simulation scenarios,
and hardware-in-the-loop test records. When hardware behavior diverges from simulation, the recorded
mismatch may inform updates to the simulator, controller, or policy before the system returns to the
validation bench.

\begin{figure}[t]
    \centering
    \includegraphics[width=\linewidth]{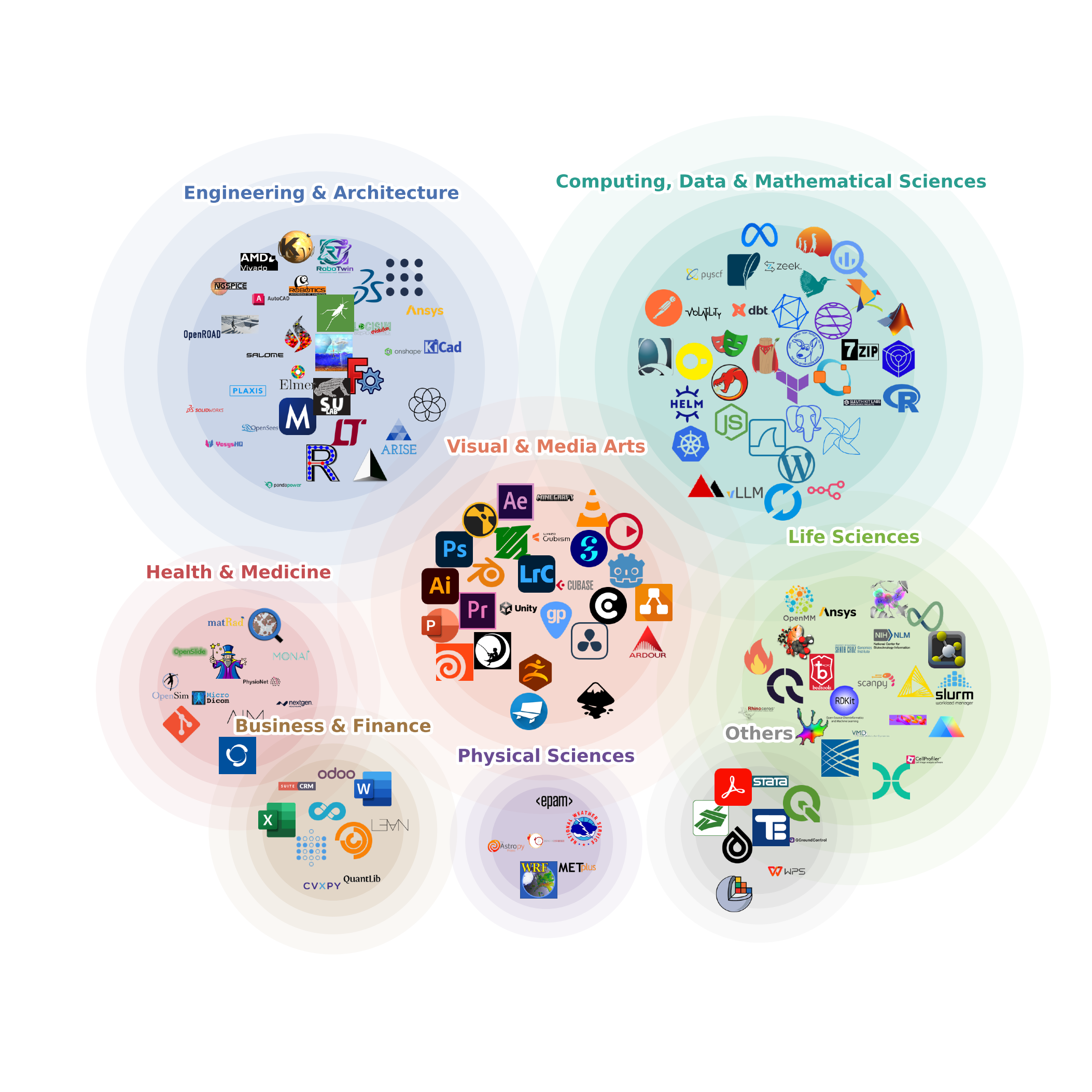}
    \caption{\textbf{Software ecosystem covered by ALE tasks.} Each icon is a distinct application or toolchain that appears in at least one task workflow, positioned within its primary ALE domain. Overlap regions hold tools that span multiple domains (e.g., creative-suite applications shared between Visual~\&~Media~Arts and Engineering). The figure is qualitative; quantitative per-subdomain instance counts appear in Figure~\ref{fig:subdomain-distribution}.}
    \label{fig:tool-constellation}
\end{figure}

\subsection{Task Construction Pipeline Details}
\label{app:construction-pipeline}

This appendix expands on the staged construction protocol summarized in Section~\ref{sec:construction-pipeline} and depicted in Figure~\ref{fig:pipeline}. The protocol converts an expert's workflow submission into a benchmark instance through five gates: expert sourcing, task submission and editing, first-pass review, task implementation, and final QC and acceptance.

\noindent \textbf{Expert sourcing.}
The pipeline begins with targeted expert outreach. To ensure coverage across the defined taxonomies, we established an advisory committee comprising leading industry professionals and expert practitioners. This committee anchors the recruitment of domain specialists who perform complex software workflows in their daily practice.

\noindent \textbf{Task submission and editing.}
Tasks originate directly from these practitioners through a dedicated web submission portal (\url{https://agents-last-exam.org/submit/new/form}). The structural overhead for experts is kept deliberately low; they are asked to upload their own past projects that historically took them days or weeks to complete. AI-assisted tools on the portal help practitioners iteratively refine their proposals until the core components are fully specified: \textit{a natural language description, the input files, the target software and tools, the expected output deliverable, and a clear evaluation specification}.
For source-grounded tasks, public papers~\citep{prasad2017video,guo2023asynchronous}, datasets, standards, and workflow documentation may also define input assets, label sets, or reference artifacts; the task bundle retains these source records as part of its material provenance.

\noindent \textbf{First-pass review.}
Upon submission, tasks undergo a first review round that functions as a screening gate. Submissions receive feedback in the form of standard conference-style decisions: \texttt{major / minor revision}, \texttt{borderline accept}, \texttt{accept}, and \texttt{strong accept}. Any revision-based decision loops the proposal back to the expert for further editing.

\noindent \textbf{Task implementation.}
Proposed tasks must then be translated from written specifications into executable benchmark environments. During implementation, an engineering team converts the expert's intent into runnable assets, provisions the necessary software containers, and codifies the evaluation logic. This process involves a rigorous engineer review and dry-run execution. If the engineer discovers gaps in the task logic or missing dependencies, the system triggers an automatic email notification, routing the task log back to the expert to unblock development.

\noindent \textbf{Final QC and acceptance.}
Before a fully implemented task is admitted to the benchmark, it passes a final quality control gate overseen by the expert committee. This peer-review step evaluates both reproducibility and evaluation integrity. Specifically, reviewers check whether the expert's reference output is fundamentally correct, whether the evaluation bounds are properly calibrated (e.g., neither impossibly narrow nor spuriously permissive), and whether the problem context provides sufficient information to reach the final state. Issues found at this stage send the task back for final adjustments prior to acceptance.

\subsection{Industry Task Cards and Metadata}
\label{app:industry-card}

\begingroup
\setlength{\parindent}{0pt}
\setlength{\parskip}{0.24em}
\setlist[itemize]{leftmargin=1.05em,itemsep=0.16em,topsep=0.12em,parsep=0em}

\definecolor{cardborder}{HTML}{BFC4C9}
\definecolor{headerbg}{HTML}{F1F3F4}
\definecolor{summarybg}{HTML}{FFF7E6}
\definecolor{inputbg}{HTML}{EDF7F6}
\definecolor{evalbg}{HTML}{EEF6EA}
\definecolor{softwarebg}{HTML}{F6F6F8}
\definecolor{analysisbg}{HTML}{FCEDEA}
\definecolor{failbg}{HTML}{FBEDEA}
\definecolor{failred}{HTML}{A83A33}
\definecolor{partialbg}{HTML}{FFF2D5}
\definecolor{partialorange}{HTML}{855A00}
\definecolor{successbg}{HTML}{EAF5EC}
\definecolor{successgreen}{HTML}{2F6F3E}

\newcommand{\Code}[1]{\texttt{\detokenize{#1}}}
\newcommand{\Chip}[1]{\begingroup\setlength{\fboxsep}{2pt}\colorbox{headerbg}{\scriptsize #1}\endgroup}
\newcommand{\StatusBadge}[3]{\begingroup\setlength{\fboxsep}{3pt}\colorbox{#2}{\textcolor{#3}{\textbf{#1}}}\endgroup}
\newcommand{\BlockTitle}[1]{{\raggedright\textbf{\footnotesize #1}\par}\vspace{2pt}}

\newtcolorbox{ColorTaskCard}[1]{
  enhanced,
  colback=white,
  colframe=cardborder,
  boxrule=0.8pt,
  arc=0pt,
  left=7pt,
  right=7pt,
  top=6pt,
  bottom=7pt,
  before skip=0.25em,
  after skip=0.25em,
  title={#1},
  coltitle=black,
  colbacktitle=headerbg,
  fonttitle=\bfseries\large
}

\newtcolorbox{CardBlock}[3][]{
  enhanced,
  colback=#2,
  colframe=#3,
  boxrule=0.25pt,
  arc=1pt,
  left=7pt,
  right=7pt,
  top=5pt,
  bottom=5pt,
  before skip=4pt,
  after skip=4pt,
  before upper={\scriptsize\raggedright},
  #1
}

\subsubsection{Representative Task Cards}
\label{app:representative-gui-task-cards}
\small
We present representative task cards to illustrate how ALE task instances are specified, executed,
and scored. Each card summarizes the agent-facing request, input materials, expected deliverables,
evaluation rubric, observed score, and observed outcome for one executed task instance. All trajectories
shown here were produced by the Claude Code harness running Claude Opus 4.7. These cards
are intended to make the benchmark construction and evaluation protocol concrete at the instance
level, complementing the aggregate coverage and performance results reported in the main text.
\clearpage
\begin{ColorTaskCard}{Task Card 1: Injection Mold-Flow Analysis \hfill \StatusBadge{PARTIAL}{partialbg}{partialorange}}
\scriptsize
{\textbf{Task ID:} \Code{manufacturing/mold-flow}}\\[-1pt]
{\textbf{Observed score:} 0.476}\quad
\Chip{Manufacturing} \Chip{Moldex3D} \Chip{CAE}

\begin{CardBlock}{summarybg}{summarybg}
\BlockTitle{TASK DESCRIPTION}
Injection mold-flow simulation supports tooling and process decisions. The task provides a meshed Moldex3D project, a process specification, and a results template. The agent must open project 230057, apply process\_spec.json, run fill, pack, cooling, and warp analysis, and write output/results.json with solver-derived pressure, force, time, volume, and weight values.
\end{CardBlock}

\begin{CardBlock}{softwarebg}{softwarebg}
\BlockTitle{SOFTWARE / EXECUTION VIEW}
\centering
\includegraphics[width=0.985\linewidth,height=3.18in,keepaspectratio]{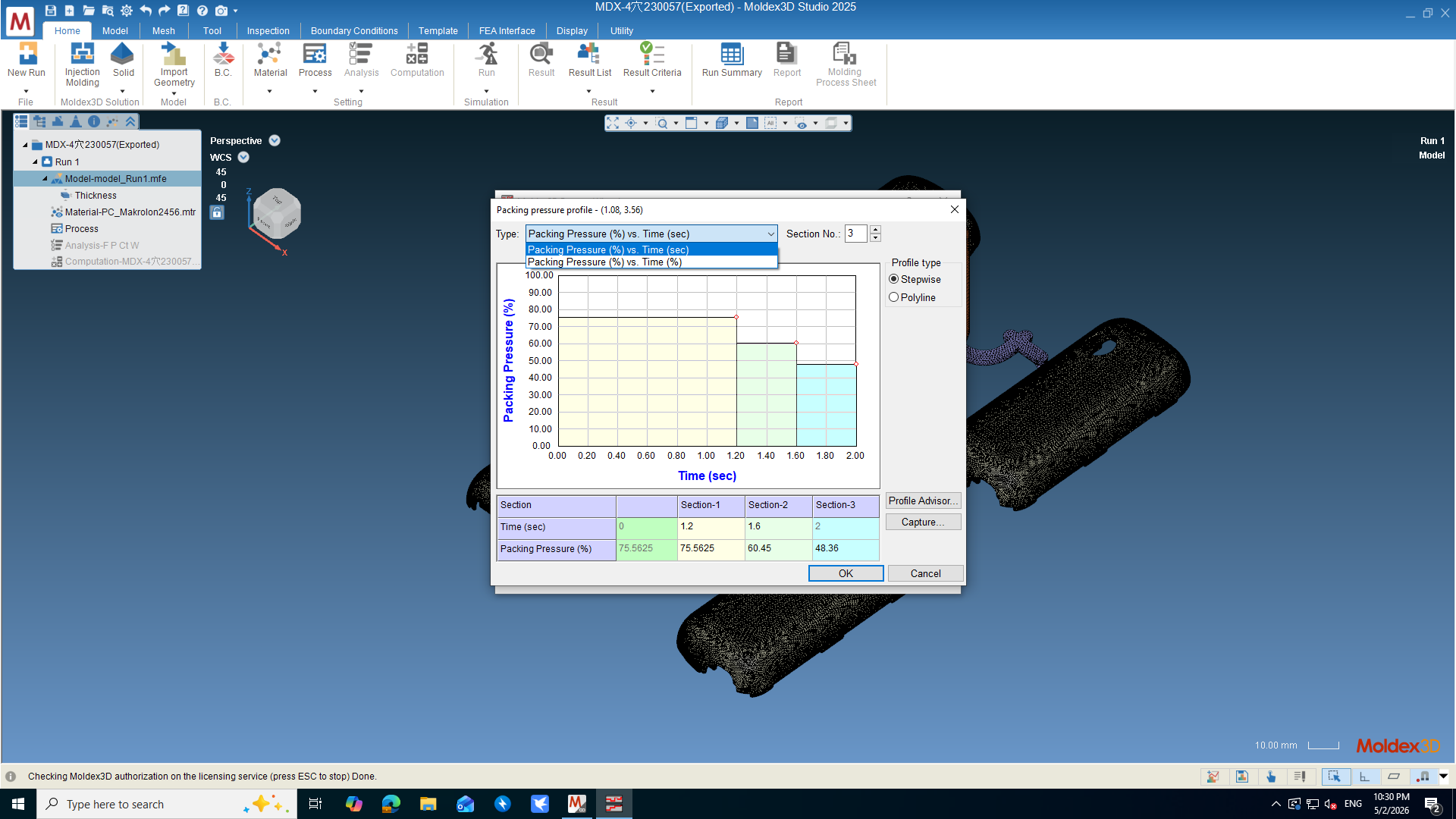}

\vspace{1pt}
{\scriptsize Moldex3D CAE workflow: the agent is editing the packing-pressure curve for a four-cavity injection-mold simulation. This is the process-setup stage; scoring depends on completing the solver run and extracting the pressure, force, cycle-time, volume, and weight metrics into results.json.}
\end{CardBlock}

\begin{minipage}[t]{0.49\linewidth}
\vspace{0pt}
\begin{CardBlock}{inputbg}{inputbg}
\BlockTitle{INPUT MATERIALS}
\begin{itemize}
\item Staged Moldex3D project with mesh, material, and four-cavity geometry.
\item process\_spec.json specifying fill, pack, cool, and warp process settings.
\item results\_template.json defining every metric required in the final JSON.
\end{itemize}
\end{CardBlock}

\begin{CardBlock}{inputbg}{inputbg}
\BlockTitle{REFERENCE OUTPUT}
\begin{itemize}
\item Saved Moldex3D project after the configured simulation run.
\item output/results.json populated with injection pressure, V/P switch pressure, clamp force, cooling/cycle times, volume, and weight fields.
\end{itemize}
\end{CardBlock}

\end{minipage}\hfill
\begin{minipage}[t]{0.49\linewidth}
\vspace{0pt}
\begin{CardBlock}{evalbg}{evalbg}
\BlockTitle{EVALUATION RUBRIC}
\begin{itemize}
\item Hard gates reject missing, malformed, or all-null results.json.
\item The main score compares each numeric field with a hidden solver reference within 1 percent relative tolerance.
\item Generated project artifacts can provide limited partial credit only when the main metrics are weak.
\end{itemize}
\end{CardBlock}

\begin{CardBlock}{analysisbg}{analysisbg}
\BlockTitle{OBSERVED OUTCOME}
\begin{itemize}
\item Score: 0.4762. The agent reached the right Moldex3D setup workflow and submitted results.json, but the score indicates only partial numeric/result credit.
\item Failure mode: the submitted metrics were not reliably extracted from completed solver outputs; some values were estimated rather than measured CAE results.
\item Interpretation: successful GUI setup was insufficient without closed-loop numerical verification: wait for result files, inspect or export solver data, then populate the JSON.
\end{itemize}
\end{CardBlock}
\end{minipage}
\end{ColorTaskCard}
\clearpage
\begin{ColorTaskCard}{Task Card 2: Orchestral Music Transcription \hfill \StatusBadge{FAIL}{failbg}{failred}}
\scriptsize
{\textbf{Task ID:} \Code{audio/music_transcription}}\\[-1pt]
{\textbf{Observed score:} 0.000}\quad
\Chip{Audio production} \Chip{Notation} \Chip{Dorico}

\begin{CardBlock}{summarybg}{summarybg}
\BlockTitle{TASK DESCRIPTION}
Orchestral transcription converts an audio recording into a printable score and multitrack MIDI. The task provides an audio brief, a Dorico environment, a named piece, tempo, instrumentation contract, and exact output-file requirements. The agent must follow the Dorico Prelude / Akinola / tempo 140 / 27-instrument specification and deliver transcription.pdf, transcription.mid, and overview.png.
\end{CardBlock}

\begin{CardBlock}{softwarebg}{softwarebg}
\BlockTitle{SOFTWARE / EXECUTION VIEW}
\centering
\includegraphics[width=0.985\linewidth,height=3.45in,keepaspectratio]{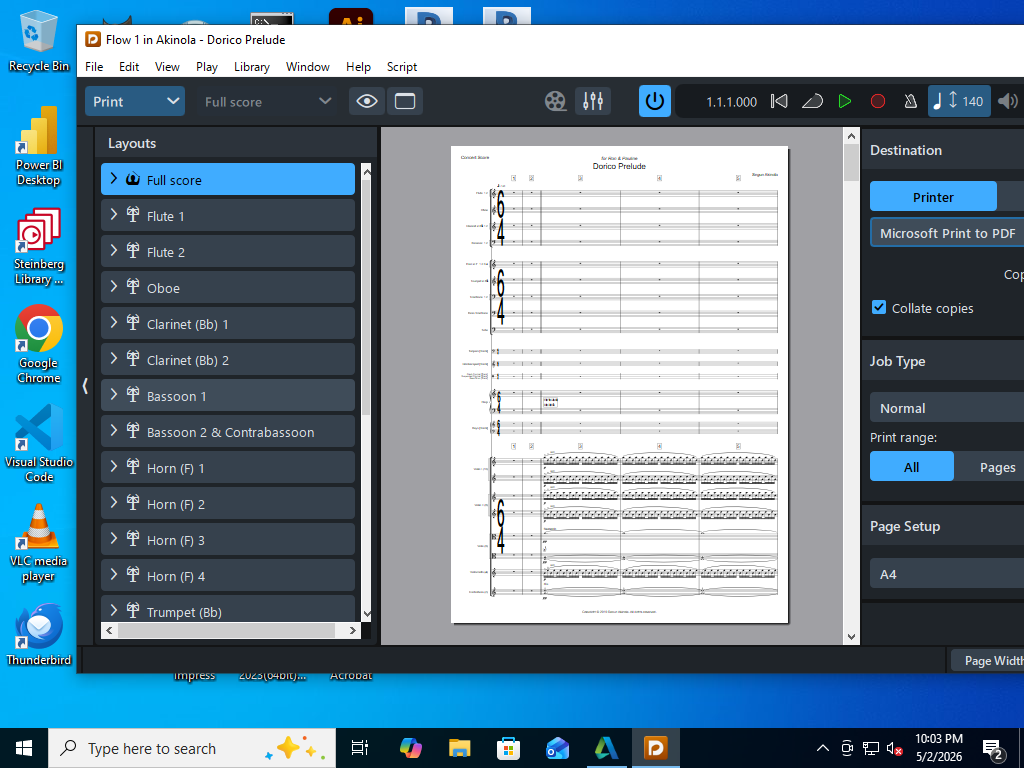}

\vspace{1pt}
{\scriptsize Music-engraving workflow: Dorico is open on an orchestral score in print/export mode. The task required converting the audio brief into a readable full score and MIDI, then exporting the PDF, MIDI, and overview screenshot to the exact output paths.}
\end{CardBlock}

\begin{minipage}[t]{0.49\linewidth}
\vspace{0pt}
\begin{CardBlock}{inputbg}{inputbg}
\BlockTitle{INPUT MATERIALS}
\begin{itemize}
\item Audio reference and task\_brief.json with title, composer, tempo, instruments, and GM programs.
\item Dorico environment for score engraving and export.
\item Output-path requirements for PDF, MIDI, and overview screenshot.
\end{itemize}
\end{CardBlock}

\begin{CardBlock}{inputbg}{inputbg}
\BlockTitle{REFERENCE OUTPUT}
\begin{itemize}
\item transcription.pdf showing a complete, readable full score.
\item transcription.mid with multitrack timing, pitches, dynamics, and GM Program Change assignments.
\item overview.png showing a notation UI with the produced score.
\end{itemize}
\end{CardBlock}

\end{minipage}\hfill
\begin{minipage}[t]{0.49\linewidth}
\vspace{0pt}
\begin{CardBlock}{evalbg}{evalbg}
\BlockTitle{EVALUATION RUBRIC}
\begin{itemize}
\item Missing PDF, missing MIDI, or missing/invalid notation screenshot triggers a hard zero.
\item After the gates pass, pitch and rhythm each contribute 30 percent, dynamics 20 percent, instrument assignment 10 percent, and score layout 10 percent.
\item The MIDI is compared to a reference MIDI after track pairing and quantization.
\end{itemize}
\end{CardBlock}

\begin{CardBlock}{analysisbg}{analysisbg}
\BlockTitle{OBSERVED OUTCOME}
\begin{itemize}
\item Score: 0.0. The agent found a matching Dorico project and exported transcription.mid, but the required transcription.pdf and overview.png were not confirmed.
\item Failure mode: the submission contained part of the required package, but the scoring hard gates require a complete score/MIDI/screenshot bundle before musical accuracy is evaluated.
\item Interpretation: partial tool use did not translate into reliable multi-artifact delivery for a creative-production task.
\end{itemize}
\end{CardBlock}
\end{minipage}
\end{ColorTaskCard}
\clearpage
\begin{ColorTaskCard}{Task Card 3: Reference-Based Chroma Key Compositing \hfill \StatusBadge{FAIL}{failbg}{failred}}
\scriptsize
{\textbf{Task ID:} \Code{media/chroma_key_from_reference}}\\[-1pt]
{\textbf{Observed score:} 0.000}\quad
\Chip{VFX} \Chip{DaVinci Resolve} \Chip{Compositing}

\begin{CardBlock}{summarybg}{summarybg}
\BlockTitle{TASK DESCRIPTION}
Chroma keying is a VFX finishing task: remove a green-screen background, preserve foreground edges, place the subject over the intended plate, and match a reference frame. The task provides a bird foreground clip, a reference image, and a DaVinci Resolve environment. The agent must create a Resolve project, key the bird from input.mp4, match the composition implied by input.png, and export output/output.mp4.
\end{CardBlock}

\begin{CardBlock}{softwarebg}{softwarebg}
\BlockTitle{SOFTWARE / EXECUTION VIEW}
\centering
\includegraphics[width=0.985\linewidth,height=3.45in,keepaspectratio]{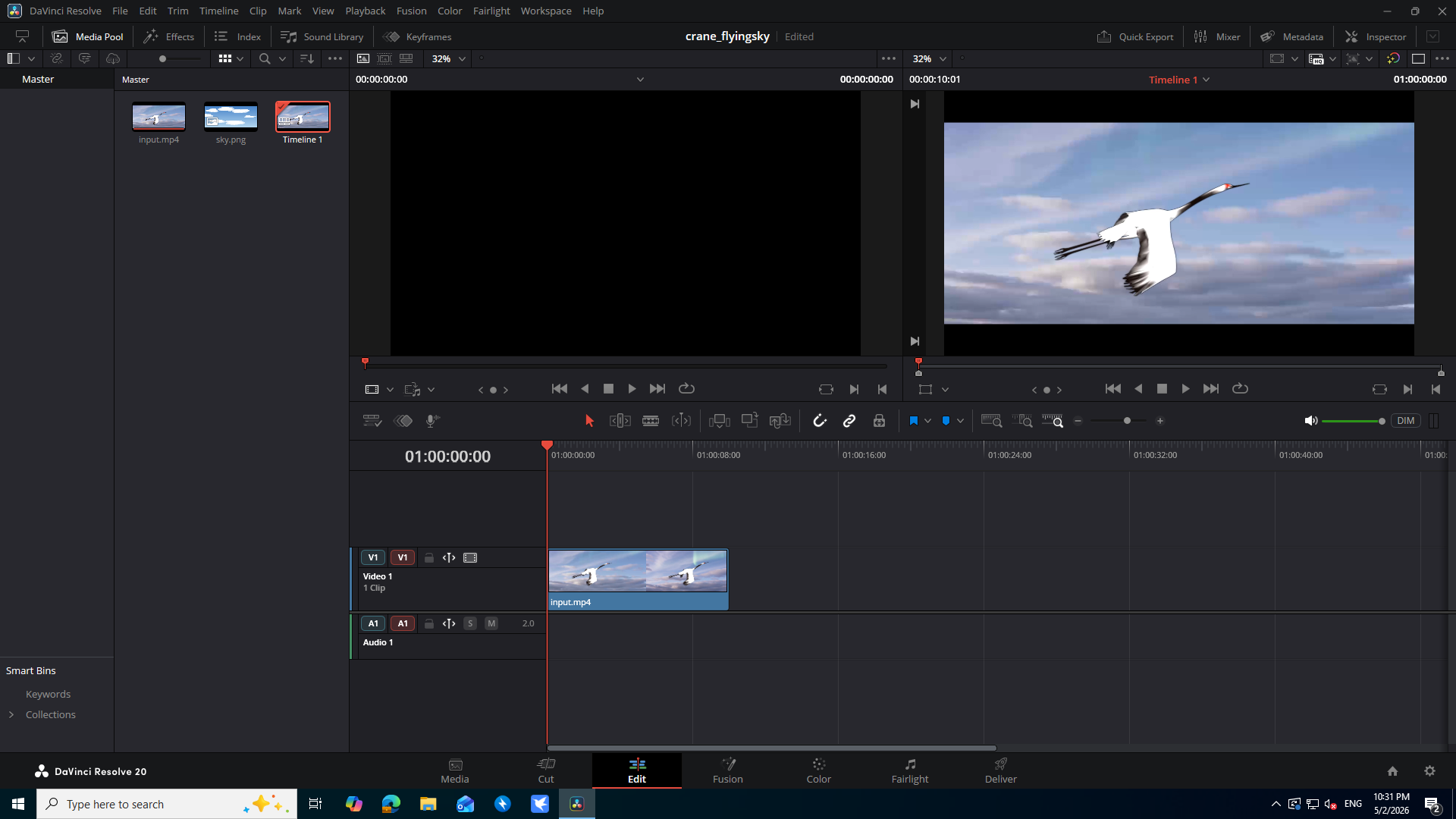}

\vspace{1pt}
{\scriptsize VFX compositing workflow: DaVinci Resolve shows a timeline and preview monitor with the bird footage. The task required identifying the green-screen foreground, keying it, compositing it over the intended sky plate, and matching the reference frame.}
\end{CardBlock}

\begin{minipage}[t]{0.49\linewidth}
\vspace{0pt}
\begin{CardBlock}{inputbg}{inputbg}
\BlockTitle{INPUT MATERIALS}
\begin{itemize}
\item input.mp4 source clip for the keying workflow.
\item input.png reference frame defining target composition and foreground placement.
\item DaVinci Resolve desktop environment and required output path.
\end{itemize}
\end{CardBlock}

\begin{CardBlock}{inputbg}{inputbg}
\BlockTitle{REFERENCE OUTPUT}
\begin{itemize}
\item output/output.mp4 containing the composited bird shot.
\item Frames with foreground preserved, green screen removed, and contour/placement close to the reference.
\end{itemize}
\end{CardBlock}

\end{minipage}\hfill
\begin{minipage}[t]{0.49\linewidth}
\vspace{0pt}
\begin{CardBlock}{evalbg}{evalbg}
\BlockTitle{EVALUATION RUBRIC}
\begin{itemize}
\item Remote scoring samples frames from reference/breakpoint.json.
\item Foreground preservation is gated by roi\_input\_cv, and quality uses full-frame plus ROI edge IoU.
\item A local vision judge must also confirm that a real keying-related edit was made rather than a raw-source copy.
\end{itemize}
\end{CardBlock}

\begin{CardBlock}{analysisbg}{analysisbg}
\BlockTitle{OBSERVED OUTCOME}
\begin{itemize}
\item Score: 0.0. The agent used Resolve and exported an MP4, but the output did not satisfy the reference-based chroma-key transformation.
\item Failure mode: the submission treated a plausible bird-over-sky clip as the target instead of preserving the intended foreground/background relationship from the reference materials.
\item Interpretation: the error was in visual task grounding rather than software operation alone.
\end{itemize}
\end{CardBlock}
\end{minipage}
\end{ColorTaskCard}
\clearpage
\begin{ColorTaskCard}{Task Card 4: Skeletal Animation Reproduction \hfill \StatusBadge{PARTIAL}{partialbg}{partialorange}}
\scriptsize
{\textbf{Task ID:} \Code{game/skeletal_animation_reproduction}}\\[-1pt]
{\textbf{Observed score:} 0.429}\quad
\Chip{Game / 3D} \Chip{Blender} \Chip{Motion matching}

\begin{CardBlock}{summarybg}{summarybg}
\BlockTitle{TASK DESCRIPTION}
Animation production turns a static character asset into a rigged, replayable performance. The task provides a Blender environment, Singing.obj, and a reference motion video. The agent must rig the character, reproduce the body motion from reference.mp4, include 19 exact bone names, and submit both an editable final.blend and a replayable preview.mp4.
\end{CardBlock}

\begin{CardBlock}{softwarebg}{softwarebg}
\BlockTitle{SOFTWARE / EXECUTION VIEW}
\centering
\includegraphics[width=0.985\linewidth,height=3.12in,keepaspectratio]{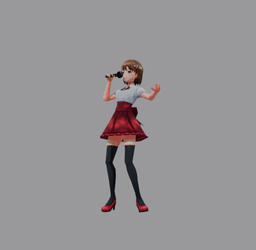}

\vspace{1pt}
{\scriptsize 3D animation output view: this render frame shows the rigged singer character from the agent's Blender submission. The task was to reproduce the body motion from a reference video, so a single valid-looking frame is only partial evidence; timing, pose range, and replay consistency determine success.}
\end{CardBlock}

\begin{minipage}[t]{0.49\linewidth}
\vspace{0pt}
\begin{CardBlock}{inputbg}{inputbg}
\BlockTitle{INPUT MATERIALS}
\begin{itemize}
\item Singing.obj static character mesh.
\item reference.mp4 body-motion target and reference/package.json bone-name requirements.
\item Blender runtime for rigging, keyframing, rendering, and replay validation.
\end{itemize}
\end{CardBlock}

\begin{CardBlock}{inputbg}{inputbg}
\BlockTitle{REFERENCE OUTPUT}
\begin{itemize}
\item final.blend containing a usable animated rig with the required bone names.
\item preview.mp4 whose body motion, timing, and pose range match the hidden clean reference.
\end{itemize}
\end{CardBlock}

\end{minipage}\hfill
\begin{minipage}[t]{0.49\linewidth}
\vspace{0pt}
\begin{CardBlock}{evalbg}{evalbg}
\BlockTitle{EVALUATION RUBRIC}
\begin{itemize}
\item Validity gates require both artifacts and a non-trivial animated rig.
\item Scoring combines video match, replay consistency from final.blend, minimal skeleton coverage, and binary vision checks.
\item Short or damped motion can lose credit even when the required files are present.
\end{itemize}
\end{CardBlock}

\begin{CardBlock}{analysisbg}{analysisbg}
\BlockTitle{OBSERVED OUTCOME}
\begin{itemize}
\item Score: 0.4285. The required Blender and preview artifacts existed and contained animation, so the run earned partial credit.
\item Failure mode: preview.mp4 was about 4 seconds while the reference was about 13.6 seconds, and the motion was damped to reduce mesh distortion.
\item Interpretation: the submission satisfied file-level validity but not the full motion-fidelity requirement.
\end{itemize}
\end{CardBlock}
\end{minipage}
\end{ColorTaskCard}
\clearpage
\begin{ColorTaskCard}{Task Card 5: MicroDicom Chest X-Ray Adjudication \hfill \StatusBadge{PARTIAL}{partialbg}{partialorange}}
\scriptsize
{\textbf{Task ID:} \Code{radiology/microdicom_nih_cxr_reader_adjudication}}\\[-1pt]
{\textbf{Observed score:} 0.333}\quad
\Chip{Radiology} \Chip{DICOM GUI} \Chip{Annotation QA}

\begin{CardBlock}{summarybg}{summarybg}
\BlockTitle{TASK DESCRIPTION}
Radiology dataset curation often requires adjudication between competing reader annotations. The task provides nine NIH CXR DICOM cases, two proposed atelectasis boxes per case, reader notes, and fixed TSV schemas. The agent must inspect each case in MicroDicom, choose the better reader annotation, and submit adjudicated\_boxes.tsv, adjudication\_log.tsv, and final\_impressions.tsv.
\end{CardBlock}

\begin{CardBlock}{softwarebg}{softwarebg}
\BlockTitle{SOFTWARE / EXECUTION VIEW}
\centering
\includegraphics[width=0.985\linewidth,height=3.45in,keepaspectratio]{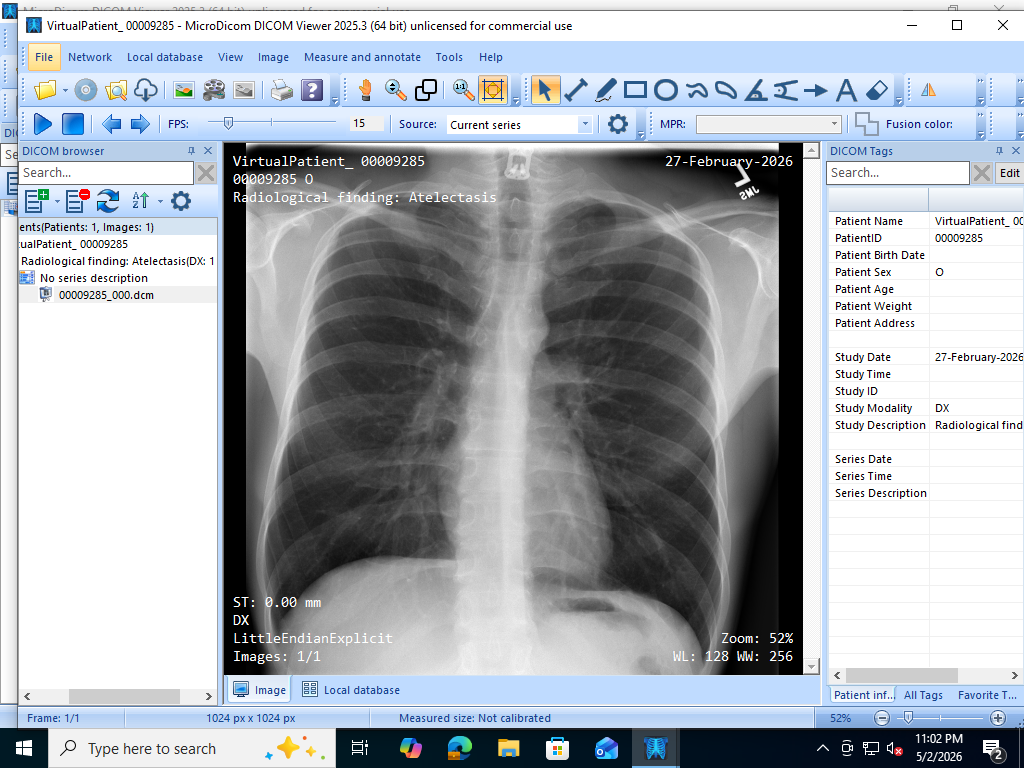}

\vspace{1pt}
{\scriptsize Radiology adjudication workflow: MicroDicom displays a chest X-ray with DICOM metadata and annotation tools. The task required reviewing each case, comparing two reader boxes for atelectasis, and writing TSV decisions; the deliverable depends on visual adjudication for each case.}
\end{CardBlock}

\begin{minipage}[t]{0.49\linewidth}
\vspace{0pt}
\begin{CardBlock}{inputbg}{inputbg}
\BlockTitle{INPUT MATERIALS}
\begin{itemize}
\item Nine DICOM chest X-ray cases with a case manifest.
\item reader\_a and reader\_b annotation TSVs plus clinical notes.
\item Rules fixing labels, allowed selected\_reader values, and output schemas.
\end{itemize}
\end{CardBlock}

\begin{CardBlock}{inputbg}{inputbg}
\BlockTitle{REFERENCE OUTPUT}
\begin{itemize}
\item adjudicated\_boxes.tsv with the final reader choice and box per case.
\item adjudication\_log.tsv and final\_impressions.tsv with the required rows and fixed labels.
\end{itemize}
\end{CardBlock}

\end{minipage}\hfill
\begin{minipage}[t]{0.49\linewidth}
\vspace{0pt}
\begin{CardBlock}{evalbg}{evalbg}
\BlockTitle{EVALUATION RUBRIC}
\begin{itemize}
\item Each TSV has hard gates for missing files, malformed headers, invalid reader values, or wrong fixed labels.
\item Log and impression files must match hidden references exactly by case\_id.
\item Box output must select the hidden-supported reader and achieve IoU at least 0.50 against the gold box.
\end{itemize}
\end{CardBlock}

\begin{CardBlock}{analysisbg}{analysisbg}
\BlockTitle{OBSERVED OUTCOME}
\begin{itemize}
\item Score: 0.3333. The agent wrote the three required TSV files covering all nine cases, but the score indicates only partial agreement with the hidden references.
\item Failure mode: detailed image review appears only for a few cases; later decisions relied on coordinates and heuristics rather than systematic visual adjudication.
\item Interpretation: file-format compliance was stronger than case-by-case visual-domain judgment.
\end{itemize}
\end{CardBlock}
\end{minipage}
\end{ColorTaskCard}
\clearpage
\endgroup

\subsubsection{Executed Task Inventory}
\label{app:complete-executed-task-inventory}

The full inventory of the 150 selected public tasks, with task name, task identifier, domain, and a brief description for each, is browsable online under the \emph{Benchmark Splits (ALE-V1, 2026/06)} tab at \url{https://agenthle.org/demo}.

\section{Evaluation Pipeline Details}
\label{app:evaluation-pipeline}

\subsection{Pipeline Architecture Details}
\label{app:pipeline-architecture}

This appendix expands on the three decoupled components summarized in Section~\ref{sec:pipeline-architecture} and depicted in Figure~\ref{fig:task-spec}: the task specification, the agent, and the environment.

\noindent \textbf{Task Specification.}
The task specification is the executable form of an expert submission. It encapsulates five elements provided during the construction pipeline (Section~\ref{sec:construction-pipeline}): a natural-language \emph{description}, \emph{input assets}, the required \emph{software}, \emph{reference assets} (ground-truth outputs), and \emph{evaluation criteria}. These are encoded in a single \texttt{main.py} file that exposes three lifecycle functions: \texttt{load()} declares the task description, metadata, and compute requirements; \texttt{start()} provisions the virtual machine into a deterministic starting state by copying input assets and launching the required software; and \texttt{evaluate()} scores the agent's output artifacts against references or rubrics, returning a normalized score in $[0, 1]$. The full protocol is detailed in Appendix~\ref{app:task-spec-protocol}.

\noindent \textbf{Agent.}
The agent is the system under evaluation, composed of a \emph{harness} (orchestration middleware) and a \emph{model} (the foundation model). Upon receiving the task configuration, which consists of a description and associated metadata, the agent enters an action loop: it observes the environment (via screenshots, shell output, or file contents), selects an action (mouse clicks, keystrokes, shell commands, file edits, or API calls), executes it, and repeats until it decides to terminate.

\noindent \textbf{Environment.}
Each task executes inside a remote virtual machine that hosts the required industrial software and exposes a standardized filesystem layout. Four directories partition the workspace: \texttt{input/} contains assets the agent reads (e.g., design files, game binaries, raw data); \texttt{software/} holds pre-installed applications and their dependencies; \texttt{output/} is the sole writable target where the agent deposits its deliverables; and \texttt{reference/} stores ground-truth artifacts used exclusively by the scoring function (the agent has no access to this directory during execution). This layout enforces a clean contract: the agent reads from \texttt{input/}, writes to \texttt{output/}, and is scored by comparison against \texttt{reference/}.

\noindent \textbf{Compute environment.}
All task instances are executed on Google Cloud Platform (GCP) virtual machines. The default configuration is \texttt{c4-standard-4} (4~vCPUs, 16\,GB RAM). Tasks that require GPU acceleration (e.g., 3D rendering, simulation) use \texttt{g2-standard-8} instances equipped with an NVIDIA L4 GPU. A small number of tasks involving heavy numerical simulation are provisioned with higher-memory or multi-core configurations as dictated by the task's compute requirements declared in \texttt{load()}. These resource assignments are determined per-task based on the software and workload involved.

\noindent \textbf{Decoupled design guarantees.}
The decoupled design yields two practical guarantees. First, any agent, regardless of its internal architecture, model backbone, or tool configuration, can be evaluated on any task provided it conforms to the action interface (shell commands, GUI interactions, and file I/O). Second, the same task specification can be deployed across different environment backends (cloud VMs or local containers) without modification.

\subsection{Task Specification Protocol}
\label{app:task-spec-protocol}

Every task specification implements three lifecycle phases that together ensure deterministic, reproducible evaluation.

\noindent \textbf{Phase 1: \texttt{load()} (Initialization).}
The \texttt{load()} function is purely declarative: it returns a structured task object containing the natural-language description visible to the agent, metadata (filesystem paths, configuration parameters, task-specific constants), and compute requirements (operating system type, hardware specifications). No remote connection is established and no environment state is modified. This phase defines \emph{what} the task is.

\noindent \textbf{Phase 2: \texttt{start()} (Environment Preparation).}
The \texttt{start()} function transforms the virtual machine into the task's deterministic starting state. Operating through a \emph{session API} that provides programmatic access to the remote desktop, including file-system operations (create, copy, delete), application management (launch, install), keyboard and mouse control, and screen capture, etc.

\noindent \textbf{Phase 3: \texttt{evaluate()} (Scoring).}
After the agent terminates, the \texttt{evaluate()} function retrieves the agent's output artifacts from the remote environment and scores them against the reference assets or rubric criteria defined in the specification. The function returns a normalized score in $[0, 1]$. The concrete evaluation methodology (deliverable-based extraction versus milestone-based reporting, rubric comparison versus reference matching, and the anti-gaming provenance gates that guard against shortcut solutions) is summarized in Section~\ref{sec:evaluation-modes} and detailed in Appendix~\ref{app:evaluation-modes}.

\subsection{Evaluation Modes: Full Taxonomy and Worked Examples}
\label{app:evaluation-modes}

This appendix expands Section~\ref{sec:evaluation-modes}. We document (i) where the scoring code runs, (ii) the artifact modes that a task author can choose for the comparison, (iii) the score-composition patterns observed across the benchmark, and (iv) the helper layer that backs LLM-as-judge evaluations. Concrete task implementations are cited throughout so that readers can inspect the full source.

\subsubsection{Execution Locale}
\label{app:eval-locale}

\noindent \textbf{Host-side scoring (default).}
The harness pulls the agent's artifact off the VM with \texttt{session.read\_file} or \texttt{session.read\_bytes}, then runs the scoring code in the host Python process. This is the default whenever (a) the artifact is small enough to transfer and (b) the scoring tooling is available off-VM. Examples:
\begin{itemize}[nosep,leftmargin=1.2em]
    \item \texttt{finance/equity\_research\_summary} reads the produced LibreOffice workbook bytes and calls \texttt{score\_workbook\_bytes} against a manifest.
    \item \texttt{cybersecurity/snake\_crackme} reads a flag file, normalizes the text, and compares the SHA-256 of every candidate against the expected digest.
    \item \texttt{photography/raw\_photo\_processing} compares exported file pairs against a reference manifest.
\end{itemize}
Host-side scoring is preferred because the scoring code is more easily reviewed, version-controlled, and re-run offline against a saved trajectory.

\noindent \textbf{VM-side verifier.}
When the artifact requires software that cannot be reasonably moved off the VM (CAD/CAM kernels, headless 3D renderers, vendor-licensed engines, very large geometry), \texttt{evaluate()} uploads a per-task script from \texttt{tasks/<task>/scripts/} into a temporary directory on the VM and invokes it with \texttt{session.run\_command}; the script prints a JSON result on stdout, which the host parses into a score. Examples:
\begin{itemize}[nosep,leftmargin=1.2em]
    \item \texttt{manufacturing/gcode} uploads \texttt{check\_collision.py}, \texttt{simulate\_agent.py}, and \texttt{verify\_stl.py} to drive PowerMill's COM API for collision detection and stock simulation, then to score the resulting STL surface against a hidden reference STL.
    \item \texttt{finance/sec\_10k\_financial\_parsing} uploads \texttt{score\_outputs.py} which uses the VM's Python environment to score the parsed filings against a multi-file reference manifest.
\end{itemize}
The contract is uniform: VM-side verifiers communicate with the host strictly through their stdout JSON, never by writing into \texttt{output/}.

\subsubsection{Artifact Modes}
\label{app:eval-modes}

Each task workflow author selects one or more of the following modes for the comparison step. Table~\ref{tab:eval-modes} summarizes the modes and representative task workflows.

\begin{table}[htbp]
    \centering
    \footnotesize
    \setlength{\tabcolsep}{4pt}
    \begin{tabular}{>{\raggedright\arraybackslash}p{2.4cm} >{\raggedright\arraybackslash}p{3.0cm} >{\raggedright\arraybackslash}p{1.6cm} >{\raggedright\arraybackslash}p{1.6cm} >{\raggedright\arraybackslash}p{4.0cm}}
        \toprule
        \textbf{Mode} & \textbf{Reference form} & \textbf{Locale} & \textbf{Judge} & \textbf{Example task workflow}\\
        \midrule
        Exact / hashed value & secret / digest & host & code & \texttt{cybersecurity/}\allowbreak\texttt{snake\_crackme}\\
        Structured tabular & manifest of (field, value, tolerance) & host or VM & code & \texttt{finance/}\allowbreak\texttt{sec\_10k\_}\allowbreak\texttt{financial\_parsing}\\
        Geometric / spatial & STL / mesh / point cloud & VM & code & \texttt{manufacturing/}\allowbreak\texttt{gcode}\\
        Visual appearance & reference screenshots & host & vision LLM & \texttt{game/}\allowbreak\texttt{mota\_reproduction}\\
        Behavioral / world state & deterministic state dump & VM & code & \texttt{architecture/}\allowbreak\texttt{parametric\_}\allowbreak\texttt{energy\_simulation}\\
        Free-text / semantic & rubric & host & LLM & \texttt{finance/}\allowbreak\texttt{equity\_research\_}\allowbreak\texttt{summary} (rubric component)\\
        Executable artifact & test set / oracle program & host or VM & code & \texttt{data\_computer\_}\allowbreak\texttt{science/}\allowbreak\texttt{data\_pipeline\_}\allowbreak\texttt{etl\_instance\_1}\\
        \bottomrule
    \end{tabular}
    \caption{Evaluation modes available to task workflow authors. Most ALE task workflows combine two or more modes (e.g., a behavioral gate with a geometric score).}
    \label{tab:eval-modes}
\end{table}

\noindent \textbf{Exact / hashed values.} The deliverable is a short string (a flag, a parsed answer, an identifier). Scoring is byte-equal or hash-equal after normalization. Because the answer is small but the search space is large, this mode is dominant in cybersecurity and a subset of mathematics task workflows.

\noindent \textbf{Structured tabular / numeric.} The deliverable is a table or a multi-field record (cells of a workbook, line items of a financial filing, parameters of a calibration). The reference is a manifest of (field, value, tolerance) tuples; per-field credit is granted within tolerance and aggregated. This mode is dominant in finance, accounting, and clinical-data-standards task workflows.

\noindent \textbf{Geometric / spatial.} The deliverable is a 3D mesh, a point cloud, or any spatially-embedded artifact. Scoring uses surface-distance functions (e.g., 10{,}000 surface samples scored against a reference STL in \texttt{gcode}, with credit for each fraction-within-threshold band).

\noindent \textbf{Visual appearance.} The deliverable's correctness is most naturally judged by human eye (a rendered scene, a recolored photo, a UI screenshot). The host calls a vision LLM judge with the agent and reference images side by side and a rubric question.

\noindent \textbf{Behavioral / world state.} The deliverable is the state of an interactive system after the agent's edits. Scoring replays the system under a fixed trajectory and dumps a comparable state (game map, simulation log, NPC events).

\noindent \textbf{Free-text / semantic.} The deliverable is a written report. Scoring is a rubric of yes/no or graded sub-criteria, evaluated by an LLM judge.

\noindent \textbf{Executable artifact.} The deliverable is a program, a model, or a pipeline. Scoring runs the artifact against a held-out test set or oracle and aggregates per-instance correctness.

\subsubsection{Score-Composition Patterns}
\label{app:eval-composition}

Per-mode scores are combined into the final $[0, 1]$ value by one of four patterns:

\noindent \textbf{Gate-and-score.} A hard precondition that forces $0$ on failure, followed by a continuous score. Used to prevent reward hacking by superficially close artifacts. Canonical example: in \texttt{manufacturing/gcode}, a PowerMill collision/gouge gate must pass before any geometric similarity is awarded; otherwise the task workflow scores $0$ regardless of how close the simulated stock model is to the reference.

\noindent \textbf{Weighted rubric.} An expert-defined weighted sum over multiple sub-metrics, e.g.\ \texttt{gcode}'s
\[
    \text{score} = 0.70 \cdot \mathrm{frac\_within}(0.3\,\text{mm}) + 0.30 \cdot \mathrm{frac\_within}(2.0\,\text{mm})
\]
on the surface-distance distribution between agent and reference STLs. Weights are part of the task spec and are reviewed during the QC stage of Section~\ref{sec:construction-pipeline}.

\noindent \textbf{Binary checklist averaging.} The deliverable is judged by $N$ independent yes/no questions, and the score is the mean. The \texttt{game/mota\_reproduction} task workflow uses this pattern over a set of vision-LLM probes (engine identification, character sprite presence, map layout match).

\noindent \textbf{Pairwise file aggregation.} When the deliverable is a directory of files matched against a reference directory, the helper \texttt{utils.evaluation.collect\_matching\_files} pairs each agent file with its reference, runs the same scoring function on every pair, and returns the mean.

\subsubsection{Judge Types and the LLM-Judge Helper Layer}
\label{app:eval-judges}

\noindent \textbf{Avoid LLM judges by default.}
ALE deliberately suppresses the use of LLM-as-judge wherever a deterministic alternative exists. The standing rule for every accepted task workflow is: if the deliverable can be reduced to bytes, fields, geometry, world state, or executable behavior, the scoring code must operate on those signals, not on a model's holistic opinion of the result. A task that proposes ``ask GPT-4 whether the result looks correct'' is rejected at the QC stage of Section~\ref{sec:construction-pipeline} and either re-engineered to expose a checkable artifact or dropped. This is enforced for three reasons: (i) judge models drift across releases, which would silently re-rank agents; (ii) general-purpose ``does this look right?'' prompts are too soft to discriminate near-correct from correct, which is exactly the regime where the benchmark needs to resolve agents; and (iii) deterministic judges can be re-run offline against a saved trajectory by anyone who has the artifact, with no API cost.

\noindent \textbf{When LLM judging is unavoidable.}
A small set of task workflows have no objective code-based reference, primarily creative or perceptual deliverables such as rendered scenes, music-production sessions, UV-mapped textures, and animation previews. For these, ALE uses the helper layer in \texttt{utils/evaluation.py}:
\begin{itemize}[nosep,leftmargin=1.2em]
    \item \texttt{llm\_vision\_yes\_no\_judge}: a single targeted yes/no visual question over an (agent image, reference image) pair.
    \item \texttt{llm\_vision\_binary\_questions\_sync} / \texttt{llm\_vision\_binary\_checklist\_judge}: a list of independent yes/no probes whose final score is a fraction.
    \item \texttt{llm\_vision\_judge} / \texttt{llm\_vision\_json\_judge}: a graded or JSON-structured rubric over a small fixed set of fields.
\end{itemize}

\noindent \textbf{Targeted probes, not general judging.}
The single most important property of ALE's LLM-judged workflows is that the prompts \emph{never} ask the model to score the artifact in the abstract. Every prompt is a narrow, evidence-anchored yes/no probe, written by the task author with reference to (a) the specific software the workflow uses and (b) failure modes the author has observed in pilot agent runs. The model is asked one structurally tight question at a time; the score is composed by code from those answers. The following are verbatim probes from current task workflows:

\begin{itemize}[nosep,leftmargin=1.2em]
    \item \texttt{game/mota\_reproduction} (replay-time engine and state checks):
    \begin{itemize}[nosep,leftmargin=1.2em]
        \item ``Does the first image show that the game is developed using RPGMakerXP? One can identify whether there is an `orange sun-like circle' in the top-left corner of the game window.'' (gate)
        \item ``Does the first image show with the same map layout as in the original game?''
        \item ``Does the first image show with the same player status as in the original game?''
    \end{itemize}
    \item \texttt{audio/timbre\_synthesis} (proves the agent actually used the DAW):
    \begin{itemize}[nosep,leftmargin=1.2em]
        \item ``Does this image show either (a) a software synthesizer / VST plugin interface with visible parameters such as oscillators, filters, envelopes, LFOs, or effects, OR (b) a DAW (e.g., Cubase, Ableton, FL Studio, Logic, Reaper) arrangement / piano-roll / mixer view that contains one or more tracks with audio or MIDI clips?'' (gate)
        \item ``Does this screenshot provide evidence that the user has actually worked on the project, for example a DAW arrangement containing multiple tracks with audio or MIDI clips, a piano roll with notes entered, or a synthesizer plugin whose knobs/sliders are clearly not all in their default/init positions?''
    \end{itemize}
    \item \texttt{game/uv\_reproduction} (UV-mapping artifact checks against a fixed reference):
    \begin{itemize}[nosep,leftmargin=1.2em]
        \item ``Is the texture placement and orientation correct enough to pass?''
        \item ``Does the candidate preserve the reference material appearance and color palette well enough to pass?''
        \item ``Are obvious UV seams, stretching, or texture artifacts absent enough for the result to pass?''
    \end{itemize}
    \item \texttt{game/high\_to\_low\_modeling} (decimation correctness):
    \begin{itemize}[nosep,leftmargin=1.2em]
        \item ``Does the candidate preserve the overall highpoly shape well enough to pass?''
        \item ``Does the candidate preserve the main silhouettes across the sampled views well enough to pass?''
        \item ``Does the candidate achieve a meaningful lowpoly reduction rather than effectively submitting the highpoly again?''
        \item ``Are obvious visual artifacts or shape collapses absent enough for the result to pass?''
    \end{itemize}
    \item \texttt{game/object\_generation} (missing-geometry restoration):
    \begin{itemize}[nosep,leftmargin=1.2em]
        \item ``Is the missing geometry restored well enough to pass?''
        \item ``Is the part placement and alignment correct enough to pass?''
        \item ``Is the whole object coherent and complete enough to pass?''
        \item ``Is the final material appearance acceptable enough to pass?''
    \end{itemize}
    \item \texttt{game/skeletal\_animation\_reproduction} (replay self-consistency):
    \begin{itemize}[nosep,leftmargin=1.2em]
        \item ``Does the submitted preview match the reference body motion well enough to pass?''
        \item ``Does the replay rendered from \texttt{final.blend} agree with the submitted preview well enough to pass?''
        \item ``Do the visible skeleton states and poses look natural and non-broken enough to pass?''
    \end{itemize}
\end{itemize}

\noindent Three patterns recur across these probes. (1) Each question targets one identifiable artifact (a circle in a corner, a track in a DAW, a UV seam, a silhouette, a pose), not the gestalt of ``is the deliverable good.'' (2) Many probes are written as gates, i.e., a binary precondition that must hold before the rest of the rubric is even checked, so that an unrelated screenshot or a placeholder file scores $0$ before any quality judgment is asked. (3) The remaining probes are phrased as ``\dots enough to pass?'' rather than as graded preferences, which converts the model into a same-vs.-different comparator against a fixed reference rather than a free-form quality oracle. Together these conventions limit the burden on the LLM to a series of decisions that a domain expert could replicate from the same image, and keep the LLM out of the role of integrator: the integration (weighting, gating, and aggregation across instances) always happens in code.

\subsubsection{Empirical Distribution of Judge Types}
\label{app:eval-judge-stats}

To make the design choice ``code-based by default, LLM only when unavoidable'' concrete, we report the actual distribution of judge types and execution locales at the task-workflow level. The proportions in Table~\ref{tab:judge-stats} are obtained by static analysis of every \texttt{main.py} in the open-sourced reference task tree together with its accompanying \texttt{scripts/} directory, scanning for direct invocations of the LLM-judge helpers in \texttt{utils/evaluation.py} and for the use of \texttt{session.run\_command} on uploaded Python verifiers.

\begin{table}[htbp]
    \centering
    \footnotesize
    \begin{minipage}[t]{0.48\textwidth}
        \centering
        \begin{tabular}{l r}
            \toprule
            \textbf{Judge type} & \textbf{Share}\\
            \midrule
            Code-based (deterministic)   & 93.2\%\\
            LLM-as-judge                 &  6.8\%\\
            \bottomrule
        \end{tabular}
        \\[0.4em]
        \textbf{(a)} Judge type per task workflow.
    \end{minipage}\hfill
    \begin{minipage}[t]{0.48\textwidth}
        \centering
        \begin{tabular}{l r}
            \toprule
            \textbf{Execution locale} & \textbf{Share}\\
            \midrule
            Host-side                       & 88.5\%\\
            VM-side verifier                & 11.5\%\\
            \bottomrule
        \end{tabular}
        \\[0.4em]
        \textbf{(b)} Execution locale of the scoring code.
    \end{minipage}
    \caption{Distribution of judge type and execution locale across the open-sourced task workflows in the ALE reference task tree.}
    \label{tab:judge-stats}
\end{table}

\noindent \textbf{Within the LLM-judged subset.}
Vision-grounded primitives dominate: the most-used helpers are \texttt{llm\_vision\_judge}, \texttt{llm\_vision\_binary\_checklist\_judge}, \texttt{llm\_vision\_binary\_questions\_sync}, and \texttt{llm\_vision\_yes\_no\_judge}. The remaining helpers (\texttt{llm\_multimodal\_binary\_questions\_sync}, \texttt{llm\_multimodal\_text}, \texttt{llm\_multimodal\_json}, \texttt{llm\_vision\_json\_judge}, and the video-rubric \texttt{gemini\_video\_json\_judge}) appear in only a handful of task workflows each. Helpers may co-occur within a single task workflow. The concentration on vision helpers reflects the fact that LLM judging is reserved for cases where the deliverable is a rendered scene, photo, screenshot, or short video that has no objective code-based reference.

\noindent \textbf{Within the VM-side subset.}
VM-side task workflows are dominated by industries whose deliverable cannot be scored without the on-VM software stack: CAD/CAM (PowerMill, SolidWorks), licensed financial workbooks, and headless 3D rendering. The contract is uniform: a per-task-workflow verifier under \texttt{tasks/<task-workflow>/scripts/} is uploaded by \texttt{evaluate()} into a temporary directory on the VM, executed via \texttt{session.run\_command}, and its JSON stdout is parsed back into a score on the host.

\noindent \textbf{Composition patterns.}
A static-analysis estimate finds that the great majority of \texttt{evaluate()} bodies contain at least two early \texttt{return [0.0]} sites preceding a continuous-score path, consistent with the gate-and-score pattern of Section~\ref{app:eval-composition}. Explicit weighted-rubric expressions of the form $a\cdot x + b\cdot y$ appear in only a small fraction of \texttt{main.py} files; in most task workflows the weighting is encoded inside per-task-workflow scoring scripts under \texttt{scripts/} (which the static count above does not pick up), so this is a lower bound on the true prevalence of weighted aggregation.

\subsubsection{Reference Isolation and Robustness}
\label{app:eval-isolation}

\noindent \textbf{Reference isolation.}
The \texttt{reference/} directory lives outside the agent's workspace and is not exposed through any session API a non-evaluator caller would invoke. Most \texttt{evaluate()} implementations begin with a sanity check that lists every required reference path and returns $0.0$ early if any path is missing, which both prevents misconfigured runs from silently producing inflated scores and documents the reference contract in the same file as the scoring logic.

\noindent \textbf{Output presence and shape checks.}
Workflows consistently treat ``no output produced'' or ``output in the wrong shape'' as score $0$ rather than as a crash, so that an agent that times out or refuses still emits a well-defined number.

\noindent \textbf{Determinism.}
Code-based judges are deterministic by construction. For LLM-judged task workflows, we record the judge model and prompt with the result and make sure that any score can be re-derived from the agent's saved artifacts.

\subsubsection{Workflows and Task Instances}
\label{app:eval-variants}

A single task workflow (one \texttt{main.py}) exposes a list of \emph{task instances}, encoded in the codebase as the \texttt{VARIANTS} tuple list: each instance carries instance-specific configuration but shares the same \texttt{evaluate()}. For example, the \texttt{manufacturing/gcode} task workflow declares 18 workpiece instances, each pointing at a different blank PowerMill project but scored by the same collision-gate-then-STL pipeline. Per-instance scores are averaged into a task workflow score, task workflow scores are averaged into industry scores, and industry scores aggregate into the cluster-level results reported in Section~\ref{sec:experiment}. The current ALE release contains \totaltasks{} task workflows and \totalvariants{} task instances in total.

\subsection{Agent Harness Internals}
\label{app:agent-harness}

This appendix elaborates on the internal structure of the agent harness introduced in Section~\ref{sec:agent-architecture}. The architecture described here is shared, at a macro level, across mainstream harness implementations such as Claude Code~\citep{anthropic2025claudecode}, Codex~\citep{openai2025codex}, and OpenClaw, and is faithfully reproduced in our own native implementation.

\noindent \textbf{Main agent loop.}
The harness operates a six-phase control loop: \circled{0}~\emph{Initialization} configures the system prompt and tool bindings; \circled{1}~\emph{Context Building} assembles the current conversation state; \circled{2}~\emph{LLM Call} queries the foundation model; \circled{3}~\emph{Decide} routes the model's output to either a final delivery or a tool invocation; \circled{4}~\emph{Collect Tool Result} gathers the execution outcome; \circled{5}~\emph{Overflow Check} evaluates whether the accumulated context exceeds a compaction threshold. If not, the loop returns to phase~1; otherwise, context compaction is triggered before the next iteration. The loop terminates when the model elects to deliver rather than act.

\noindent \textbf{System prompt builder.}
At initialization, the harness constructs a system prompt from modular components: \emph{Identity} (agent persona), \emph{Memory} (persistent cross-session state), \emph{Tool Guidance} (usage conventions for each tool), \emph{Runtime} (environment metadata), \emph{Behavioral Rules} (safety and policy constraints), and \emph{Skills} (domain-specific capabilities). These components are typically authored through configuration files such as \texttt{CLAUDE.md} or \texttt{AGENTS.md}.

\noindent \textbf{Tool system.}
The harness exposes a unified tool interface that the model invokes by name: file operations (read, write, glob, grep), shell execution, web search and fetch, and sub-agent management (spawn, list, wait, terminate). Each tool returns structured results that are appended to the conversation context.

\noindent \textbf{GUI-as-Tool: CUA MCP bridge.}
The GUI-as-Tool mode extends the tool system with 14 desktop-action tools exposed through an MCP server that wraps a CUA (Computer-Use Agent) HTTP API running on the VM. Table~\ref{tab:gui-tools} lists the full tool surface.

\begin{table}[h]
\caption{\textbf{GUI-as-Tool: 14 desktop-action tools} exposed via the CUA MCP bridge.}
\label{tab:gui-tools}
\centering
\small
\begin{tabular}{@{}llp{7.2cm}@{}}
\toprule
Group & Tool & Description \\
\midrule
\multirow{5}{*}{Keyboard}
 & \texttt{key} & Press and release one or more keys (supports hotkeys, e.g.\ \texttt{["ctrl","c"]}) \\
 & \texttt{key\_down} & Press keys down without releasing (for modifier holds) \\
 & \texttt{key\_up} & Release previously held keys \\
 & \texttt{type} & Type text into the currently focused input field \\
 & \texttt{hold\_key} & Hold keys for a specified duration, then release \\
\midrule
\multirow{6}{*}{Mouse}
 & \texttt{mouse\_move} & Move the cursor to a coordinate \\
 & \texttt{click} & Click at a coordinate (left/right/middle; single/double/triple) \\
 & \texttt{drag} & Drag from a start coordinate to an end coordinate \\
 & \texttt{mouse\_down} & Press a mouse button without releasing \\
 & \texttt{mouse\_up} & Release a mouse button \\
 & \texttt{scroll} & Scroll in a direction (up/down/left/right) by a specified amount \\
\midrule
\multirow{3}{*}{Utility}
 & \texttt{screenshot} & Capture the current screen; optionally save to a VM path \\
 & \texttt{cursor\_position} & Return the current cursor coordinates \\
 & \texttt{wait} & Pause execution for a specified duration \\
\bottomrule
\end{tabular}
\end{table}

\subsubsection{Tool Surface and Terminology}
\label{app:tool-surface}

\noindent \textbf{Tool taxonomy and per-agent availability.}
Tool names differ across harnesses, so the analysis in Figure~\ref{fig:tool-usage} maps raw tool calls into a common taxonomy before aggregation.
\emph{Bash} denotes direct shell or terminal execution, including tools named \texttt{Bash}, \texttt{bash}, \texttt{shell}, \texttt{exec}, \texttt{run\_shell\_command}, \texttt{terminal}, \texttt{Execute}, \texttt{bash\_command}, and \texttt{execute\_code}.
\emph{File} denotes direct file-system tools such as \texttt{Read}, \texttt{Write}, \texttt{Edit}, \texttt{Glob}, \texttt{Grep}, \texttt{read\_file}, \texttt{write\_file}, \texttt{edit\_file}, \texttt{patch}, and \texttt{list\_directory}.
\emph{GUI} denotes the CUA desktop-action surface in Table~\ref{tab:gui-tools}, including wrapper-specific names such as \texttt{mcp\_\_cua\_\_click}, \texttt{cua\_\_\_screenshot}, and \texttt{mcp\_cua\_key}.
\emph{Web} denotes browser or retrieval tools such as \texttt{WebSearch}, \texttt{WebFetch}, \texttt{web\_search}, \texttt{web\_fetch}, \texttt{webSearch}, \texttt{webFetch}, and \texttt{browser\_navigate}.
\emph{Planning/delegation} denotes explicit planning, task-tracking, memory, or sub-agent tools such as \texttt{TodoWrite}, \texttt{task}, \texttt{think}, \texttt{delegate}, \texttt{subagents}, and \texttt{memory\_get}.
\emph{Other} captures process, session, finish, or harness-internal utilities that do not fit the preceding groups.
Finally, \emph{Azure desktop} refers to the hosted Windows remote-desktop backend used for Windows GUI task execution; it is an execution substrate, not a model provider and not a separate tool class.

\noindent \textbf{Sub-agents.}
Complex tasks benefit from delegation. The harness can spawn specialized sub-agents (a \emph{General} sub-agent with access to all tools, an \emph{Explore} sub-agent restricted to read-only operations, among others) that operate in isolated context windows and return summarized results to the parent loop. This mechanism enables parallel exploration and limits context consumption.

\noindent \textbf{Context manager.}
Long-horizon professional tasks routinely generate context that exceeds model limits. The context manager implements a three-tier compaction strategy: (1)~\emph{Microcompaction} clears stale tool results in place; (2)~\emph{LLM-based summarization} compresses older conversation segments into structured checkpoints; (3)~\emph{Truncation} enforces hard context-window limits (e.g., 400K or 1M tokens). This graduated approach preserves recent detail while retaining long-range planning state.

\noindent \textbf{ALE-Claw vs.\ OpenClaw.}
OpenClaw is a personal AI assistant with two main components: a user-interaction runtime and an agent loop.
For ALE-Claw, we removed the components that keep a long-lived, multi-user AI assistant alive in production: the scheduled-prompt subsystem, including cron and heartbeats; multi-channel gateways; the skills system; and the plugin framework with lifecycle hooks.
These components are not needed to solve a single benchmark task.
The simplification reduces the system prompt by ${\sim}65\%$.

The agent loop is similar in principle to the design in Figure~\ref{fig:agent-harness}: it takes a task instruction and turns it into a sequence of tool calls and observations until the task is complete.
OpenClaw was originally developed in TypeScript~\citep{openclaw2026agent}, which makes direct adaptation to the CUA framework difficult.
To resolve this, we rewrote the agent loop in Python and added a small set of CUA-specific adaptations: a composite computer-use tool that matches CUA's native GUI surface, and a vision-driven GUI sub-agent, \texttt{delegate\_gui}, with no OpenClaw analogue.
We retained OpenClaw's load-bearing context-management primitives near-verbatim.

ALE-Claw is also of independent interest.
By isolating OpenClaw's agent loop and porting it to Python, we make the design accessible to the broader Python research ecosystem and to the CUA framework specifically, where the original TypeScript implementation cannot be plugged in directly.
The Python port also enables dynamic ablation of the agent harness.
Components can be swapped or removed in place, and the resulting performance shift can be measured directly; this workflow is impractical against the original TypeScript runtime.
As one concrete future direction, ALE-Claw can serve as a fixed scaffold for benchmarking different GUI models against the same task suite.

\section{Extended Experiment Results and Analysis}
\label{app:extended-results}

\subsection{Public-Subset Representativeness}
\label{app:fullpool}

\begin{figure}[htbp]
\centering
\includegraphics[width=0.55\linewidth]{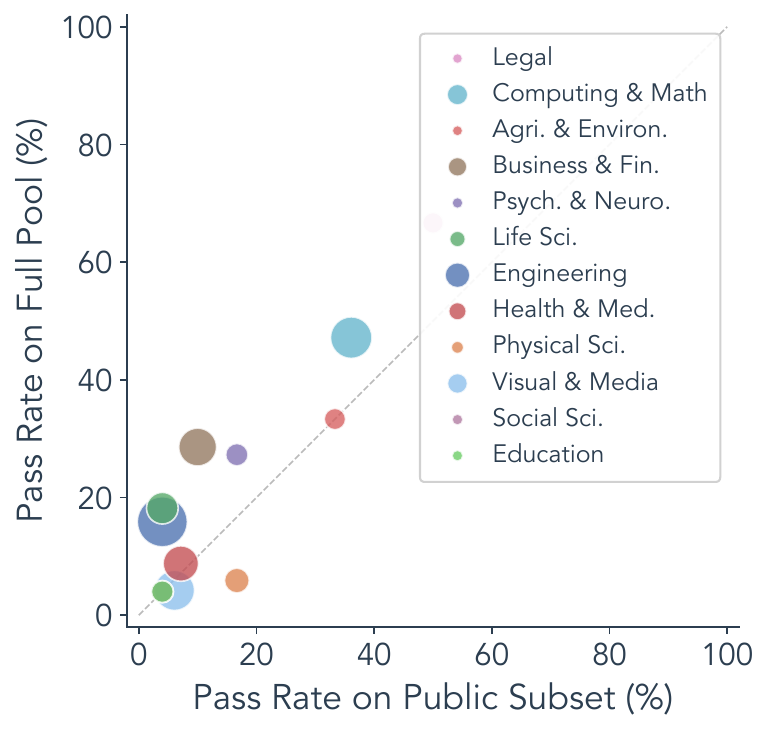}
\caption{\textbf{Public-subset representativeness.}
Pass rate per taxonomy cluster on the public subset ($x$) vs.\ the full task pool ($y$) for Claude Code + Opus~4.7.
Point size $\propto$ total task instances per cluster.
The strong correlation ($r{=}0.89$) confirms the public subset is representative.}
\label{fig:fullpool-scatter}
\end{figure}

To verify that the public subset is representative of the broader benchmark despite its limited size (Section~\ref{sec:construction-pipeline}), we ran Claude Code + Opus~4.7 on the full task pool.
The full task pool yield a higher pass rate.
The gap arises because the public set includes the full Last-Exam tier, while the private pool contains proportionally more Near-Term-level tasks.
Figure~\ref{fig:fullpool-scatter} compares pass rates per taxonomy cluster on the public subset versus the full task pool.
The two axes are strongly correlated (Pearson $r{=}0.89$, $p{<}0.001$), indicating that the public subset faithfully reflects full-pool difficulty across domains.
Most clusters lie above the diagonal because the private pool's higher share of easier tasks raises each cluster's full-pool pass rate.

\subsection{Timeout Analysis}
\label{app:timeout-analysis}

ALE evaluations use a five hour wall clock cap per run.
When a run reaches the cap, the harness stops the agent and the normal evaluator scores whatever artifacts are present in the output directory.
In the runs used for Table~\ref{tab:main-results}, 3.8\% of evaluated runs reached the cap.
Runs that reached the cap have a mean score of 20.7, compared with 33.2 for runs that ended earlier.

\begin{table}[h]
\centering
\caption{Timeout frequency by difficulty tier. Scores are mean normalized scores on the same 0 to 100 scale used in Table~\ref{tab:main-results}.}
\label{tab:timeout-by-tier}
\begin{tabular}{lrrr}
\toprule
Tier & Timeout rate & Timeout score & Other score \\
\midrule
Near-Term & 3.1\% & 39.2 & 48.6 \\
Full-Spectrum & 3.2\% & 17.7 & 25.9 \\
Last-Exam & 6.4\% & 1.3 & 7.0 \\
\bottomrule
\end{tabular}
\end{table}

\begin{table}[h]
\centering
\caption{Timeout frequency by harness for harnesses with at least one run that reached the cap.}
\label{tab:timeout-by-harness}
\begin{tabular}{lrrr}
\toprule
Harness & Timeout rate & Timeout score & Other score \\
\midrule
OpenClaw & 5.7\% & 21.3 & 29.3 \\
Cursor & 4.8\% & 34.6 & 44.2 \\
Claude Code & 4.5\% & 22.0 & 40.1 \\
Terminus & 4.4\% & 0.0 & 35.8 \\
Gemini CLI & 4.2\% & 3.3 & 35.4 \\
Codex & 2.0\% & 35.1 & 36.8 \\
ForgeCode & 2.0\% & 0.0 & 33.3 \\
ALE-Claw & 1.2\% & 6.7 & 40.5 \\
Grok CLI & 1.0\% & 0.0 & 17.4 \\
Droid & 0.6\% & 3.0 & 45.9 \\
Hermes & 0.4\% & 0.0 & 34.8 \\
OpenHands & 0.3\% & 0.0 & 21.5 \\
\bottomrule
\end{tabular}
\end{table}

\subsection{Failure Taxonomy Classification}
\label{app:failure-taxonomy}

This appendix documents the two-stage pipeline behind the failure root-cause taxonomy presented in Figure~\ref{fig:failure-taxonomy}.

\subsubsection{Stage 1: Trajectory analysis}
For each failed task, an LLM (OpenAI Codex) was given access to the full run artifact directory, including the agent interaction log (\texttt{interaction\_log.json}), run metadata (\texttt{run\_result.json}, \texttt{agent\_result.json}), evaluation output (\texttt{debug/eval/result.json}), and event traces (\texttt{events.jsonl}).
The LLM was prompted to produce a structured \emph{analysis card} in Markdown with five mandatory sections:

\begin{enumerate}[leftmargin=*,itemsep=2pt]
    \item \textbf{Conclusion}: a one-sentence verdict, an overall judgment (success / partial success / failure), and the single most important problem.
    \item \textbf{Task description}: a plain-language explanation of what the task asks, including key constraints and runtime metadata.
    \item \textbf{What the agent did right}: correct behaviors, with evidence pointers to specific log entries.
    \item \textbf{What the agent did wrong}: observed errors, with evidence pointers. The prompt requires separating confirmed errors from uncertain or inferred causes.
    \item \textbf{Scoring}: the final score, raw score breakdown (if available), inferred evaluation criteria, and a confidence rating.
\end{enumerate}

Each claim in the analysis card must cite a specific artifact file and field as evidence.
The prompt prohibits reading the full transcript (\texttt{transcript.jsonl}) to keep generation cost bounded; the interaction log provides a sufficient behavioral summary.

\subsubsection{Stage 2: Taxonomy classification}
Each analysis card was then classified into a two-level failure taxonomy using GPT-4o (temperature 0).
The classification prompt defines the following hierarchy:

\begin{itemize}[leftmargin=*,itemsep=2pt]
    \item \textbf{Understanding}: the agent lacked knowledge or fabricated information.
    \begin{itemize}[itemsep=1pt]
        \item \emph{Domain Knowledge Gap}: the agent's errors trace back to missing specialized expertise. The prompt instructs: ``Would a domain expert have avoided this mistake? If yes, classify here.''
        \item \emph{Hallucination/Fabrication}: the agent invented data or results instead of computing them.
    \end{itemize}
    \item \textbf{Approach}: the agent understood the domain but chose the wrong plan.
    \begin{itemize}[itemsep=1pt]
        \item \emph{Wrong Strategy}: the agent violated explicit task constraints or chose a fundamentally wrong method not attributable to domain ignorance.
        \item \emph{Incomplete/Abandoned}: the agent stopped early or failed to produce required deliverables.
    \end{itemize}
    \item \textbf{Execution}: the approach was sound but the implementation was flawed.
    \begin{itemize}[itemsep=1pt]
        \item \emph{Implementation Bug}: logic errors, calculation mistakes, or data processing bugs.
        \item \emph{Output Format Error}: output in wrong format, location, or structure.
    \end{itemize}
    \item \textbf{Infrastructure}: external constraints unrelated to agent capability.
    \begin{itemize}[itemsep=1pt]
        \item \emph{GUI/Browser Failure}: GUI or browser interaction failed due to tool issues.
        \item \emph{Timeout/Resources}: the agent ran out of time or computational resources.
    \end{itemize}
\end{itemize}

\subsubsection{Distribution}
Nearly half (47\%) of classifiable failures stem from Approach errors: wrong strategy (30\%) or premature abandonment (17\%).
Understanding failures account for 31\%, dominated by domain knowledge gaps (25\%) with a smaller fraction (6\%) involving hallucination or data fabrication.
The remaining 22\% are Execution errors: output format mismatches (10\%), implementation bugs (8\%), and GUI interaction failures (4\%).
Timeout and resource-exhaustion cases are excluded from this breakdown because they reflect environment constraints rather than agent reasoning failures; their prevalence is analyzed separately in Appendix~\ref{app:timeout-analysis}.

\subsection{Model vs.\ Harness Effect}
\label{app:harness-vs-model}

\begin{figure}[htbp]
\centering
\includegraphics[width=0.65\linewidth]{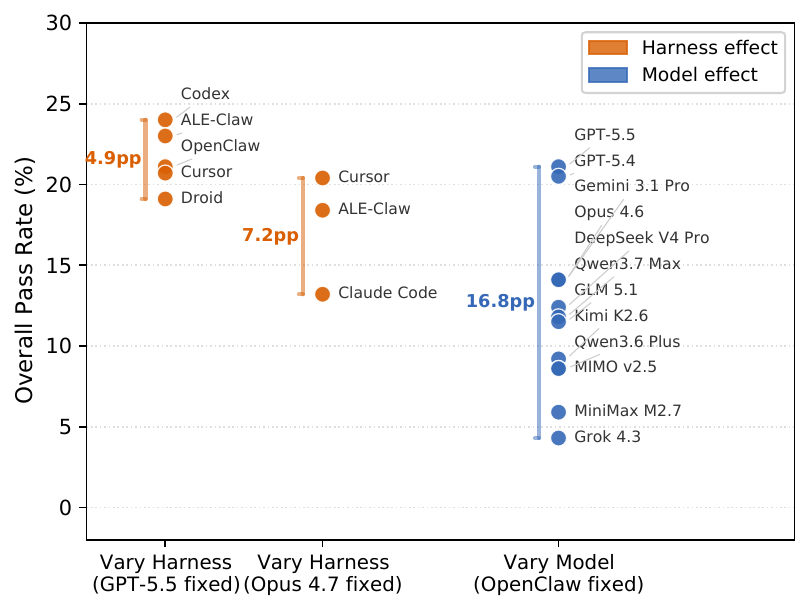}
\caption{\textbf{Model choice vs.\ harness choice.}
Each dot is one configuration; the vertical bracket shows the full range of overall pass rates.
Varying the backbone model under a fixed harness (OpenClaw, 12 models) produces a 16.8\,pp spread, roughly 3$\times$ the spread observed when varying the harness under a fixed backbone (4.9--7.2\,pp).}
\label{fig:harness-vs-model}
\end{figure}

A natural question raised by Table~\ref{tab:main-results} is whether performance differences are driven primarily by the choice of foundation model or the choice of agent harness.
Figure~\ref{fig:harness-vs-model} isolates the two factors.
Under a fixed OpenClaw harness, swapping the backbone model produces an overall pass-rate spread of 16.8 percentage points (from 4.3\% for Grok~4.3 to 21.1\% for GPT-5.5).
Under a fixed backbone, swapping the harness yields a much narrower spread: 4.9\,pp when the backbone is GPT-5.5 (five harnesses, 19.1--24.0\%) and 7.2\,pp when the backbone is Claude Opus~4.7 (three harnesses, 13.2--20.4\%).

The pattern is consistent across both backbone choices: among the competitive harnesses evaluated here, engineering differences in prompting strategy, tool routing, and context management account for only a modest share of overall performance variation.
The dominant factor is the foundation model's reasoning and domain knowledge, which aligns with the failure analysis in Section~\ref{sec:experiment-analysis}, where Understanding and Approach errors (both rooted in model capability) constitute the majority of failures.

\subsection{Cost, Time, and Token Efficiency}
\label{app:cost-performance}

\begin{figure}[htbp]
\centering
\includegraphics[width=\linewidth]{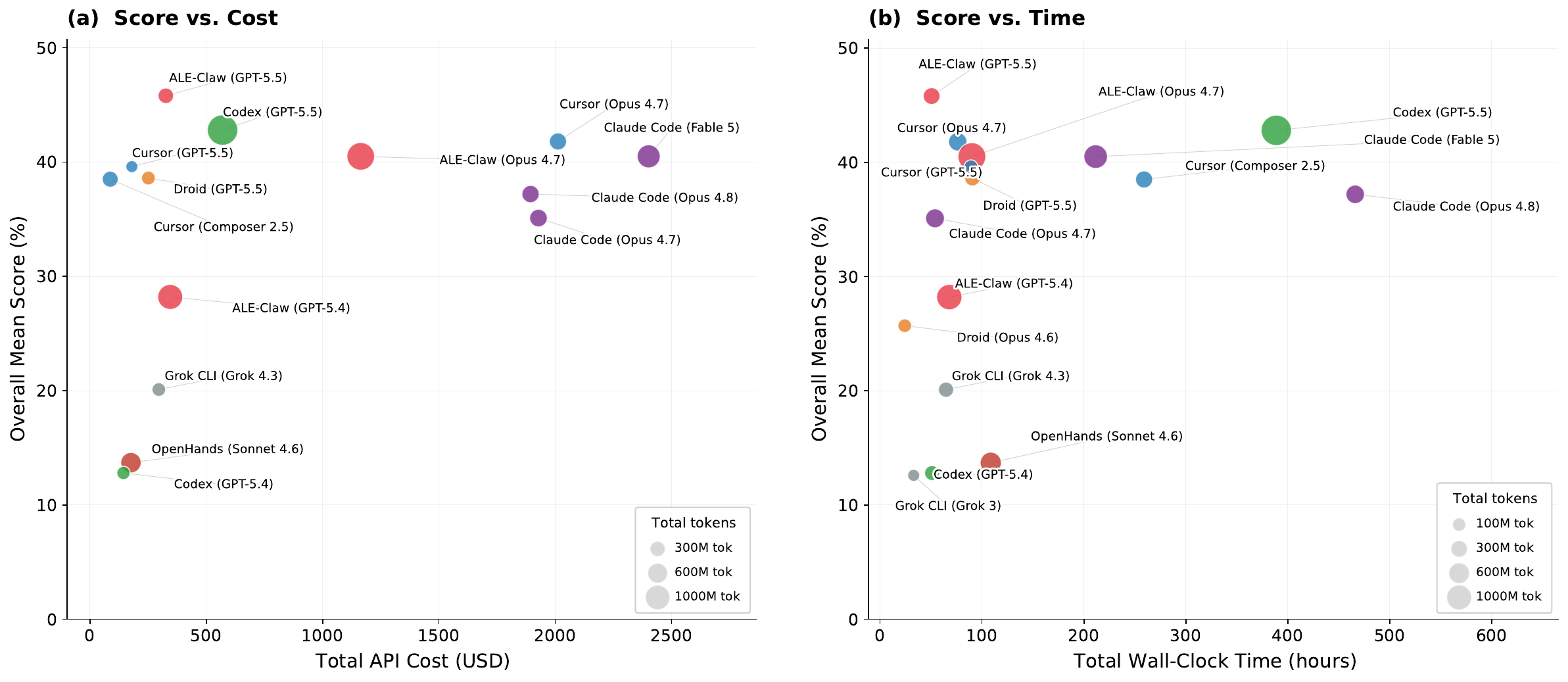}
\caption{\textbf{Performance vs.\ resource consumption for mainstream agent harnesses.}
Each bubble represents one harness--backbone configuration from Table~\ref{tab:main-results};
bubble area is proportional to total token consumption.
\textbf{(a)}~Overall mean score vs.\ total API cost (configurations with available cost data).
\textbf{(b)}~Overall mean score vs.\ total wall-clock time (all 16 configurations).
The ideal operating point is the upper-left corner of each panel (high score, low resource use).}
\label{fig:cost-performance}
\end{figure}

Figure~\ref{fig:cost-performance} visualizes the relationship between performance and resource consumption across the 16 mainstream harness--backbone configurations in Table~\ref{tab:main-results}.
Three observations emerge.

\noindent \textbf{Cost and performance are only loosely correlated.}
In panel~(a), ALE-Claw with GPT-5.5 achieves the highest overall mean score (45.8\%) at \$326 total API cost, while the same harness paired with Opus~4.7 spends $3.6\times$ more (\$1\,164) yet scores 5.3 percentage points lower (40.5\%).
Cursor with Composer~2.5 is the most frugal configuration, reaching 38.5\% for \$87, whereas Claude Code with Fable~5 spends \$2\,402, the highest of any configuration, for a comparable 40.5\%.
The spread indicates that higher spending does not reliably translate to better results; the backbone model's fit to the task distribution and the harness's token efficiency jointly determine the cost--performance tradeoff.

\noindent \textbf{Time efficiency varies widely.}
Panel~(b) reveals that wall-clock time is largely decoupled from score.
ALE-Claw (GPT-5.5) achieves the top score in approximately 51 hours of total wall-clock time, whereas Claude Code (Opus~4.8) requires 466 hours for a lower score (37.2\%).
Droid (Opus~4.6) is the fastest configuration (24 hours) but scores only 25.7\%.
The variation reflects differences in per-task timeout behavior, retry strategies, and the degree of parallelism in each harness's action loop.

\noindent \textbf{Token consumption does not predict performance.}
Bubble sizes in both panels show that token-heavy configurations are not necessarily higher-scoring.
ALE-Claw (Opus~4.7) consumes 1\,373M tokens yet scores slightly below Cursor (Opus~4.7) at 460M tokens (40.5\% vs.\ 41.8\%).
Conversely, Cursor (GPT-5.5) uses only 160M tokens while reaching 39.6\%, suggesting that concise tool use and efficient context management can compensate for raw token volume.

\subsection{Per-Task Instance Score Heatmaps}
\label{app:per-task-heatmap}

\begin{figure}[p]
\centering
\makebox[\textwidth][c]{\includegraphics[width=1.25\textwidth]{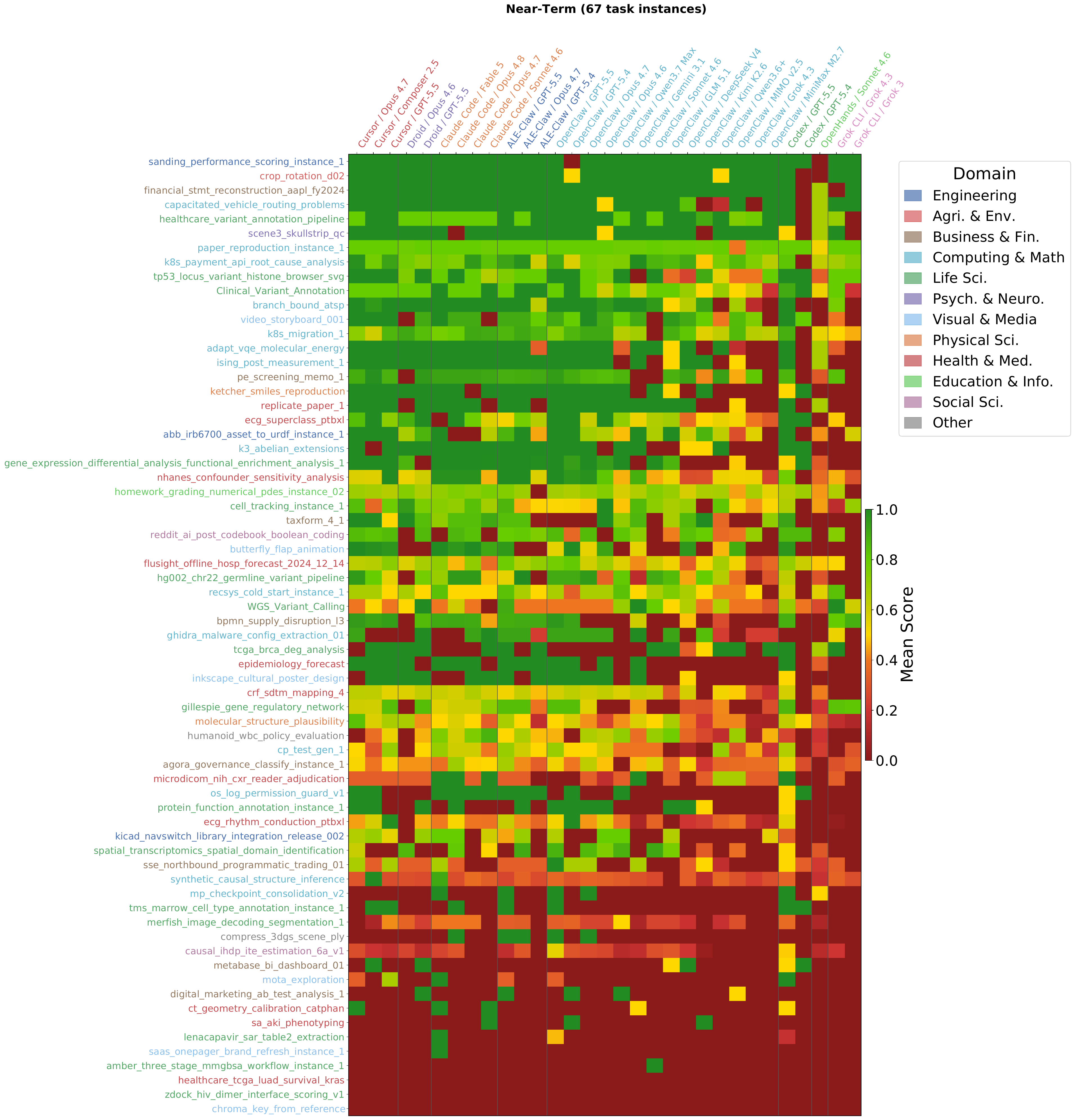}}
\caption{\textbf{Per-task instance scores: Near-Term tier} (\neartermcount{} task instances).}
\label{fig:heatmap-nearterm}
\end{figure}

\begin{figure}[p]
\centering
\makebox[\textwidth][c]{\includegraphics[width=1.25\textwidth]{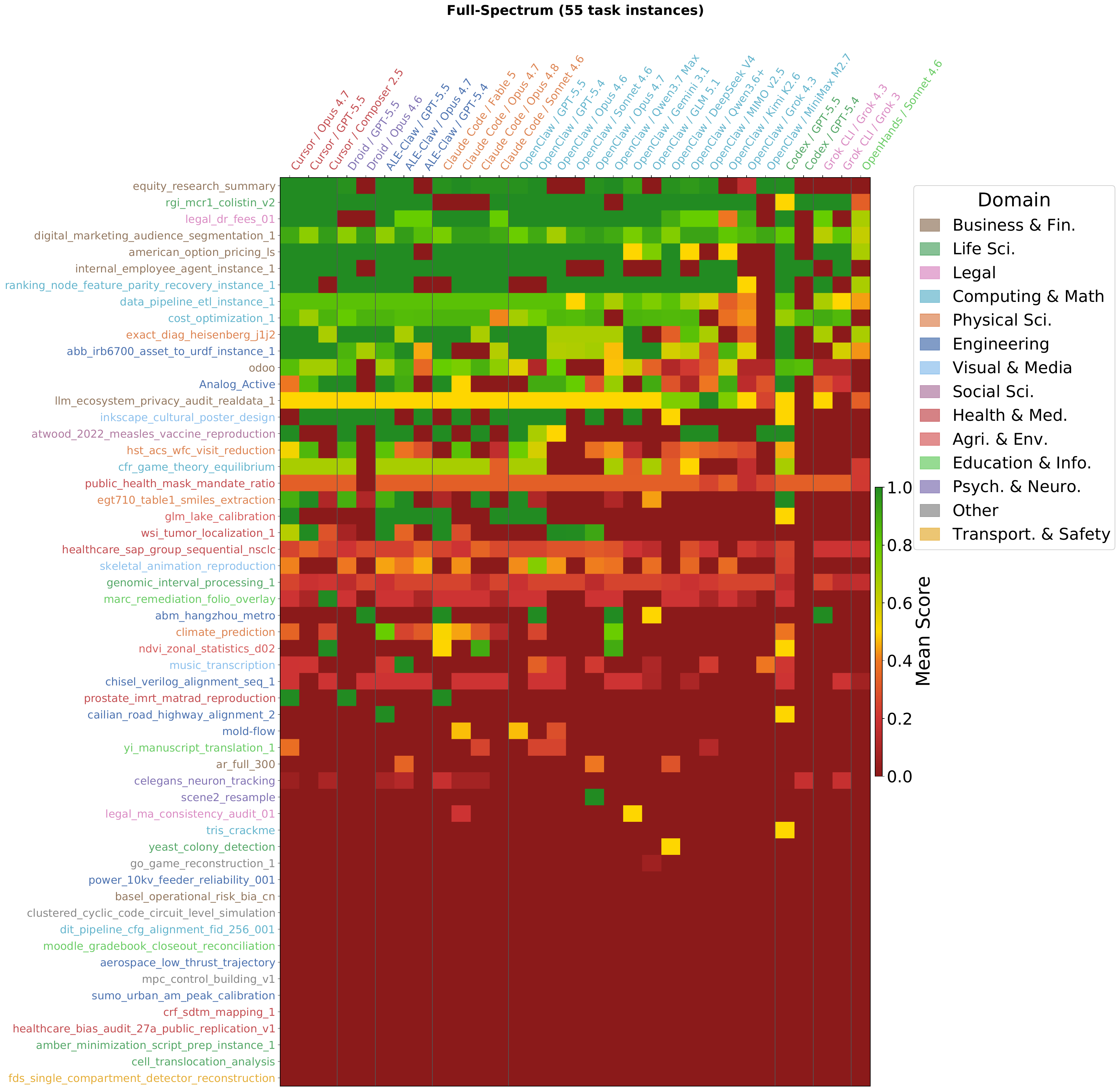}}
\caption{\textbf{Per-task instance scores: Full-Spectrum tier} (\fullspectrumcount{} task instances).}
\label{fig:heatmap-fullspectrum}
\end{figure}

\begin{figure}[p]
\centering
\makebox[\textwidth][c]{\includegraphics[width=1.25\textwidth]{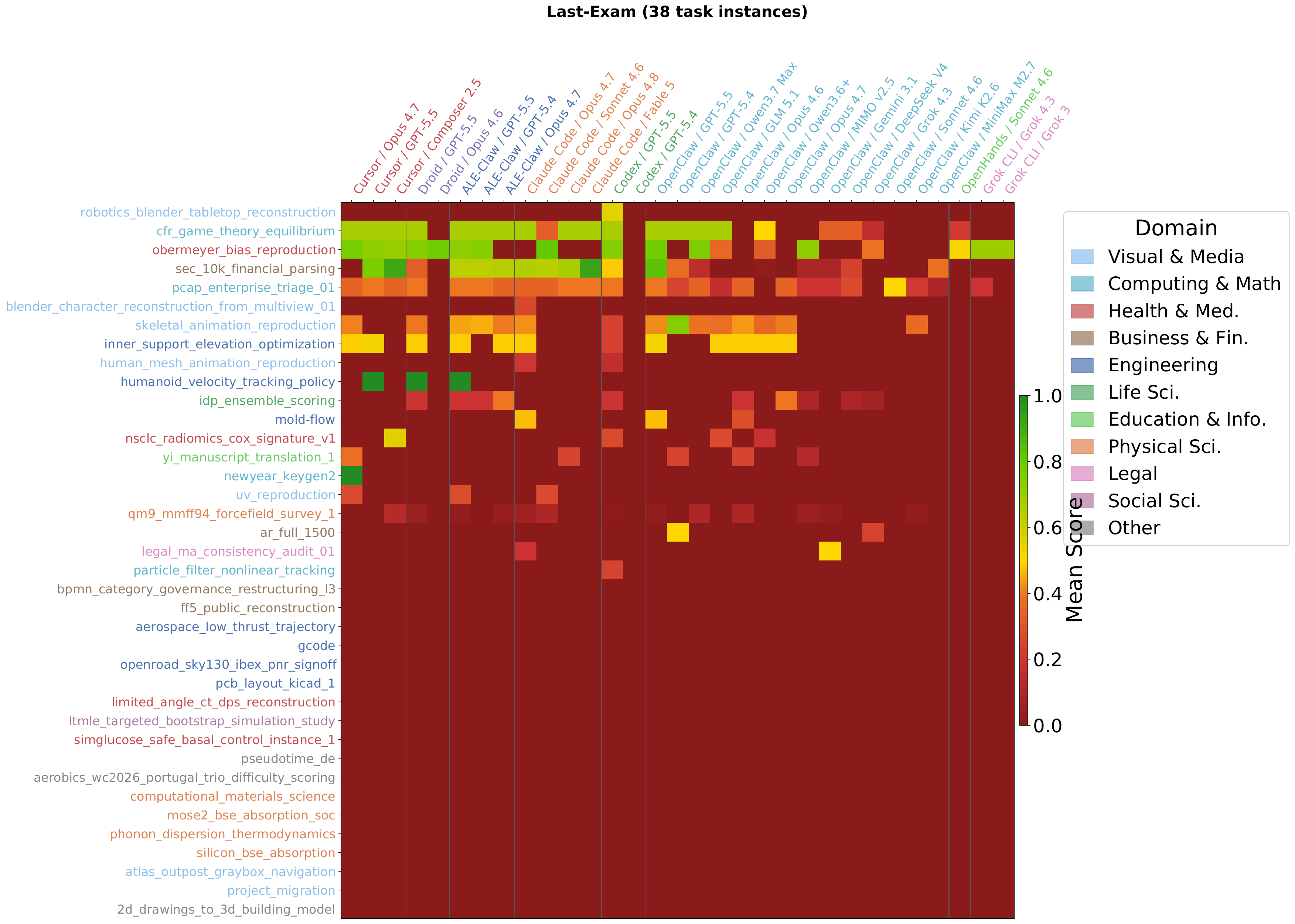}}
\caption{\textbf{Per-task instance scores: Last-Exam tier} (\lastexamcount{} task instances).}
\label{fig:heatmap-lastexam}
\end{figure}

Figures~\ref{fig:heatmap-nearterm}--\ref{fig:heatmap-lastexam} show the mean score of every task instance under each evaluated agent system.
Rows are sorted by descending average score across all systems; columns are grouped by harness and sorted by descending average score within each group.
Task instance labels are colored by taxonomy domain (legend inset).
Gray cells indicate missing runs.

\FloatBarrier

\end{document}